\documentclass[manuscript,screen]{acmart}

\makeatletter
\@namedef{r@tocindent4}{0pt}
\@namedef{r@tocindent5}{0pt}
\makeatother

\usepackage{subfigure}
\usepackage{color}
\usepackage{soul}
\usepackage{multicol}
\usepackage{multirow}
\usepackage{tabularx}
\usepackage{booktabs}
\usepackage{xcolor}
\usepackage{xargs}
\usepackage{graphbox}
\usepackage{fontawesome5}
\usepackage{listings}
\usepackage{algorithmic}
\usepackage{tcolorbox}
\usepackage[colorinlistoftodos,prependcaption,textsize=small]{todonotes}
\usepackage{graphicx}
\usepackage{longtable}
\usepackage{lscape}
\usepackage{wrapfig,lipsum,booktabs}

\AtBeginDocument{%
  \providecommand\BibTeX{{%
    \normalfont B\kern-0.5em{\scshape i\kern-0.25em b}\kern-0.8em\TeX}}}

\setcopyright{acmcopyright}
\copyrightyear{2022}
\acmYear{2022}
\acmDOI{XXXXXXX.XXXXXXX}

\acmJournal{CSUR}

\usepackage{color, colortbl}
\usepackage{pifont}
\usepackage{xspace}

\definecolor{myRed}{rgb}{0.808,0.067,0.149}
\definecolor{myGreen}{rgb}{0.067,0.708,0.149}

\usepackage{csquotes}
\usepackage{comment}
\usepackage{bm}
\usepackage{pgfplots}
\usepackage{pgfplotstable}
\usepackage{subcaption}
\usepackage{tikz}
\usepackage{tkz-tab}
\usepackage{caption}
\usepackage{latexsym}
\usepackage{amsmath}
\usepackage{graphicx}

\makeatletter
\DeclareRobustCommand\onedot{\futurelet\@let@token\@onedot}
\def\@onedot{\ifx\@let@token.\else.\null\fi\xspace}

\def\etal{\emph{et al}\onedot}
\makeatother

\usepackage{tabulary}
\usepackage{array}
\newcolumntype{x}[1]{>{\arraybackslash\hspace{0pt}}p{#1}}

\begin{document}

\title{Foundation Models for Video Understanding: A Survey}


\author{Neelu Madan}
\orcid{0000-0001-5778-3470}
\email{nema@create.aau.dk}
\affiliation{%
 \institution{Visual Analysis and Perception Lab, Aalborg University}
 \city{Aalborg}
 \country{Denmark}}
 
\author{Andreas M\o{}gelmose}
\orcid{0000-0003-0328-382X}
\affiliation{%
 \institution{Visual Analysis and Perception Lab, Aalborg University}
 \city{Aalborg}
 \country{Denmark}}

\author{Rajat Modi}
\email{rajatmodi@knights.ucf.edu}
\orcid{0009-0006-7845-4434}
\affiliation{%
\institution{Center for Research in Computer Vision, University of Central Florida}
\city{Florida}
\country{USA}}

\author{Yogesh S.~Rawat}
\email{yogesh@crcv.ucf.edu}
\orcid{0000-0003-4052-6798}
\affiliation{%
\institution{Center for Research in Computer Vision, University of Central Florida}
\city{Florida}
\country{USA}}
 
\author{Thomas B. Moeslund}
\orcid{0000-0001-7584-5209}
\affiliation{%
 \institution{Visual Analysis and Perception Lab, Aalborg University}
 \city{Aalborg}
 \country{Denmark}}


\renewcommand{\shortauthors}{Madan et al.}

\begin{abstract}
Video Foundation Models (ViFMs) aim to learn a general-purpose representation for various video understanding tasks. Leveraging large-scale datasets and powerful models, ViFMs achieve this by capturing robust and generic features from video data. This survey analyzes over 200 video foundational models, offering a comprehensive overview of benchmarks and evaluation metrics across 14 distinct video tasks categorized into 3 main categories. Additionally, we offer an in-depth performance analysis of these models for the 6 most common video tasks.
We categorize ViFMs into three categories: 1) Image-based ViFMs, which adapt existing image models for video tasks, 2) Video-Based ViFMs, which utilize video-specific encoding methods, and 3) Universal Foundational Models (UFMs), which combine multiple modalities (image, video, audio, and text etc.) within a single framework. By comparing the performance of various ViFMs on different tasks, this survey offers valuable insights into their strengths and weaknesses, guiding future advancements in video understanding. Our analysis surprisingly reveals that image-based foundation models consistently outperform video-based models on most video understanding tasks. Additionally, UFMs, which leverage diverse modalities, demonstrate superior performance on video tasks.  We share the comprehensive list of ViFMs studied in this work at: \url{https://github.com/NeeluMadan/ViFM_Survey.git} 



\end{abstract}

\ccsdesc[300]{Computing methodologies~Large-scale Pretraining, Computer Vision}

\keywords{Video Understanding, Large-scale Pretraining, Vision-language Models, Zero-shot Learning}

\maketitle

\section{Introduction}

The increasing availability of powerful 
computing resources and ever-growing datasets has fueled the development of foundation models \cite{Bommasani-arxiv-2021, Awais-arxiv-2023}. These versatile AI models, trained on massive amounts of data using self-supervised or semi-supervised learning, can be fine-tuned for various downstream tasks. Initial successes focused on static images  \cite{Clip-ICML-2021, Align-ICML-2021}, with models like CLIP \cite{Clip-ICML-2021}, and SAM \cite{Sam-ICCV-2023} achieving impressive results. Recent research \cite{Videoclip-EMNLP-2021, Video_Llama-arxiv-2023} has extended this to video domain where several pretraining strategies have been developed for Video Foundational Models (ViFMs).

While video analysis and generation has been of interest to the computer vision community for decades \cite{Karpathy-CVPR-2014, I3D-CVPR-2017, C3D-ICCV-2015, Timesformer-ICML-2021, Tong-NeurIPS-2022, Videopoet-arxiv-2023}, and the problem has largely been challenging due to complexity in tasks, additional time dimension, and volume of data. The initially developed approaches are mostly based on processing individual frames with standard image analysis techniques and additional temporal aspect on top \cite{I3D-CVPR-2017, Feichtenhofer-CVPR-2016}. 
Alternatively, more advanced techniques were developed specifically designed for videos, such as 3D convolutions \cite{Yu-CVPR-2016}, recurrent networks, use of optical-flow, and transformers \cite{AArnab-ICCV-2021, Timesformer-ICML-2021}, operating directly on videos providing better temporal modeling 
Furthermore, there has been significant research exploring the role of multiple modalities to enhance video understanding \cite{Hori-ICCV-2012, Lsmdc-CVPR-2015}.

We see a similar trend in ViFMs and their evolution also follows extending images (Image-based ViFMs), separate video modeling (Video-based ViFMs), and incorporating additional modalities, e.g., Automatic Speech Recognition (ASR) (Universal FMs). 


\begin{figure*}[t] 
    \centering
    \includegraphics[width=0.95\textwidth]{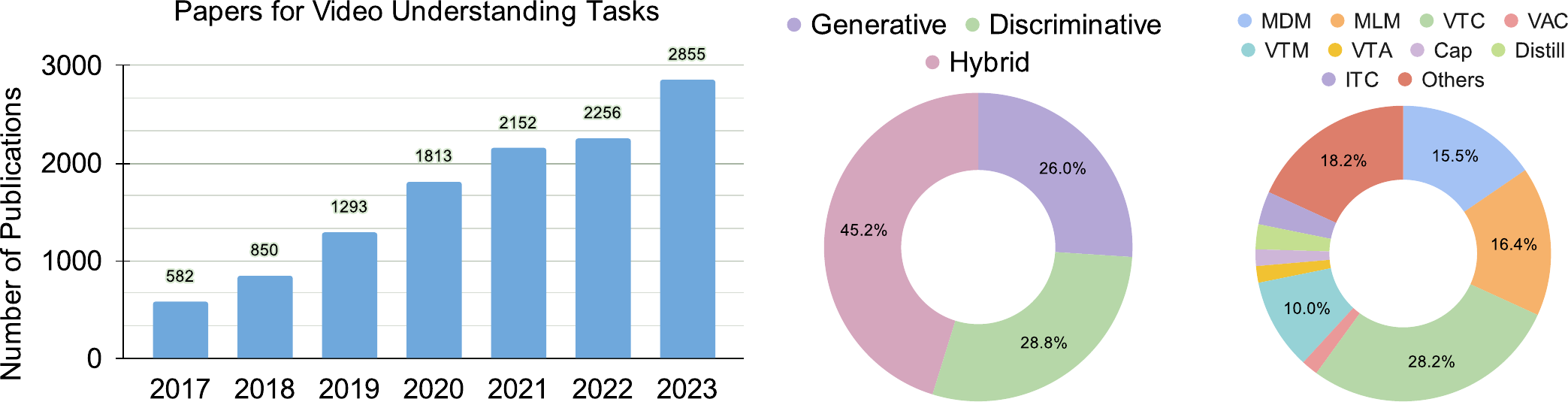} 
    \caption{Overview of recent research trends in video understanding. The \textbf{left bar chart} shows a significant increase in publications on this topic, based on data from prestigious conferences and journals. The figure presents statistics showcasing research focusing on generative, discriminative, and hybrid pretraining objectives, as depicted in the \textbf{center pie chart}. Specific pretraining objectives such as Masked Data Modeling (MDM), Masked Language Modeling (MLM), Vision-Text Contrastive (VTC), Vision-Audio Contrastive (VAC), Vision-Text Matching (VTM), Vision-Text Alignment (VTA), Captioning Loss (CAP), and Distillation Loss (Distill) are highlighted in the \textbf{right pie chart}. Best viewed in color.
    } 
    \vspace{-0.4cm}
    \label{fig:video_papers} 
\end{figure*}

\begin{figure*}[t]
    \centering
    \includegraphics[width=0.98\textwidth]{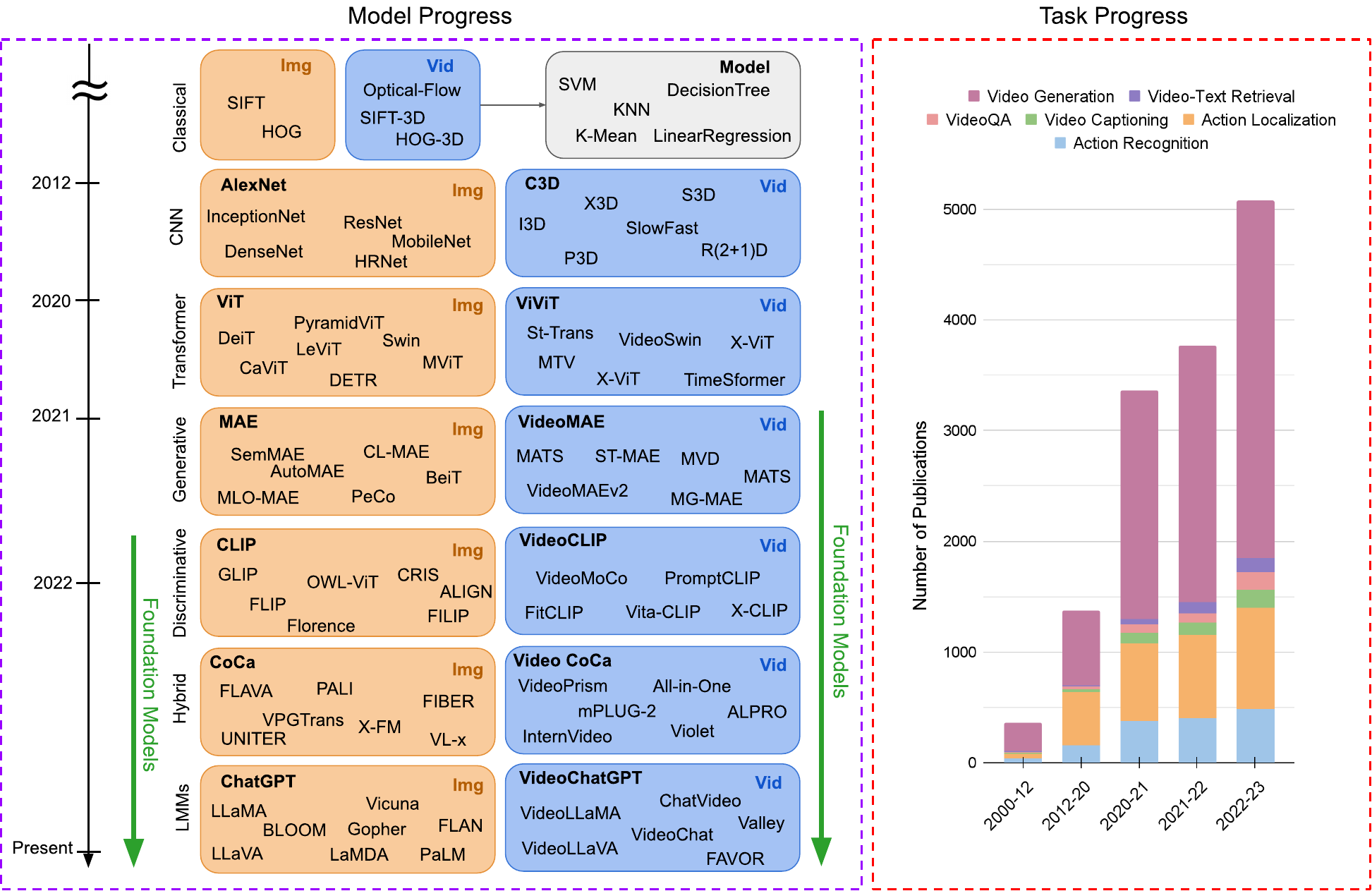}
    \caption{Figure contrasts classical (separate feature extraction, model training) and deep learning (unified framework) approaches in computer vision. It also shows the progression of deep learning approaches for both image and video processing over time (best viewed in color).}
    \vspace{-0.4cm}
    \label{fig:level1_hierarchy}
\end{figure*}

\textbf{Motivation and Contribution.}
The field of video understanding is undergoing significant advancement, as evidenced by the increasing number of research publications focused on various video understanding tasks (Figure \ref{fig:video_papers}). This growth coincides with the development of large-scale pretraining techniques. These techniques have demonstrated remarkable capabilities in adapting to diverse tasks, requiring minimal additional training with robust generalization. As a result, researchers are actively investigating the role of these foundational models to address a broad spectrum of video understanding challenges.
To navigate this rapidly evolving research landscape (See Figure \ref{fig:level1_hierarchy}), a systematic review of video understanding models is essential. We attempt to fill this critical gap by providing a comprehensive analysis of foundational models employed in video understanding tasks. We hope that this survey helps to provide a roadmap for future research directions associated with video understanding. The main contributions of our survey are:
\begin{itemize}
    \item This work presents the first comprehensive survey of foundational models (ViFMs) deployed for diverse video understanding tasks. Our survey categorizes ViFMs into three groups: i) \textit{Image-based ViFMs:} Trained solely on image data. ii) \textit{Video-based ViFMs:} Leveraging video data during training. iii) \textit{Universal Foundation Models (UFMs):} Combining various modalities (image, video, audio, text) during pretraining. 
    \item We uniquely categorize video understanding tasks based on their involvement in the temporal aspect. We further provide an extensive list of datasets and evaluation metrics associated with each categorized task.
    \item We conduct a comprehensive comparison of ViFMs from each category, analyzing various research findings. This analysis reveals valuable insights regarding the most effective ViFMs for different video understanding tasks.
    \item This survey further identifies crucial challenges faced by ViFMs, highlighting open problems that require further research attention. Additionally, we discuss promising future directions for ViFM development, paving the way for advancements in video understanding. 
\end{itemize}

\textbf{Related Surveys.} 
While several surveys have delved into specific video understanding tasks \cite{Zhou-TPAMI-2022, Zhang-TPAMI-2023} or foundational models for images \cite{Awais-arxiv-2023}, with surveys such as Shiappa et al. \cite{Schiappa-ACM-2023}, which offers an extensive review of self-supervised approaches for video understanding, the landscape has evolved significantly in recent years. With the rise of large-scale foundational models, there is a need for a comprehensive review specifically focused on these models in the context of video understanding. To the best of our knowledge, our survey is the first to provide such a comprehensive overview of foundation models for video understanding.

\begin{figure*}[htb]
    \centering
    \includegraphics[width=0.98\textwidth]{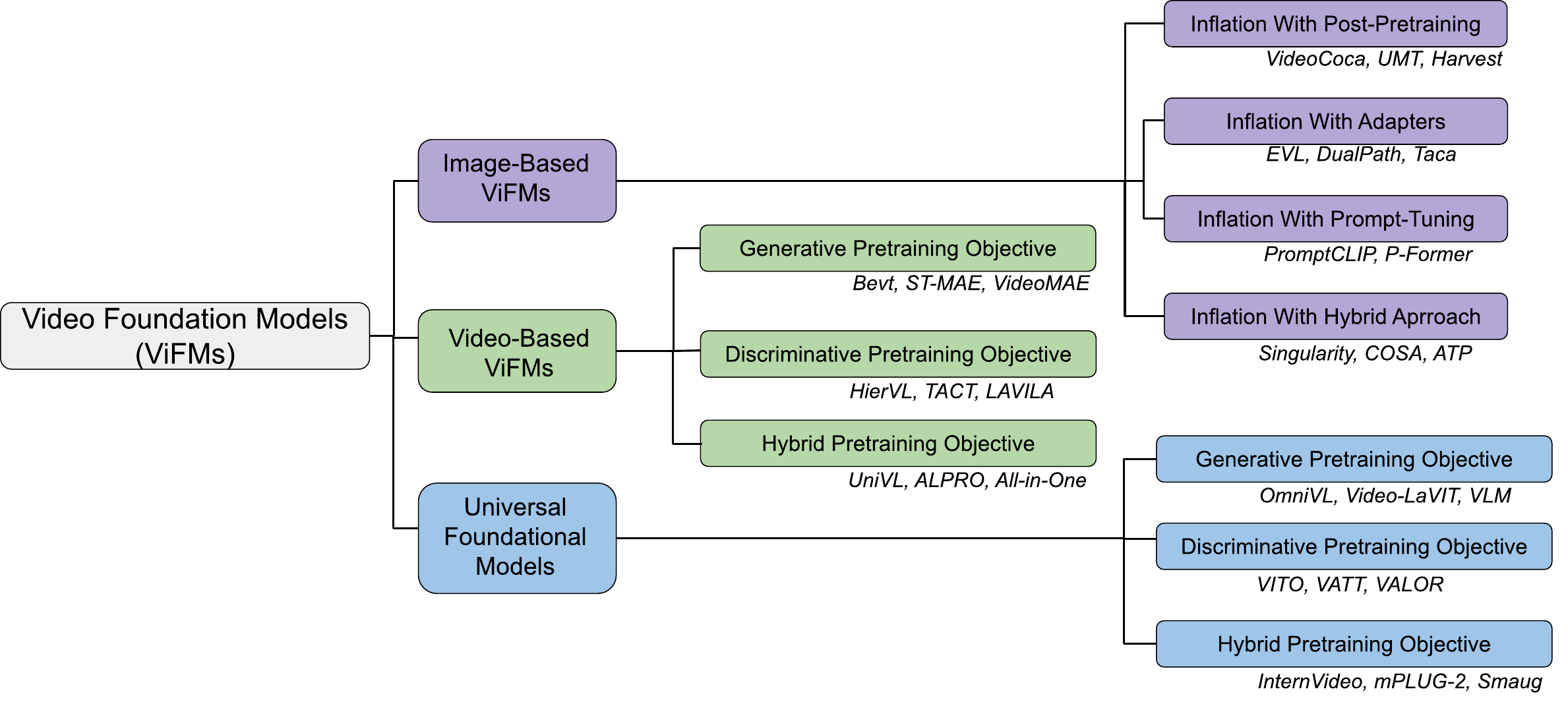}
    \caption{We need to kinda change the classification categories after the level 1(Type): \textbf{a) Adapting  Image Models}: Post-pretraining (3.1), Adapters (3.2), and Prompt-tuning (3.3); \textbf{b) Direct Video Models}: Generative (4.1), Discriminative (4.2), and Hybrid (4.3) \textbf{c) Joint Image-Video Models}: Generative (5.1), Discriminative (5.2), Hybrid (5.3), \textbf{d} Generative (6.1) and Conversational (6.2)}
    \label{fig:taxonomly}
\end{figure*}

\textbf{Paper Organization.}
In the first part of the paper (section \ref{sec:preliminaries}), we cover a wide range of video analysis tasks ranging from video classification to generation. We discuss widely used architectures and loss functions, as well as datasets relevant for large-scale pre-training. Next, we explain the main categories of ViFMs namely: Image-based ViFMs (Sec \ref{sec:image_based_vifm}), Video-based ViFMs (Sec \ref{sec:video_based_vifm}), and Universal FMs (Sec \ref{sec:universal_fms}) (See Figure \ref{fig:taxonomly} for the taxonomy). Finally (sections \ref{sec:discussion}-\ref{sec:future}), we compare and discuss the performance of the presented models, as well as present challenges and future directions for the field.







\section{Preliminaries}
\label{sec:preliminaries}

In this section, we lay the groundwork for understanding this survey. We begin by defining the diverse tasks involved in video understanding, allowing the reader to grasp various goals and challenges associated with analyzing video data. Next, we delve into the major architectural styles adopted by different foundation models. Finally, we offer a concise overview of the large-scale pretraining process, the necessary datasets used for training, and the methods used to adapt these generic models for different video tasks. 

\subsection{Video Understanding Tasks}
\label{video_understanding}

In this section, we discuss various tasks along with their popular corresponding benchmarks and evaluation metrics.

\subsubsection{Video Content Understanding} 

Computer vision tasks for video understanding fall into three primary levels. \emph{a) Abstract understanding} infers the video's overall event (e.g., sports highlight, cooking tutorial). \emph{b) Temporal understanding} zooms in, pinpointing the exact moments in a video (e.g., goal scored, ingredient added). \emph{c) Spatio-temporal understanding} goes further, identifying both the when and where of events within video frames (e.g., player celebrating on the field, chef placing vegetables in the pan). By traversing this progression, deep models build a sophisticated comprehension of video content, similar to how humans gradually grasp information. \\

\noindent
\textbf{a) Abstract Understanding Tasks.}

\noindent
\textbf{Classification or Recognition.} The task is to assign a category to a video. Different benchmarks for activity classification are: Kinetic-400 \cite{Kinetics400-arxiv-2017}, Kinetic-600 \cite{Kinetics600-arxiv-2018}, Kinetic-700 \cite{Kinetics700-arxiv-2019},  Something-Something-V1 (SSv1) \cite{SomethingSomethingv2-ICCV-2017}, Something-Something-V2 (SSv2) \cite{SomethingSomethingv2-ICCV-2017}, ActivityNet \cite{Activitynet-CVPR-2015}, HACS \cite{Hacs-ICCV-2019}, HMDB51 \cite{Jhmdb-ICCV-2011}, UCF-101 \cite{Ucf101-arxiv-2012}, TinyVIRAT \cite{Tinyvirat-ICPR-2021}, and Diving-48 \cite{Diving_48-ECCV-2018}. This task is evaluated using Top-K accuracy as a metric.  

\vspace{0.1cm}
\noindent
\textbf{Retrieval.} The task involves finding videos containing specific actions, objects, or scenes. The common metric for evaluating this task is Recall at K (R@K), specifically R@1 meaning accuracy of the first retrieved result. This task exists in the literature with different names  including: 
\emph{i) Multi-Instance Retrieval (MIR):} MIR \cite{Ashutosh-CVPR-2023}, focuses on both text-to-video (T2V) and video-to-text (V2T) retrieval. The common benchmark for this task is EPIC-Kitchen-100 \cite{EpicKitchen-ECCV-2018}, and the metrics are \textit{mAP} for V2T retrieval and \textit{normalized Discounted Cumulative Gain (nDCG)} for T2V retrieval. 
\emph{ii) Paragraph-to-video (P2V) retrieval:} P2V retrieval \cite{Sun-NeurIPS-2022} bridges the gap between language and video, finding videos relevant to a paragraph (several sentences). Different datasets to solve this task includes ActivityNet Captions \cite{ActivityNetCaptions-ICCV-2017}, QuerYD \cite{Queryyd-ICASSP-2021}, and CondensedMovie \cite{CondensedMovies-ACCV-2020}. 
\emph{iii) Text-to-video (T2V) retrieval:} T2V retrieval \cite{Hitea-CVPR-2023} finds videos using textual descriptions (single sentence). Different benchmarks for this task are Kinetic-Geb \cite{KGeb-ECCV-2022}, MSRVTT \cite{Msrvtt-CVPR-2016}, DiDeMo \cite{Didemo-ICCV-2017}, YouCook2 \cite{Youcook2-AAAI-2018}, LSMDC \cite{Lsmdc-CVPR-2015}. \\


\noindent
\textbf{b) Temporal Understanding Tasks.}

\noindent
\textbf{Temporal Action Localization (TAL).} TAL \cite{Internvideo-arxiv-2022} aims to pinpoint the exact moments within videos where specific actions occur. Common benchmarks for this tasks are THUMOS-14 \cite{Thumos-CVIU-2017}, ActivityNet-v1.3 \cite{Activitynet-CVPR-2015}, HACS Segment \cite{Hacs-ICCV-2019}, FineAction \cite{Fineaction-TIP-2022}, BreakFast \cite{Breakfast-CVPR-2014}, Charades \cite{Charades-ECCV-2016}, Ikea-ASM \cite{IkeaAsm-WACV-2021}. The evaluation metric is the mean average precision (mAP) and the average precision (AP) \cite{Thumos-CVIU-2017, Activitynet-CVPR-2015} for each action category. 

\vspace{0.1cm}
\noindent
\textbf{Fine-grained Clasification.} This task extends the classification task for long-form videos \cite{Sun-NeurIPS-2022}, where COIN \cite{Coin-CVPR-2019} and LVU \cite{Lvu-CVPR-2021} are the benchmarking datasets. COIN \cite{Coin-CVPR-2019} proposes Procedural Activities Classification (PAC), where the task is to divide complex actions into meaningful subactions and then learn the correct order and hierarchical-relationship among these subactions. LVU \cite{Lvu-CVPR-2021} proposes 9 tasks including content understanding (relationship, speaking style, scene/place) for the comprehensive video understanding. These fine-grained classification tasks thus require spatio-temporal understanding of videos. \\

\noindent
\textbf{c) Spatio-Temporal Understanding Tasks.}

\noindent
\textbf{Spatiotemporal Action Localization (SAL).}  SAL aims to find both "when" and "where" specific actions unfold within a video\cite{Internvideo-arxiv-2022}. Notable datasets for this particular category are UCF101-24 \cite{Ucf101-arxiv-2012}, JHMDB-22 \cite{jhuang2013towards}, and UCF-MAMA\cite{modi2022video}. These datasets contain annotation for each video frame. Datasets like Ava\cite{Ava-CVPR-2018}, Ava-Kinetics\cite{Ava_Kinetic-arxiv-2005} contain box-annotations at 1Hz sampling frequency over a clip of 15 mins. Evaluation metric for this task is the f-mAP and video-mAP\cite{duarte2018videocapsulenet} measuring frame-level and video-level localization performance respectively. (mAP: mean average precision)



\vspace{0.1cm}
\noindent
\textbf{Tracking.} The task aims at identifying and following the movement of objects throughout a video. KITTI \cite{Geiger-IJRR-2013}, UA-DETRAC \cite{Rujikietgumjorn-AVSS-2017}, LaSOT \cite{Lasot-CVPR-2019}, MOT16/MOT17 \cite{Milan-arXiv-2016}, MOT20 \cite{Dendorfer-Arxiv-2020}, MOTSynth \cite{Fabbri-ICCV-2021}, BDD-100K \cite{Bdd100k-CVPR-2020}, TAO \cite{Tao-ECCV-2020}, BURST \cite{Burst-WACV-2023}, and LV-VIS \cite{Lvvis-ICCV-2023} are popular benchmarking datasets; and HOTA \cite{Hota-IJCV-2021}, and Clear-MOT \cite{ClearMOT-EURASIP-2008} are evaluation metrics for this task. Recently, a more fine-grained tracking approach known as \textit{point tracking} has emerged, which tracks specific points on an object's surface regardless of pixel location. Datasets for point tracking include PointOdyssey \cite{Pointodyssey-ICCV-2023}, TAP-Vid-DAVIS \cite{TapVid-NeurIPS-2022} and CroHD \cite{Sundararaman-CVPR-2021}, while evaluation focuses on Mean Trajectory Error (MTE) and position accuracy.

\vspace{0.1cm}
\noindent
\textbf{Referring Video Segmentation (RVS).} RVS \cite{Glee-arxiv-2023} segments the objects referred by either textual descriptions or first frame's segmentation. The benchmarks for RVS using textual description are RefCOCOg \cite{RefCocoG-ECCV-2016}, Refer-Youtube-VOS \cite{Seo-ECCV-2020}, Refer-DAVIS \cite{Khoreva-ACCV-2019}, A2D-Sentences \cite{Gavrilyuk-CVPR-2018,duarte2019capsulevos}, and JHMDB-Sentences \cite{Gavrilyuk-CVPR-2018};  and evaluation metrics for this task is meanAP and meanIoU.  Whereas, The benchmarks for RVS using first frame's segmentation as reference are Youtube-VOS , DAVIS \cite{Davis-CVPR-2016}.

\vspace{0.1cm}
\noindent
\textbf{Video Object Detection (VOD).} VOD typically aims to detect object across a video-stream. Videos typically contain a lot of redundant temporal information. This helps detectors detect an object in the current frame and  \textit{anticipate} it's position in the subsequent\cite{liu2023objects,zhou2022transvod}. Reported Metrics are generally mAP with results reported for different speed of motion of objects. Benchmark datasets are ImageNet-VID\cite{russakovsky2015imagenet}.



\begin{figure*}[t!]
    \centering
    \includegraphics[width=\textwidth]{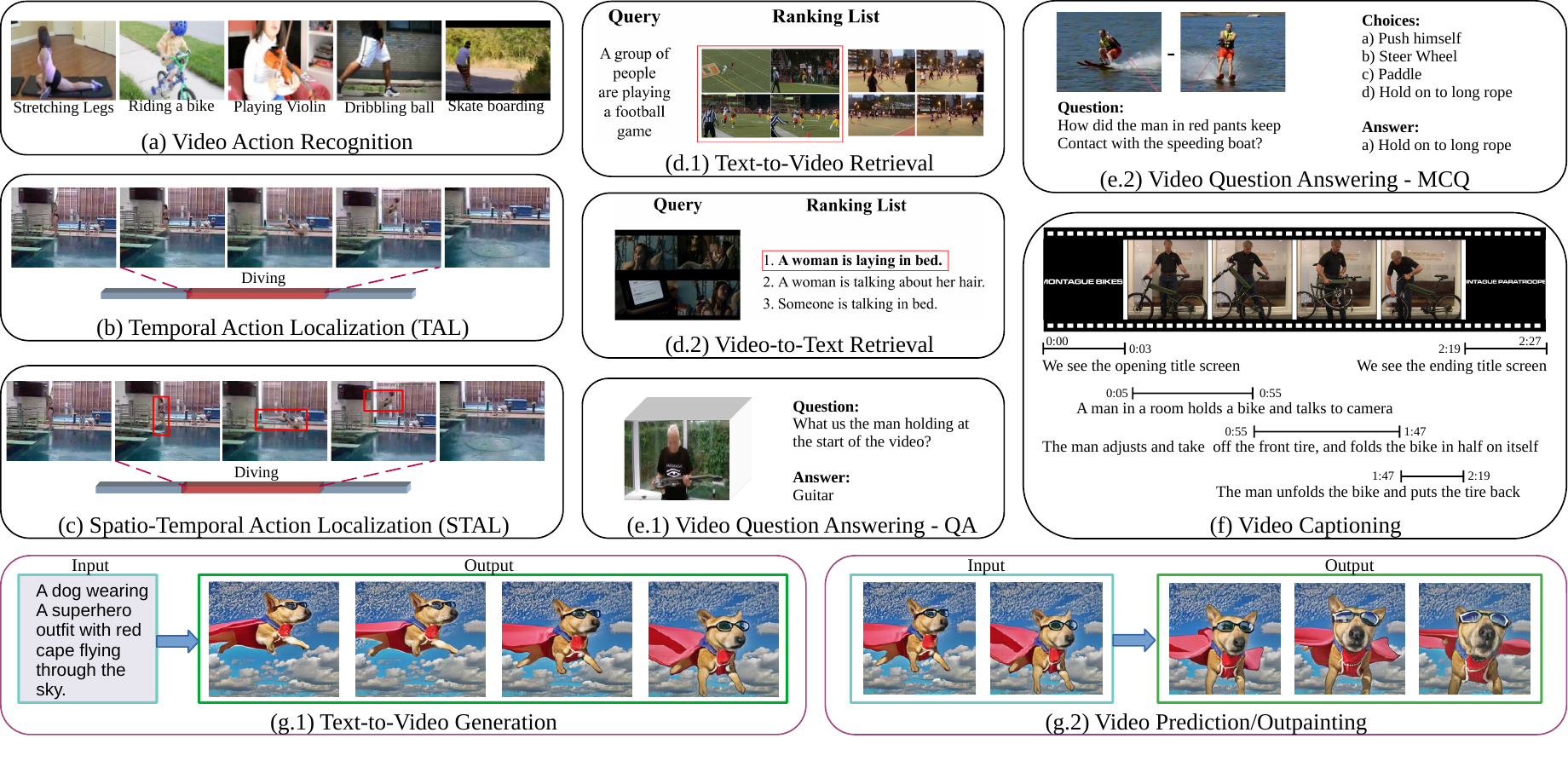}
    \vspace{-0.6cm}
    \caption{Figure presents different video tasks. Task in first column: (a) Video Action Recognition, b) Temporal Action Localization (TAL), and c) Spatio-temporal Action Localization (STAL) require only video understanding. Tasks in second and third column: d) Video-Text Retrieval, e) VideoQA, and f) Video Captioning requires both video and language understanding. Best viewed in color.
    }
    \label{fig:video_understanding_tasks}
\end{figure*}

\subsubsection{Descriptive Understanding Tasks} 
This section covers benchmarks and evaluation metrics associated with Video Question Answering (VQA), and Video Captioning tasks. Both VQA and Captioning tasks focus on understanding of the textual description of the video content. \\

\noindent
\textbf{Video question answering (VQA).}
VQA answers questions about the video content based on visual information and potentially textual queries. VQA is evaluated using Top-1, Top-K accuracy, DA score and ANLS as a metric\cite{wu2023omni, mathew2021docvqa}. According to the literature, this task is sub-divided into three sub-categories: \textit{i) Multiple-Choice (MC):} MC-VQA addresses multiple-choice question answering. Common benchmarks for this subtask are TGIF-Action and TGIF-Transition \cite{Tgif_vqa-CVPR-2017}, MSRVTT-MC \cite{Msrvtt_vqa-ECCV-2018}, and LSMDC-MC \cite{Lsmdc_vqa-arxiv-2016}. \textit{ii) Open-Ended (OE):} OE-VQA answers subjective, creative, and logical questions. Common benchmarks for this subtask are TGIF-Frame \cite{Tgif_vqa-CVPR-2017}, MSRVTT-QA \cite{Msrvtt_vqa-ECCV-2018}, MSVD-QA \cite{Msvd_vqa-ACMMM-2017}, LSMDC-FiB \cite{Lsmdc_vqa-arxiv-2016}, ActivityNet-QA \cite{Activitynet_vqa-AAAI-2019}. \textit{iii) Long-Form (LF):} LF-VQA \cite{Sun-NeurIPS-2022} goes beyond single answers, generating comprehensive explanations that understand video content, reason temporally, and adapt to diverse question types. Common benchmarks for this subtasks are ActivityNet-QA \cite{Activitynet_vqa-AAAI-2019}, How2QA \cite{How2qa-arxiv-2020} and VIOLIN \cite{Violin-CVPR-2020}. 

\noindent
\textbf{Video Captioning.} 
Video captioning generates textual descriptions of video content \cite{Hitea-CVPR-2023}. MSRVTT \cite{Msrvtt-CVPR-2016}, Youcook2 \cite{Youcook2-AAAI-2018}, and MSVD \cite{Msvd-ACLHLT-2011} are the common benchmarks to solve this task. The evaluation metric for this task are BLEU@4 \cite{Bleu-ACL-2002}, METEOR \cite{Meteor-ACLW-2005}, ROUGE \cite{Rouge-ACLW-2004}, and CIDEr \cite{Cider-CVPR-2015}. 


\subsubsection{Video Content Generation and Manipulation}
This section covers benchmarks and evaluation metrics for various generative video tasks. \\

\noindent
\textbf{Video Prediction.} This area of research encompasses two main sub-tasks: a) \textit{Video future prediction (VFP)}: VFP predicts future frames, given an input video of variable length. Literature uses K600 \cite{Kinetics600-arxiv-2018} as a benchmarking dataset and FVD as an evaluation metric for this task; b) \textit{Long-Term Anticipation (LTA)}: LTA \cite{Ashutosh-CVPR-2023} predicts next 20 actions given the current action (verb, noun). The common benchmark for this task is Ego4D \cite{Ego4D-CVPR-2022} and the metric to evaluate the performance is \textit{Edit Distance (ED)} \cite{Ego4D-CVPR-2022}. 

\noindent
\textbf{Text-to-video (T2V) Generation.}  The process \cite{Videopoet-arxiv-2023} involves generating video frames based on a textual prompt. Common benchmarks for this task are MSR-VTT \cite{Msrvtt-CVPR-2016} and UCF-101 \cite{Ucf101-arxiv-2012}, while using Fr\'{e}chet Video Distance (FVD) \cite{FVD-arxiv-2018}, CLIP Similarity Score (CLIPSim) \cite{Godiva-arxiv-2021} and Inception Score (IS) \cite{Saito-IJCV-2020} as evaluation metrics. 

\noindent
\textbf{Video inpainting/outpainting.} The task involves predicting the video with the contents filled-in on a masked video using SSv2 \cite{SomethingSomethingv2-ICCV-2017} as benchmarking dataset and FVD \cite{FVD-arxiv-2018} as evaluation metrics. 

\noindent
\textbf{Video stylization.} 
This task involves generating a video whose style is governed by an additional modality such as text or optical flow\cite{Videopoet-arxiv-2023}. Existing methods try to preserve high-level content of the video, and generating a temporally-consistent stylized version. Benchmark dataset and evaluation metrics for this task are DAVIS 2016 \cite{Davis-CVPR-2016} and CLIPSim \cite{Godiva-arxiv-2021} respectively.

\begin{figure*}[htb]
    \centering
    \includegraphics[width=0.98\textwidth]{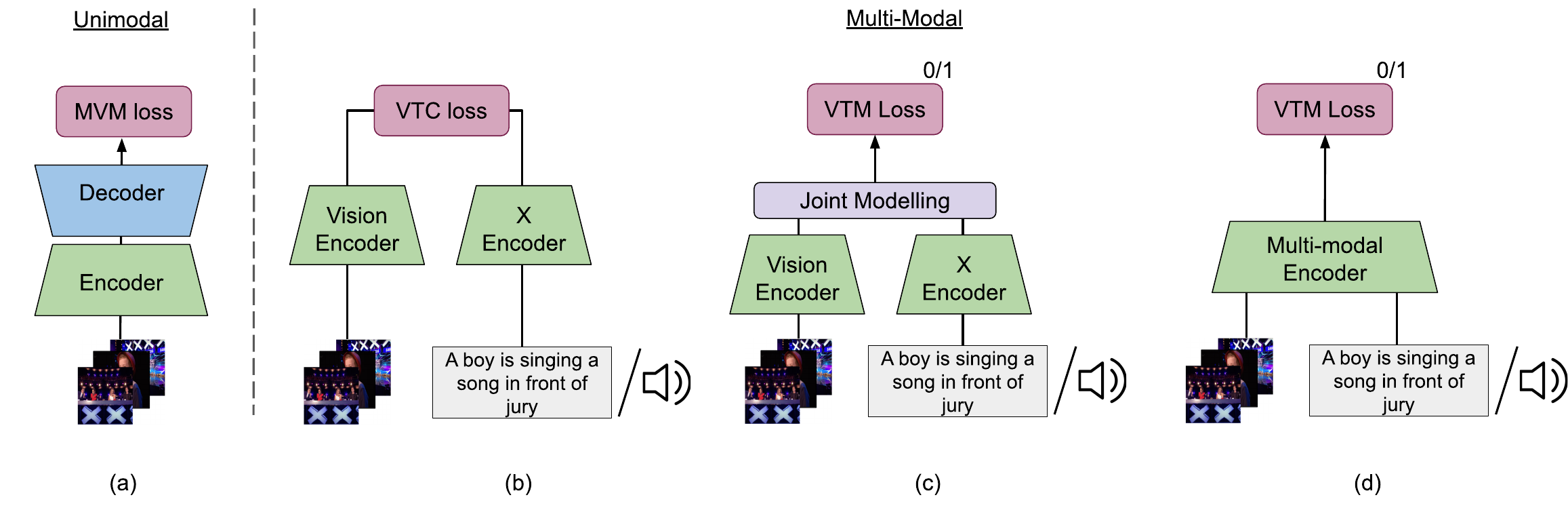}
    \caption{Figure shows different architectures adopted by Video Foundation Models (ViFMs): Uni-modal ViFMs follows usually (a) Encoder-Decoder network, and multi-modal foundation model follows either (b) Joint-Encoder (c) Dual-Encoder (n==2) or Multi-Encoder (n$>$2), and (d) Mix-Encoder. Best viewed in color.}
    \label{fig:architectures_vifm}
    \vspace{-0.2cm}
\end{figure*}

\subsection{Architectures and Loss Functions}

This section provides a brief description of the architectures and different training objectives adopted by ViFMs. Figure~\ref{fig:architectures_vifm} illustrates different patterns adopted by ViFMs. These patterns can be broadly categorized based on modality (type of data processed) and loss function. 

\subsubsection{Architectures} Different architecture patterns in ViFMs can be broadly categorized as: \emph{unimodal} and \emph{multi-modal}. Multi-modal ViFM further show different patterns as discussed below. \\

\noindent
\textbf{Uni-Modal.}
Unimodal ViFMs \cite{VideoMAEv2-CVPR-2023a, Feichtenhofer-NeurIPS-2022} focus on a single modality (e.g., video) and typically employ generative objectives like mask reconstruction. They often follow an \emph{Encoder-Decoder }framework where the encoder extracts features from the input video, and the decoder reconstructs the masked or missing parts. The training process is guided by a reconstruction loss function that measures the difference between the original and reconstructed video.

\vspace{0.1cm}
\noindent
\textbf{Multi-Modal.}
Multi-modal ViFMs \cite{Li-ICCV-2023, Clover-CVPR-2023} handle multiple modalities (e.g., video and text) and usually rely on contrastive learning objectives. They typically use encoder-only networks where separate encoders might be employed for each modality. Here, the focus is on learning representations that capture the relationships between different modalities. A contrastive loss function (\emph{VTC}) is used during training to pull together similar representations and push apart dissimilar ones. Within multi-modal models, the encoding style for different modalities can be further classified into three categories:
i) \emph{Joint-Encoder} \cite{Omnivore-CVPR-2022, Omivl-NeurIPS-2022, Omnimae-CVPR-2023} utilizes a single encoder to process all modalities simultaneously. This is computationally efficient but may not capture modality-specific nuances.
ii) \emph{Dual/Multi-Encoder} \cite{Clip-ICML-2021, Ag_Adapter-ICCV-2023, Fitclip-arxiv-2022, PromptClip-ECCV-2022} employs separate encoders for each modality. While it allows for more specialized feature extraction, it increases computational complexity. Dual encoders are used for two modalities, while architectures handling more than two modalities are referred to as multi-encoders.
iii) \emph{Mixed-Encoder} \cite{Alpro-CVPR-2022, Hdvila-CVPR-2022, Clover-CVPR-2023, Violetv2-CVPR-2023a} offers a compromise between joint and separate encoders. It first uses lightweight encoders to extract initial features from individual modalities. These features are then combined and processed by a shared encoder before reaching the final loss function. Both unimodal and multi-modal architectures often leverage transformer blocks as their basic building blocks. 

\subsubsection{Loss Functions}

Video Foundation Models (ViFMs) rely on pre-training with specific objectives (loss functions) to learn effective representations. These objectives can be broadly categorized into two main approaches: generative and discriminative. This section delves into the details of each category and the specific objectives associated with them. \\

\noindent
\textbf{Discriminative.}
The most common objective functions for ViFMs pretraining are \textit{Video-Text Contrastive (VTC)} \cite{Videoclip-EMNLP-2021} and \textit{Video-Text Matching (VTM)} \cite{Alpro-CVPR-2022}. VTC pulls together similar representations and pushes apart dissimilar ones, while VTM aims to maximize the matching score between a given video-text pair. These objectives are well-suited for multi-modal architectures but might not be directly applicable to uni-modal architectures. However, contrastive objectives using data augmentation to generate positive pairs have been explored for uni-modal settings \cite{Videomoco-CVPR-2021, Byol-NeurIPS-2020, Simclr-ICLR-2020}. Notably, VIMPAC \cite{Vimpac-arxiv-2021} leverages this approach for ViFMs pretraining.

Beyond VTC and VTM, several variants have been proposed. \textit{Verb-Focused Contrastive (VFC)} \cite{Momeni-ICCV-2023} focuses on fine-grained verb alignment.  \textit{Video-Text Joint (VTJ)} \cite{Luo-arxiv-2020} learns a joint representation from video and text (Figure \ref{fig:architectures_vifm} (d) for reference). \textit{Multimodal Temporal Contrastive (MTC)} \cite{Sun-NeurIPS-2022} expands contrastive learning to other modalities, while \textit{Video Clip Contrastive (VCC)} \cite{Vimpac-arxiv-2021} utilizes contrastive learning between video clips. Recent works explore \textit{Tri-Modal Alignment (TMA)} \cite{Clover-CVPR-2023} for simultaneous cross-modal alignment and fusion, as well as \textit{Omni-Modality Video-Caption Contrastive (OM-VCC)} \cite{Vast-arxiv-2023} and \textit{Video-Caption Matching (OM-VCM)} \cite{Vast-arxiv-2023} losses. Additionally, \textit{Video-Audio Contrastive (VAC)} \cite{Gong-arxiv-2023, Akbari-NeurIPS-2021} and \textit{Attention-Guided Contrastive (AGC)} \cite{Parthasarathy-NeurIPS-2023} objectives have been introduced.

The presence of the additional temporal dimension in video allows for significant flexibility in designing discriminative objectives. Several objectives aim to improve temporal modeling capabilities. \textit{Multi-modal temporal relation exploration (MTRE}) \cite{Hitea-CVPR-2023} and \textit{cross-modal moment exploration (CME)} \cite{Hitea-CVPR-2023} leverage text guidance to enhance the model's ability to capture temporal context in video. \textit{Time-Order Consistency Check (TOCC)} \cite{Bagad-ICLRW-2023} ensures the correct order of events, while \textit{Control Task (CT)} \cite{Bagad-ICLRW-2023} enforces matching between video events and corresponding text descriptions. \textit{Discriminative Video Dynamics Modeling (DVDM)} \cite{Paxion-NeurIPS-2023} specifically promotes nuanced temporal understanding. \textit{Frame-Transcript Matching (FTM)} \cite{Merlot-NeurIPS-2021} matches video frames with their corresponding transcripts, and \textit{Temporal Reordering (TR)} \cite{Merlot-NeurIPS-2021} predicts the correct order of scrambled video frames.

\noindent
\textbf{Generative.}
This category encompasses various objectives that focus on reconstructing masked information within the video data. Examples include \emph{Masked Language Modeling (MLM)} for text data \cite{Hdvila-CVPR-2022, Violetv2-CVPR-2023a}, \emph{Mask Video Modeling (MVM)} for videos \cite{VideoMAEv2-CVPR-2023a, Mgmae-ICCV-2023}, \emph{Mask Signal/Data Modeling (MDN, MSM)} for general signals \cite{Gong-arxiv-2023}, \emph{Mask Frame Modeling (MFM)} for video frames \cite{VLM-IJCNLP-2021}, and \emph{Mask Image Modeling (MIM)} for images \cite{Wang-CVPR-2022}. The primary goal here is to reconstruct the masked parts, such as predicting a masked frame in MFM.  Building upon these core objectives UniVL \cite{Luo-arxiv-2020} proposes \emph{Conditional MLM (CMLM)} \cite{Luo-arxiv-2020}, and \emph{Conditional MFM (CMFM)} \cite{Luo-arxiv-2020}. These objectives are primarily used for training unimodal architectures. However, for pre-training multi-modal architectures, generative objectives are often combined with a contrastive loss (acting as a discriminative objective). This combination leverages the strengths of both approaches: reconstructing masked information and learning relationships between modalities.

Beyond the aforementioned objectives, there are less commonly used generative approaches for multi-modal architectures. These include:
\textit{Auto-regressive training objectives} like Language Modeling (LM) \cite{Hitea-CVPR-2023}, PrefixLM \cite{Hitea-CVPR-2023}, and next (Image/Motion/Text) \cite{VideoLavit-arxiv-2024} token generation, which predicts the next element (token) in a sequence based on the current one. 
\emph{Captioning Loss} \cite{Yan-arxiv-2023b} predicts next token based on past video and text. VAST \cite{Vast-arxiv-2023} modifies this concept with \emph{Omni-Modality Video Caption Generation (OM-VCG)} loss.  \emph{Audio Video Continuation (AVCont)} \cite{Videopoet-arxiv-2023} predicts next frame from audio, \emph{Image Inpainting and Outpainting}, \cite{Videopoet-arxiv-2023} \emph{Text-to-Image/Video Generation} \cite{Videopoet-arxiv-2023}, \emph{Video-to-Text Completion (VTC)} \cite{Fu-CVPR-2023b}, and \emph{Frame Prediction} \cite{Videopoet-arxiv-2023}.  

Beyond general-purpose objectives, task-specific options exist. \textit{Prompting Entity Modeling (PEM)} \cite{Alpro-CVPR-2022} focuses on fine-grained region-entity alignment and action understanding. \textit{Multi-Grained Aligning (MGA)} \cite{Zeng-arxiv-2023} aligns visual concepts (objects) with text descriptions, while \textit{Multi-Grained Localization (MGL)} \cite{Zeng-arxiv-2023} locates these concepts in images based on textual descriptions. \textit{Multi-Choice Modeling (MCM)} \cite{Xu-arxiv-2023} improves modality alignment and representation learning. \textit{Distillation loss} \cite{Fitclip-arxiv-2022, Taca-arxiv-2023}, where a student network mimics the representation of a stronger teacher network, is another common ViFM pretraining objective employed for knowledge transfer, particularly for training lightweight student networks.


\subsection{Training Strategy}
This sections discusses various datasets involved with large-scale pretraining of foundation models. Moreover, this section also briefly mentions the recipe for deploying these models for different video tasks. 

\subsubsection{Self-supervised Pretraining Datasets}
We discuss pretraining datasets for both unimodal and multi-modal architectures in ViFMs in this section. \\

\noindent
\textbf{Unimodal.}
Large-scale models in single modality mostly used a combination of action recognition dataset for self-supervised pre-training. K400 \cite{Kinetics400-arxiv-2017}, K600 \cite{Kinetics600-arxiv-2018}, K700 \cite{Kinetics700-arxiv-2019}, SomethingSomethingV1 (SSv1) \cite{SomethingSomethingv2-ICCV-2017}, and SomethingSomethingV2 (SSv2) \cite{SomethingSomethingv2-ICCV-2017} are few datasets used for such cases.

\noindent
\textbf{Multi-modal.}
Video-text dataset used for training multi-modal video foundation models are listed as: WebVid-2M \cite{WebVid2M-CVPR-2021}, HowTo100M \cite{Howto100m-ICCV-2019}, EpicKitchen \cite{EpicKitchen-ECCV-2018}, Flinstones \cite{Flintstones-ECCV-2018}, Mugen \cite{Mugen-ECCV-2022}. As we have limited number of multi-modal (Vision-Language) dataset for the video domain. To fulfill this requirement, some foundation models \cite{Alpro-CVPR-2022, Violet-arxiv-2022, Violetv2-CVPR-2023a, Wang-CVPR-2023, Valor-arxiv-2023, Tang-WACV-2023} used image-text dataset such CC3M \cite{CC3M-ACL-2018}, and CC12M \cite{Conceptual12m-CVPR-2021}, and SBU Captions \cite{Sbcaptions-NeurIPS-2011} for the pretraining of multi-modal video foundation models. Moreover, few foundation models \cite{Hdvila-CVPR-2022, Sun-NeurIPS-2022, Vast-arxiv-2023} further curates their own datasets such HD-VILA-100M \cite{Hdvila-CVPR-2022} and LF-VILA-8M \cite{Hdvila-CVPR-2022}, and VAST-27M \cite{Vast-arxiv-2023} in order to provide diverse and large-scale dataset for multi-modal pretraining.

\subsubsection{Semi-supervised Pretraining Datasets}

The recent trajectory of multimodal foundation model research reveals a compelling trend towards the development of increasingly generic models, capable of tackling a broad spectrum of tasks across both video and image domains \cite{VLM-IJCNLP-2021, Omivl-NeurIPS-2022, Omnivore-CVPR-2022, Zhu-arxiv-2023, Valor-arxiv-2023, Merlot-NeurIPS-2021, Videopoet-arxiv-2023, Mplug2-arxiv-2023, Gong-arxiv-2023, Tang-WACV-2023}. \\

\noindent
\textbf{Combining Datasets}
The grounding task in the visual domain is not merely solved by self-supervised learning. Preparing large-scale annotated datasets is also a tedious task in such cases. Therefore, different models use a combination of datasets:  Object365 \cite{Objects365-ICCV-2019}, OpenImages \cite{Openimages-IJCV-2020}, and COCO \cite{Coco-ECCV-2014} for object detection; RefCOCO \cite{RefCoco-ECCV-2016}, RefCOCO+ \cite{RefCoco-ECCV-2016}, RefCOCOg \cite{RefCocoG-ECCV-2016}, and VisualGenome \cite{Krishna-IJCV-2017} for visual grounding; LVIS \cite{Lvis-CVPR-2019}, BDD \cite{Bdd100k-CVPR-2020} for tracking; YTVIS19 \cite{Ytvis19-ICCV-2019}, YTVIS21 \cite{Ytvis21-2021}, RVOS \cite{Seo-CVPR-2022} and OVIS \cite{Ovis-IJCV-2022} for video segmentation.

\noindent
\textbf{Pseudo-labelled Datasets}
The requirement for large-scale annotated data remains a challenge in computer vision. A recent trend involves leveraging a few powerful teacher models to provide high-quality labels associated with different visual tasks. This approach was pioneered by GRIT \cite{Grit-arxiv-2023}, which utilizes a teacher model to generate labels for grounding tasks. Following this success, SAM \cite{SegmentAnything-ICCV-2023} proposes an active learning approach to generate high-quality labeled data specifically for the segmentation task. This approach resulted in the creation of a very large-scale dataset, SA-1B \cite{SegmentAnything-ICCV-2023}, containing 1 Billion high-quality annotations. Similarly, Distill VLM \cite{Zhao-arxiv-2024} leverages a teacher model to generate captions for existing video datasets like VideoCC \cite{Nagrani-ECCV-2022} and InternVid \cite{InternVid-arxiv-2023}. This process creates two new pseudo-captioned datasets: VideoCC$^+$, and InternVid$^+$. \\

\begin{figure}[tb]
    \centering
    \includegraphics[width=.65\textwidth]{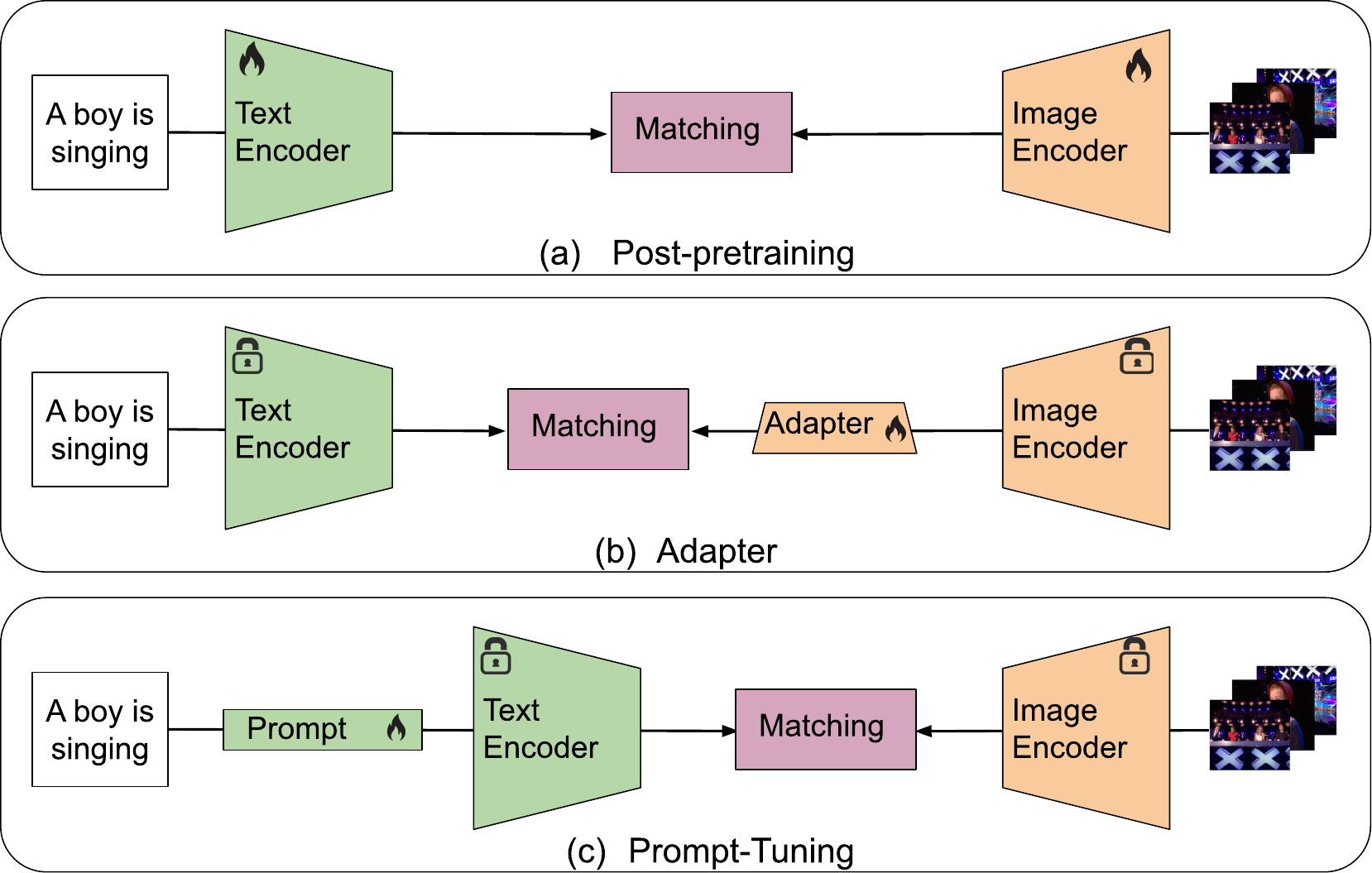}
    \caption{Figures illustrate different methods for inflating image models for video understanding task: a) Post-pretraining (encoders are trainable), b) Adapters (encoders are frozen with few additional trainable parameters), and c) Prompt-tuning (trainable parameter at the input of the text encoder). Best viewed in color.
    }
    \vspace{-0.7cm}
    \label{fig:image_based_vifms}
\end{figure}

\subsubsection{Deploying Foundation Models for Video Understanding}

We can establish a foundation model trained on a large-scale dataset, but deploying it for video understanding tasks (Section \ref{video_understanding}) requires further steps. \\


\noindent
\textbf{Fine-Tuning.}
Fine-tuning a model remains a powerful technique for adapting it to specific video tasks \cite{Luo-Neurocomputing-2022, tang2021clip4caption}. However, its capabilities extend beyond that. Fine-tuning can also be used to improve the model's generic representation, meaning its overall ability to understand and handle various types of information. This type of fine-tuning is often referred to as \emph{post-pretraining} \cite{Li-ICCV-2023, Yan-arxiv-2023b, Harvest-arxiv-2023, Fitclip-arxiv-2022} because it essentially trains the model again to expand its knowledge base.

One approach to achieve this effective integration of visual modalities with LLMs is through a process called \emph{instruction-tuning} \cite{InstructBlip-NeurIPS-2023}. This process involves fine-tuning the additional module, or sometimes the entire LLM, on an instructional dataset. The idea behind instruction-tuning was first introduced by InstructBLIP \cite{InstructBlip-NeurIPS-2023}. Instruction has become a common practice for both static images \cite{Blip-ICML-2022, Instructblip-arxiv-2023, Llama-arxiv-2023} and dynamic videos \cite{Video_Chatgpt-arxiv-2023, MacawLLM-arxiv-2023}.

\noindent
\textbf{Adapters.}
For video understanding, adapters \cite{Adapter-ICML-2019} offer a powerful and efficient approach. These lightweight neural network modules are strategically integrated within large pre-trained models.  Their key strength lies in requiring training only a limited number of parameters, significantly reducing the computational burden compared to fine-tuning the entire model. This efficiency makes them ideal for video tasks, which often involve processing vast amounts of data. Adapters excel in this domain due to their dual functionality: 1) improving the model's representation for specific tasks \cite{fang2021clip2video, Aim-ICLR-2022, St_adapter-NeurIPS-2022, M2clip-arxiv-2024, ZeroI2V-arxiv-2023} and 2) extending its capabilities to enhance the overall understanding of videos \cite{Ag_Adapter-ICCV-2023, Evl-ECCV-2022, Qing-ICCV-2023, Wang-ACMMM-2023, Taca-arxiv-2023}, effectively creating a more generic video representation.

\noindent
\textbf{Prompt-Tuning.}
Similar to adapter networks, prompt-tuning \cite{Prompttuning-EMNLP-2021} offers a computationally efficient approach for adapting large pre-trained models to new tasks. It achieves this by introducing a small number of additional trainable parameters at the model's input, in the form of a prompt. This prompt essentially guides the pre-trained model towards the desired task by providing specific instructions or context. Similar to adapter, they are integrated with large-scale models to improve: 1) representation for specific task \cite{Huang-CVPR-2023, Wasim-CVPR-2023, Engin-ICCVW-2023}, and 2) extending the overall video understanding. \cite{PromptClip-ECCV-2022, Videoprompter-arxiv-2023} Alternatively, we can design generic prompts that enhance the model's overall performance on various tasks, e.g., \textit{"image of {object}"} as used by CLIP \cite{Clip-ICML-2021}. By focusing on a small set of trainable parameters, prompt-tuning significantly reduce the computation complexity.

\begin{table*}
\centering
\caption{Table shows the video foundation models (ViFMs) generated via inflating image models, further divided into sub-categories as: a) Post-pretraining (further fine-tuning on image-models), b) Adapters (introduce few trainable layers inside a pretrained image model), c) Prompt-Tuning (introduce fe trainable parameters at the input of a pretrained model), d) Hybrid (using combination of mentioned techniques). These approaches depict two prominent architectural patterns namely: ED (Encoder-Decoder) and DE (Dual-Encoder). Additionally, the code corresponding to the approach can be followed through the hyperlink. The highlighted column shows that the method requires no additional pretraining.}
\setlength\tabcolsep{4.0pt}
\footnotesize
\resizebox{1\textwidth}{!}{%
\begin{tabular}{|c|l|x{2.2cm}x{1cm}|x{2cm}|lx{2.8cm}|r|}
\toprule
 & \multirow{3}{*}{Method} & \multicolumn{2}{|c|}{\underline{Pretraining data}} & & \multicolumn{2}{|c|}{\underline{Architecture}} & \multirow{3}{*}{Venue} \\ 
 & & Dataset(s) & Size & Pretraining & Type & Base & \\ 
 & & & & Objectives & & & \\ 
\midrule

\multirow{18}{*}{\rotatebox{90}{Post-pretraining}} & VideoCoCa \cite{Yan-arxiv-2023b} &  VideoCC3M \cite{Nagrani-ECCV-2022}, HowTo100M \cite{Howto100m-ICCV-2019} & 103M & VTC, Captioning & ED & ViT \cite{Vit-arxiv-2020}, Transformer \cite{Vaswani-NeurIPS-2017} & arxiv'22 \\

& \href{https://github.com/OpenGVLab/unmasked_teacher}{UMT} \cite{Li-ICCV-2023} & K710 \cite{Uniformerv2-arxiv-2022}, WebVid-2M \cite{WebVid2M-CVPR-2021}, CC3M \cite{CC3M-ACL-2018}, COCO \cite{Coco-ECCV-2014}, Visual Genome \cite{Krishna-IJCV-2017}, SBU Captions \cite{Sbcaptions-NeurIPS-2011}, CC12M \cite{Conceptual12m-CVPR-2021} & 25M & VTM, MLM, MVM &  DE & ViT \cite{Vit-arxiv-2020}, BERT \cite{Bert-NAACL-2019}, CLIP-ViT \cite{Clip-ICML-2021} & ICCV'23 \\

& \href{https://github.com/lucidrains/MaMMUT-pytorch}{MaMMUT} \cite{Mammut-MLR-2023} &  Web alt-text \cite{Jia-ICML-2021} & 1.8B & ITC, Captioning & ED & TubeViT \cite{Vit-arxiv-2020} & TMLR'23 \\

& \href{https://github.com/OpenGVLab/InternVideo}{Harvest} \cite{Harvest-arxiv-2023} &  WebVid-10M \cite{WebVid2M-CVPR-2021} & 10M & VTC, MLM &  ED & UniformerV2 \cite{Uniformerv2-arxiv-2022}, Transformer \cite{Vaswani-NeurIPS-2017}, CLIP \cite{Clip-ICML-2021} & arxiv'23 \\

& \href{https://github.com/microsoft/XPretrain/tree/main/CLIP-ViP}{CLIP-ViP} \cite{Xue-arxiv-2023} & WebVid-2.5M \cite{WebVid2M-CVPR-2021}, HD-VILA-100M \cite{Hdvila-CVPR-2022} & 102M & VTC & DE & ViT \cite{Vit-arxiv-2020} & ICLR'23  \\

& \href{https://github.com/bryant1410/fitclip}{FitCLIP} \cite{Fitclip-arxiv-2022} & WebVid-2.5M \cite{WebVid2M-CVPR-2021} & 4M & VTC, Distill & DE & CLIP-ViT \cite{Clip-ICML-2021} & BMVC'22  \\

& Distill-VLM \cite{Zhao-arxiv-2024} & S-Mit \cite{Smit-CVPR-2021}, WebLI \cite{Webli-arxiv-2022} & 400K & VTC & DE & Pali-3 \cite{Pali3-arxiv-2023}, Vit-G \cite{Vitg-CVPR-2022}, UL-2 \cite{UL2-ICLR-2022} & CVPR'24 \\

\hline 
\multirow{7}{*}{\rotatebox{90}{Adapters}} & \cellcolor{blue!10} \href{https://github.com/OpenGVLab/efficient-video-recognition}{EVL} \cite{Evl-ECCV-2022} & \multicolumn{2}{c|}{\cellcolor{blue!10} CLIP Pretrained} & \cellcolor{blue!10} VTC & \cellcolor{blue!10} DE & \cellcolor{blue!10} CLIP \cite{Clip-ICML-2021}, Transformer \cite{Vaswani-NeurIPS-2017} & \cellcolor{blue!10} ECCV'22 \\

& \cellcolor{blue!10} \href{https://github.com/park-jungin/DualPath}{DualPath} \cite{Park-CVPR-2023} & \multicolumn{2}{c|}{\cellcolor{blue!10} CLIP Pretrained} & \cellcolor{blue!10} VTC & \cellcolor{blue!10} DE & \cellcolor{blue!10} ViT \cite{Vit-arxiv-2020}, Transformer \cite{Vaswani-NeurIPS-2017} & \cellcolor{blue!10} CVPR'23 \\

& \cellcolor{blue!10} AG-Adapter \cite{Ag_Adapter-ICCV-2023} & \multicolumn{2}{c|}{\cellcolor{blue!10} CLIP Pretrained} & \cellcolor{blue!10} VTC & \cellcolor{blue!10} DE & \cellcolor{blue!10} CLIP \cite{Clip-ICML-2021}, LLM \cite{Llama-arxiv-2023} & \cellcolor{blue!10} ICCV'23  \\

& \href{https://github.com/alibaba-mmai-research/DiST}{DiST} \cite{Qing-ICCV-2023} & Kinetics-710 \cite{Kinetics700-arxiv-2019} & 0.5M & VTC & DE & CLIP-ViT \cite{Clip-ICML-2021} & ICCV'23 \\

& \cellcolor{blue!10} \href{https://github.com/SCZwangxiao/RTQ-MM2023}{RTQ} \cite{Wang-ACMMM-2023} & \multicolumn{2}{c|}{\cellcolor{blue!10} BLIP Pretrained} & \cellcolor{blue!10} VTC, VTM & \cellcolor{blue!10} DE & \cellcolor{blue!10} BLIP-ViT \cite{Blip-ICML-2022} & \cellcolor{blue!10} ACMMM'23 \\

& TaCA \cite{Taca-arxiv-2023} & LAION-400M \cite{Laion-arxiv-2021} & 400M & VTC, Distillation & DE & ViT \cite{Vit-arxiv-2020}, BERT \cite{Bert-NAACL-2019} & arxiv'23  \\

& PaLM2-VAAapter \cite{Palm2_VAadapter-arxiv-2024} & WebLI \cite{Webli-arxiv-2022}, VTP \cite{Flamingo-NeurIPS-2022}, SMiT \cite{Smit-CVPR-2021} & - & VTA & DE & CoCa \cite{Coca-arxiv-2022}, PaLM 2 \cite{Palm2-arxiv-2023} & arxiv'24 \\

\hline

 \multirow{6}{*}{\rotatebox{90}{Prompt-Tuning}} & \cellcolor{blue!10} \href{https://github.com/StanfordVL/atp-video-language}{ATP} \cite{buch2022revisiting} & \multicolumn{2}{c|}{\cellcolor{blue!10} CLIP Pretrained} & \cellcolor{blue!10} - & \cellcolor{blue!10} DE & \cellcolor{blue!10} CLIP \cite{Clip-ICML-2021} & \cellcolor{blue!10} CVPR'22 \\

 & \href{https://github.com/yiren-jian/BLIText}{P-Former} \cite{jian2023bootstrapping} &  LAION \cite{Laion-NeurIPS-2022}, COCO \cite{Coco-ECCV-2014}, Visual Genome \cite{Krishna-IJCV-2017}, CC-3M \cite{CC3M-ACL-2018}, SBU Captions \cite{Sbcaptions-NeurIPS-2011} & 16M & ITC, ITM, ITG & ED & EVA-CLIP \cite{Evaclip-arxiv-2023}, LLM \cite{Opt-arxiv-2022}, Q-Former \cite{Blip2-arxiv-2023} & NeurIPS'23 \\

 & \cellcolor{blue!10} VideoPrompter \cite{Videoprompter-arxiv-2023} & \multicolumn{2}{c|}{\cellcolor{blue!10} CLIP Pretrained} & \cellcolor{blue!10} - & \cellcolor{blue!10} DE & \cellcolor{blue!10} CLIP \cite{Clip-ICML-2021}, GPT-3.5 \cite{Gpt4-2023}, ViFi-Clip \cite{VifiClip-CVPR-2023}, AIM \cite{Aim-ICLR-2022}, Action-CLIP \cite{Actionclip-arxiv-2021} & \cellcolor{blue!10} arxiv'23 \\

\hline

 \multirow{6}{*}{\rotatebox{90}{Hybrid}} & \href{https://github.com/jayleicn/singularity}{Singularity} \cite{Lei-arxiv-2022} & COCO \cite{Coco-ECCV-2014}, VG \cite{Krishna-IJCV-2017}, SBU Captions \cite{Sbcaptions-NeurIPS-2011}, CC3M \cite{CC3M-ACL-2018}, CC12M \cite{Conceptual12m-CVPR-2021}, WebVid-2M \cite{WebVid2M-CVPR-2021} & 17M & VTC, MLM & DE &  ViT \cite{Vit-arxiv-2020}, BERT \cite{Bert-NAACL-2019} & ACL'23  \\

 & \cellcolor{blue!10} \href{https://github.com/ju-chen/Efficient-Prompt}{PromptCLIP} \cite{PromptClip-ECCV-2022} & \multicolumn{2}{c|}{\cellcolor{blue!10} CLIP Pretrained} & \cellcolor{blue!10} VTC & \cellcolor{blue!10} DE & \cellcolor{blue!10} CLIP-ViT \cite{Clip-ICML-2021} & \cellcolor{blue!10} ECCV'22 \\

 & \href{https://github.com/TXH-mercury/COSA}{COSA} \cite{Chen-arxiv-2023} & CC14M \cite{Valor-arxiv-2023}, WebVis-2.5M \cite{WebVid2M-CVPR-2021} & 14M & CITC, CITM, CMLM, CGM & DE & ViT \cite{Vit-arxiv-2020}, BERT \cite{Bert-NAACL-2019} & ICLR'24 \\

\bottomrule
\end{tabular}
}
\label{tab:AIM}
\end{table*}

\section{Image-based Video Foundation Models}
\label{sec:image_based_vifm}

Image-based Video-Foundation Models (ViFMs) are constructed by leveraging pre-trained Image Foundation Models (IFMs) and inflating them for video tasks. This paper explores three primary approaches for inflating IFMs for video: post-pretraining (Section~\ref{post_pretraining}), adapters (Section~\ref{adapters}), and prompt-tuning (Section~\ref{prompt_tuning}). Each of these approaches can be employed to generate either a general-purpose ViFM (see Table~\ref{sec:image_based_vifm}) or a task-specific model tailored for video question answering (VQA), recognition, or captioning.

\subsection{Inflation With Post-pretraining}
\label{post_pretraining}

Post-pretraining refines pre-trained IFMs on large video datasets, enhancing their capability for video-centric tasks. We explore both general-purpose ViFMs (detailed in Table~\ref{sec:image_based_vifm}) in Section \ref{post_pretraining_generalist} and specialized models in Section \ref{post_pretraining_specialist} obtained via inflating IFMs.

\noindent
\subsubsection{Generalist Models}
\label{post_pretraining_generalist}

In this section, we explore major trends in post-pretraining for video understanding, focusing on leveraging pre-trained image foundation models (IFMs). This approach involves adapting powerful image models to video tasks.

Several methods achieve this by directly applying pre-trained models to video frames. \emph{ClipBERT} \cite{Clipbert-CVPR-2021} utilizes CLIP \cite{Clip-ICML-2021} as the IFM, employing techniques like sparse sampling and end-to-end learning. Conversely, \emph{VideoCoCa} \cite{Yan-arxiv-2023b} utilizes a pre-trained contrastive captioner (CoCa) \cite{Coca-arxiv-2022} as the IFM, directly applying CoCa's functionalities to video frames. However, these methods didn't fully optimize video-pretraining. Addressing this, \emph{Harvest} \cite{Harvest-arxiv-2023} focuses on post-training efficiency. It refined pre-trained image models like CLIP \cite{Clip-ICML-2021} through novel methods during post-training on large datasets, significantly boosting training speed, fostering cross-modal fusion, and achieving state-of-the-art performance on diverse video-language tasks.
Building on post-training efficiency, \emph{CLIP-ViP} \cite{Xue-arxiv-2023} proposes auxiliary captions to bridge the domain gap between images and videos. CLIP-ViP introduces video proxy tokens for efficient processing and Omnisource Cross-modal Learning to jointly learn from diverse data sources, significantly improving CLIP's \cite{Clip-ICML-2021} video-text retrieval performance and providing valuable insights for effective post-pretraining strategies in video-text representation learning.

\vspace{0.1cm}
\noindent
\textbf{Student-Teacher Framework for Video Adaptation.}
\emph{FitCLIP} \cite{Fitclip-arxiv-2022} enables the student to learn from video-text pairs and distills knowledge from a pre-trained CLIP model (teacher). \emph{Unmasked Teacher (UMT)} \cite{Li-ICCV-2023} trains the student model in two stages, aligning visible token representations with the teacher (a pretrained IFM) in Stage 1 and refining the student's understanding in Stage 2 through text alignment and knowledge distillation. \emph{Distill-VLM} \cite{Zhao-arxiv-2024} utilizes a two-stage training process to adapt the pre-trained IFM, effectively distilling knowledge for video tasks and generating high-quality, detailed video captions crucial for video understanding and various video-language tasks. This enables Distill-VLM \cite{Zhao-arxiv-2024} for generating pseudo-labels for large datasets like VideoCC+ ($\sim$10M Clips) and InternVid+ ($\sim$234M clips), which improves the representation \cite{Zhao-arxiv-2024, Internvideo-arxiv-2022} when employed during the pretraining. 

\subsubsection{Specialist Models}
\label{post_pretraining_specialist}
We narrow our focus to instances where post-pretraining is employed to tailor IFMs for specific video tasks. 

\noindent
\textbf{Post-pretraining for Retrieval and Captioning.} Clip4Clip \cite{Luo-Neurocomputing-2022} and Clip4Caption \cite{tang2021clip4caption} present empirical studies (involving feature pooling, and fine-tuning etc.) on video-text post-training's efficiency and its impact on video-text representation for retrieval and captioning tasks. Clip2Caption \cite{tang2021clip4caption} first post-pretrains CLIP \cite{Clip-ICML-2021} on a video dataset and then finetunes it for the captioning task. On the other hand, \emph{CLIP4Clip} \cite{Luo-Neurocomputing-2022} surpasses state-of-the-art on various datasets by leveraging pre-trained CLIP \cite{Clip-ICML-2021} for video-text retrieval.

\subsection{Inflation With Adapters}
\label{adapters}

Instead of fine-tuning an entire pre-trained image model, we use lightweight adapter layers for temporal modeling. This adapts the model for video tasks without needing extra pre-training. The sections explores the method using adapters to inflate pretrained image models. 

\subsubsection{Generalist Models} 

These adapters integrate with pre-trained image foundation models (IFMs) using only a small number of additional parameters. This allows for efficient generic video representation learning while maintaining the power of pre-trained models. \textit{DiST} \cite{Qing-ICCV-2023} disentangles spatial and temporal learning. This framework leverages a pre-trained image recognition model for spatial understanding and a lightweight temporal encoder to capture dynamic changes, offering an efficient and effective solution for video understanding tasks.
\textit{RTQ} \cite{Wang-ACMMM-2023} clusters redundant tokens and refines redundant information. It further extends the method for complicated video understanding by modeling temporal relations using adapters and querying task-specific information.
\textit{EVL} \cite{Evl-ECCV-2022} presents a framework that directly trains video models on frozen CLIP \cite{Clip-ICML-2021} features (powerful visual representations learned from image-text pairs). Unlike fine-tuning, EVL uses a lightweight Transformer decoder and a local temporal module to efficiently learn spatiotemporal features without retraining the image backbone.
\textit{DualPath} \cite{Park-CVPR-2023} employs two distinct paths: a) Spatial path, which encodes individual frame appearance with minimal tuning, using only a few frames at a low frame rate; and b) Temporal path, which captures dynamic relationships by constructing a grid-like frameset from consecutive low-resolution frames. It then introduces light adapters to both paths for inflating image models for videos.
\textit{AGAdapter} \cite{Ag_Adapter-ICCV-2023} proposes two adapter modules: KaAdapter aligns the video-text representation, and PgAdapter employs prompt-tuning to leverage LLMs for captioning.
\textit{PaLM2-VAdapter} \cite{Palm2_VAadapter-arxiv-2024} proposes an adapter module that effectively aligns vision encoders and LLMs using a progressive training strategy, effectively integrating visual information into LLMs.
Task-agnostic Compatible Adapter (TaCA) \cite{Taca-arxiv-2023} enables seamless integration of new foundation models into existing frameworks without retraining, preserving model strengths through lightweight adapters.

\subsubsection{Specialist Models} 

Adapters also illustrate a wide range of applications for inflating IFMs of specific video tasks. We show some distinguished examples in this section. 

\noindent
\textbf{Adapters for Text-Video Retrieval.} 
Adapters play a crucial role in text-video retrieval tasks, as demonstrated by various approaches such as PromptSwitch \cite{Deng-ICCV-2023}, Clip2Video \cite{Fang-TMM-2022}, Cross-Modal Adapter \cite{Jiang-arxiv-2022}, AdaCLIP \cite{Hu-ACMMM-2023}, and CrossVTR \cite{Dai-arxiv-2023}. \textit{PromptSwitch} \cite{Deng-ICCV-2023} introduces a "Prompt Cube" to CLIP's image encoder, capturing global video semantics efficiently, while \textit{CrossVTR} \cite{Dai-arxiv-2023} introduces a decoupled video-text cross-attention module to handle spatial and temporal multimodal information separately. \textit{CLIP2Video} \cite{Fang-TMM-2022} simplifies the task into spatial representation and temporal relations, achieving state-of-the-art performance on retrieval benchmarks. Meanwhile, \textit{Cross-Modal Adapter} \cite{Jiang-arxiv-2022} achieves significant parameter efficiency through adapter-based layers, facilitating realignment of CLIP's feature spaces. \textit{AdaCLIP} \cite{Hu-ACMMM-2023} addresses the challenge of frame selection and aggregation, offering a tailored system for practical deployment on resource-constrained devices and cloud pipelines.

\noindent
\textbf{Adapters for VQA.} Tem-Adapter \cite{Chen-ICCV-2023} tackles Video Question Answering (VideoQA) challenges by bridging domain gaps in image-based pre-trained models. It introduces visual and textual aligners: the Visual Temporal Aligner predicts future states based on past video and textual cues, while the Textual Semantic Aligner refines textual embeddings using question-answer pairs and video sequences, enabling effective adaptation for VideoQA tasks.

\noindent
\textbf{Adapters for Video Action Recognition.} While powerful, adapting pre-trained image models for video tasks can be expensive and prone to overfitting. \emph{AIM} \cite{Aim-ICLR-2022}, \emph{M$^2$-CLIP} \cite{M2clip-arxiv-2024}, \emph{ST-Adapter} \cite{St_adapter-NeurIPS-2022}, and \emph{ZeroI2V} \cite{ZeroI2V-arxiv-2023} tackles this challenge using parameter-efficient adapter modules. AIM \cite{Aim-ICLR-2022}, ST-Adapter \cite{St_adapter-NeurIPS-2022}, and M$^2$-CLIP \cite{M2clip-arxiv-2024} introduces \textit{joint spatial-temporal adapters}, Temporal Enhancement and Difference modeling (\textit{TED-Adapters}), and Spatio-Temporal Adapter (\textit{ST-Adapter}) respectively into a pre-trained CLIP model. Different from these, ZeroI2V \cite{ZeroI2V-arxiv-2023} provides a novel \textit{zero-cost} transfer learning method and proposes a new \textit{spatio-temporal attention} mechanism that captures video dynamics without extra effort. 

\noindent
\textbf{Adapters for Temporal Action localization (TAL).}
To address challenges in Zero-shot temporal action detection (TAD), \textit{ZEETAD} \cite{Phan-arxiv-2023} introduces two key modules: 1) a dual-localization module pinpointing action-relevant regions and generating proposals using semantic embeddings; 2) a zero-shot proposal classification module using efficiently fine-tuned CLIP \cite{Clip-ICML-2021} models with adapters for better transferability to the video domain.

\subsection{Inflation With Prompt-Tuning}
\label{prompt_tuning}

\begin{figure*}[h]
    \centering
    \includegraphics[width=0.98\textwidth]{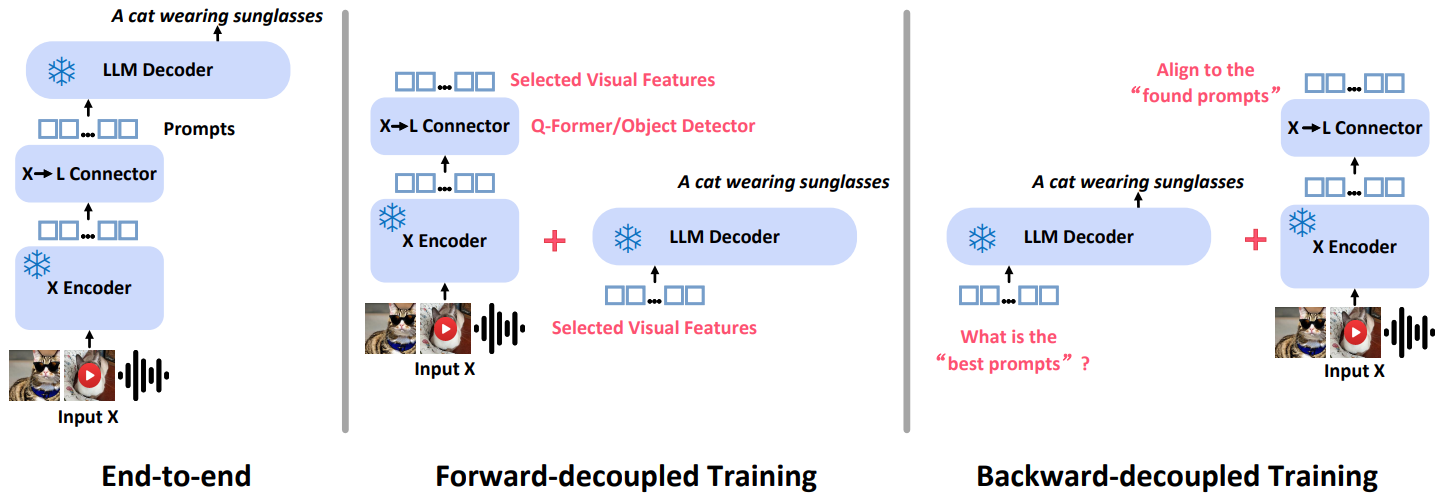}
    \caption{Figure shows three different approaches integrating X (can be image/video/audio) modality with LLMs namely: i) \textbf{End-to-end} (left) \cite{WebVid2M-CVPR-2021, Clipcap-arxiv-2021}: First aligning the features using 'X$\rightarrow$L Connector' and provide them as prompts, ii) \textbf{Forward-decoupled Training} (middle) \cite{Blip2-arxiv-2023, XLLM-arxiv-2023}: 'X$\rightarrow$L Connector' is trained to extract relevant features, which are then provided directly to LLMs. iii) \textbf{Backward-decoupled Training} (right) \cite{Jian-arxiv-2023}: First identify a reference frame for LLMs and then align feature using 'X$\rightarrow$L Connector'. Image taken from \cite{Jian-arxiv-2023}. Best Viewed in color.}
    \label{fig:promptingLLMs}
    \vspace{-0.4cm}
\end{figure*}

Similar to adapters, prompt tuning improves efficiency by only fine-tuning a few additional parameters. These parameters however are incorporated at the input layer (See Fig. \ref{fig:promptingLLMs})

\subsubsection{Generalist Models}
Our review highlights the increasing use of prompt-tuning with large language models (LLMs) for video understanding. \textit{Jian \etal}~\cite{Jian-arxiv-2023} propose a \emph{Prompt-Transformer (P-Former)}, training solely on language data to predict optimal prompts for LLMs. This backward-decoupling approach (See Fig \ref{fig:promptingLLMs}) significantly boosts performance across various tasks and architectures, making it adaptable and modality-agnostic. \textit{VideoPrompter} \cite{Videoprompter-arxiv-2023} enhances zero-shot performance of existing VLMs by leveraging video-specific information. It employs LLMs with video-specific prompts to generate detailed descriptions and attributes for each class, enriching their representation beyond just the name. This approach offers plug-and-play compatibility with existing VLMs and consistent performance improvements across various video understanding tasks.

A different approach, \textit{Atemporal Probe (ATP)} \cite{buch2022revisiting} leverages pre-trained image-language models to extract a single key frame. This key frame then acts as a prompt for a pre-trained vision encoder. Interestingly, their results show that this approach can achieve strong performance on video tasks like question answering and retrieval, suggesting that some video understanding can be achieved by analyzing a single well-chosen frame.

\subsubsection{Specialist Models}
This subsection explores the approaches using prompt-tuning to inflate IFMs for specific video tasks. 

\noindent
\textbf{Prompt Tuning for Retrieval.} In text-video retrieval, \textit{VoP} \cite{Huang-CVPR-2023} proposes an efficient adaptation method with minimal parameters by introducing lightweight text and video prompts. These prompts guide the model towards relevant information while retaining zero-shot capabilities.

\noindent
\textbf{Prompt Tuning for Video Action Recognition.}
\textit{Vita-CLIP} \cite{Wasim-CVPR-2023} freezes the pre-trained backbone to retain zero-shot capabilities and introduces learnable prompts on both vision and text sides. These prompts capture video-specific information, enhancing representation capabilities.

\noindent
\textbf{Prompt Tuning for VQA.} 
\textit{Q-ViD} \cite{Romeroa-arxiv-2024} provides instruction prompts to \textit{InstructBLIP} \cite{Instructblip-arxiv-2023} for generating video captions, improving VQA. \textit{ViTiS} \cite{Engin-ICCVW-2023} addresses the challenge of adapting large, pre-trained vision-language models to VideoQA under limited data, preserving generalization and minimizing parameters while empowering efficient task-specific fine-tuning with multimodal prompt learning and a visual mapping network.

\subsection{Inflation With Hybrid Approaches}

Some approaches either adapt \cite{Chen-arxiv-2023, buch2022revisiting} or involve combinations \cite{VifiClip-CVPR-2023, Wang-ACMMM-2023a} of post-pretraining, adapters, and prompt-tuning to inflate IFMs into ViFMs. We discuss such approaches in this subsection. 

\subsubsection{Generalist Models}
Some methods explore a combination of different inflating approaches for generic video understanding. \textit{PromptCLIP} \cite{PromptClip-ECCV-2022} efficiently adapts pre-trained models by learning prompt vectors instead of handcrafted ones. These capture video information and act as virtual tokens within the text encoder. PromptCLIp further introduces lightweight transformers (adapter) in the network to capture temporal dynamics.
\textit{Singularity} \cite{Lei-arxiv-2022} consolidates video clips into single frames using a few additional attention layers, it further modifies the prompts in order to model the temporal relationships exist in videos. 
\textit{COSA} \cite{Chen-arxiv-2023} further supports this notion by constructing "pseudo long-form videos" from existing image-text data. This approach randomly combines multiple images, creating a sequence of static scenes with richer contexts and detailed captions. These "pseudo videos" allow powerful IFMs, primarily trained on images, to be repurposed for video tasks without explicit temporal modeling.

\subsubsection{Specialist Models}

We discuss approaches that combine inflating techniques (detailed in Figure~\ref{fig:image_based_vifms}) and methods that leverage visual prompting by adapting pre-trained \textit{Segment Anything Model (SAM)} \cite{Sam-ICCV-2023} for specific video tasks. 

\noindent
\textbf{Inflating IFMs for Text-To-Video Retrieval.} CenterCLIP \cite{Zhao-SIGIR-2022}, X-Pool \cite{Gorti-CVPR-2022}, and \textit{MEME} \cite{kang2023meme} presents \textit{clustering-based} approaches for the retrieval task. These approaches bridge the semantic gap between textual queries and video content by grouping similar data points together. \textit{CenterCLIP} \cite{Zhao-SIGIR-2022} identifies the most representative token, \textit{X-Pool} \cite{Gorti-CVPR-2022}  uses text as a condition to guide the aggregation of video tokens, and \textit{MEME} \cite{kang2023meme} proposes graph patch spreading (GPS) to cluster similar patches together. \textit{ProST \etal}~\cite{Li-ICCV-2023b} improves the retrieval performance by focusing on fine-grained visual objects (spatial) and interaction (temporal) among them during video-text pretraining and thus inflate IFMs for videos. 

\noindent
\textbf{Inflating IFMs Video Action Recognition.} IFMs are used as basis for multiple video action recognition approaches including: ViFi-CLIP \cite{VifiClip-CVPR-2023}, X-CLIP \cite{Ni-ECCV-2022a}, Action-CLIP \cite{Actionclip-arxiv-2021}, Wang \etal~\cite{Wang-ACMMM-2023a}, BIKE \cite{Wu-CVPR-2023}. \textit{ViFi-CLIP} \cite{VifiClip-CVPR-2023} first fine-tune CLIP \cite{Clip-ICML-2021} on video data and thus implicitly captures temporal cues without additional modules. It further enhances the performance by learning prompts using a \textit{bridge and prompt} approach in low-data settings. \textit{Wang \etal}~\cite{Wang-ACMMM-2023a} captures motion cues through a two-stream adapter block, enriching video representations without sacrificing CLIP's generalization. Additionally, it generates dynamic, motion-aware prompts that describe actions more effectively, guided by captured motion cues. Finally, a pre-matching step aligns video and text representations before feeding them to CLIP, further boost performance. \textit{BIKE} \cite{Wu-CVPR-2023} utilizes a bidirectional knowledge exploration framework (T2V and V2T) from pre-trained IFMs and improves the representation for video recognition.

\noindent
\textbf{Adapting SAM \cite{SegmentAnything-ICCV-2023} for Video Segmentation.}
Following the success of the SAM \cite{Sam-ICCV-2023} for image segmentation, researchers are actively adapting it for video tasks. \emph{SAM-Track} \cite{Cheng-arxiv-2023} empowers users to interactively segment and track objects through clicks, strokes, or text, while \emph{TAM} \cite{Yang-arxiv-2023a} achieves high-performance interactive video tracking and segmentation with minimal clicks. For multi-object scenarios, \emph{HQTrack} \cite{Zhu-arxiv-2023b} utilizes SAM for segmentation followed by mask refinement, and \emph{DEVA} \cite{Deva-ICCV-2023} incorporates temporal coherency into per-frame SAM segmentation for improved consistency. In the unsupervised realm, \emph{UVOSAM} \cite{Zhang-arxiv-2023b} leverages SAM for video object segmentation without costly annotations. Additionally, \emph{RefSAM} \cite{Li-arxiv-2023d}  refines SAM for referring video object segmentation (RVOS) by using multi-view information (text, different frames), and \emph{SAM-PT} \cite{Rajič-arxiv-2023} employs point selection and propagation for zero-shot object tracking and segmentation. These diverse adaptations showcase SAM's potential for various video segmentation tasks, including interactive experiences, leveraging temporal information, and unsupervised learning, marking significant progress in the field.


\begin{table*}
\centering
\caption{Table presents the ViFMs using video-based pretraining. These methods are divided into three categories based on the pretraining objective as: generative, discriminative, and hybrid (combining both generative and discriminative). The hyperlink in the approach points to the corresponding implementation.}
\footnotesize
\resizebox{1\textwidth}{!}{%
\begin{tabular}{|l|x{2.8cm}x{1.4cm}|x{1.6cm}x{2cm}|lx{3.0cm}|c|}
\toprule
 \multirow{2}{*}{Method} & \multicolumn{2}{|c}{\underline{Pretraining data}} & \multicolumn{2}{|c}{\underline{Pretraining Objectives}} & \multicolumn{2}{|c|}{\underline{Architecture}} & \multirow{2}{*}{Venue} \\ 
 & Dataset(s) & Size & Contrastive & Generative & Type & Base &  \\ 
\midrule

\href{https://github.com/xyzforever/BEVT}{Bevt} \cite{Wang-CVPR-2022} & IN-1K \cite{ImageNet-CVPR-2009}, K400 \cite{K400-CVPR-2017} & 400M & - & MIM, MVM & ED & Video-Swin \cite{VideoSwin-CVPR-2022}, VQ-VAE \cite{Vqvae-NeurIPS-2017} & CVPR'22 \\

\href{https://github.com/facebookresearch/mae_st}{ST-MAE} \cite{Feichtenhofer-NeurIPS-2022} &  IN-1K \cite{ImageNet-CVPR-2009}, K400 \cite{Kinetics400-arxiv-2017}, K600 \cite{Kinetics600-arxiv-2018}, K700 \cite{Kinetics700-arxiv-2019} & 710M & - & MVM & ED & ViT \cite{Vit-arxiv-2020} & NeurIPS'22 \\

\href{https://github.com/MCG-NJU/VideoMAE}{VideoMAE} \cite{Tong-NeurIPS-2022} &  K400 \cite{K400-CVPR-2017}, SSv2 \cite{SomethingSomethingv2-ICCV-2017} & 400M & - & MVM & ED & ViT \cite{Vit-arxiv-2020} & NeurIPS'22 \\

MAM$^2$ \cite{Song-arxiv-2022} &  K400 \cite{K400-CVPR-2017} & 400M & - & MVM & ED & ViT \cite{Vit-arxiv-2020}, VQ-VAE \cite{Vqvae-NeurIPS-2017} & arxiv'22 \\

\href{https://github.com/MCG-NJU/MGMAE}{MG-MAE} \cite{Mgmae-ICCV-2023} &  SSv2 \cite{SomethingSomethingv2-ICCV-2017} & 400M & - & MVM & ED & ViT \cite{Vit-arxiv-2020} & ICCV'23 \\

AudVis MAE \cite{Georgescu-ICCV-2023} & VGG Sound \cite{Vggsound-ICASSP-2020} & 0.2M & - & MVM & ED & MAE \cite{MAE-CVPR-2022} & ICCV'23 \\

\href{https://github.com/microsoft/LAVENDER}{LAVANDER} \cite{Li-CVPR-2023}  & WebVid-2M \cite{WebVid2M-CVPR-2021}, CC3M \cite{CC3M-ACL-2018} & 5M & - & MLM & JE & Video-Swin \cite{VideoSwin-CVPR-2022}, BERT-base \cite{Bert-NAACL-2019} & CVPR'23 \\

MMVG \cite{Fu-CVPR-2023b} & EpicKitchen \cite{EpicKitchen-ECCV-2018}, Flintstones \cite{Flintstones-ECCV-2018}, Mugen \cite{Mugen-ECCV-2022} & 0.5M & - & TVC  & JE & VQ-VAE \cite{Vqvae-NeurIPS-2017}, CLIP-Tokenizer \cite{Clip-ICML-2021}, VideoSWIN \cite{VideoSwin-CVPR-2022} & CVPR'23 \\

\href{https://github.com/ruiwang2021/mvd}{MVD} \cite{Mvd-CVPR-2023} & K400 \cite{K400-CVPR-2017} & 400M & - & MFM & DD & ViT \cite{Vit-arxiv-2020} & CVPR'23 \\

\href{https://github.com/OpenGVLab/VideoMAEv2}{VideoMAEv2} \cite{VideoMAEv2-CVPR-2023a} &  Unlabeled Hybrid & 135M & - & MVM & ED & ViT \cite{Vit-arxiv-2020} & CVPR'23 \\

\href{https://github.com/ali-vilab/videocomposer}{VideoComposer} \cite{Videocomposer-NeurIPS-2023} & WebVid-10M \cite{WebVid2M-CVPR-2021}, LAION-400M \cite{Laion-arxiv-2021} & 410M & - & GVM & ED & VLDM \cite{Denoising-NeurIPS-2020, Sohl_Dickstein-ICML-2015}, CLIP-ViT-H \cite{Clip-ICML-2021} & NeurIPS'23 \\

MATS \cite{Mats-arxiv-2023} &  K400 \cite{K400-CVPR-2017}, SSv2 \cite{SomethingSomethingv2-ICCV-2017}, UCF101 \cite{Ucf101-arxiv-2012}, HDMB51 \cite{Jhmdb-ICCV-2011}, Ego4D \cite{Ego4D-CVPR-2022}& $>>$400M & - & MIM, MVM & ED & ViT \cite{Vit-arxiv-2020} & arxiv'23 \\

VideoBERT \cite{Sun-ICCV-2019} & Web scraping & 300k & - & MLM, MVM & JE & BERT \cite{Bert-NAACL-2019}, Transformer \cite{Vaswani-NeurIPS-2017} & ICCV'19 \\
\midrule
\href{https://github.com/facebookresearch/HierVL}{HierVL} \cite{Ashutosh-CVPR-2023} & Ego4D \cite{Ego4D-CVPR-2022} & 3M & VTC & - & Mul-E & Frozen \cite{WebVid2M-CVPR-2021}, DistillBERT \cite{Distilbert-arxiv-2019} & arxiv'19 \\

\href{https://github.com/bpiyush/TestOfTime}{TACT} \cite{Bagad-ICLRW-2023} & Synthetic Dataset & 180M & TOCC, CT & - & - & VideoCLIP & ICLRW'23 \\

\href{https://github.com/google-research/scenic/tree/main/scenic/projects/verbs_in_action}{VFC} \cite{Momeni-ICCV-2023} & SMiT \cite{Monfort-CVPR-2021} & 0.5M & VTC, VFC & - & DE & PaLM \cite{Palm-JMLR-2023}, CLIP-ViT \cite{Clip-ICML-2021} & ICCV'23 \\

\href{https://github.com/openai/gpt-2}{LAVILA} \cite{Zhao-CVPR-2023} & Ego4D \cite{Ego4D-CVPR-2022}, HowTo100M \cite{Howto100m-ICCV-2019} & 141M & VTC & - & DE & GPT-2 \cite{GPT2-2019} & CVPR'23 \\

\href{https://github.com/MikeWangWZHL/Paxion.git}{PAXION} \cite{Paxion-NeurIPS-2023} & ActionBench & 0.4M & VTC, DVDM & -  & DE & InternVideo, CLIP-ViP, Singularity-temporal & arxiv'23 \\

\href{https://github.com/OpenGVLab/InternVideo/tree/main/Data/InternVid}{ViCLIP} \cite{InternVid-arxiv-2023} & InterVid \cite{InternVid-arxiv-2023} & 234M & VTC & - & DE & Vit-L \cite{Vit-arxiv-2020} & arxiv'23 \\

\href{https://github.com/facebookresearch/fairseq/tree/main/examples/MMPT}{VideoCLIP} \cite{Videoclip-EMNLP-2021}  & HowTo100M \cite{Howto100m-ICCV-2019} & 136M & VTC & - & DE & Vit-L \cite{Vit-arxiv-2020}, Transfomer \cite{Vaswani-NeurIPS-2017} & EMNLP'21 \\
\midrule

\href{https://github.com/microsoft/UniVL}{UniVL} \cite{Luo-arxiv-2020} & HowTo100M \cite{Howto100m-ICCV-2019} & 136M & VTJ, VTA & CMFM, CMLM, LM  & ME & BERT \cite{Bert-NAACL-2019}, Transformer \cite{Vaswani-NeurIPS-2017} & arxiv'20 \\

\href{https://github.com/salesforce/ALPRO}{ALPRO} \cite{Alpro-CVPR-2022} & WebVid-2M \cite{WebVid2M-CVPR-2021}, CC3M \cite{CC3M-ACL-2018} & 5M & VTC, VTM, PEM & MLM & ME & TimeSformer \cite{Timesformer-ICML-2021} & CVPR'22 \\

\href{https://github.com/microsoft/XPretrain/tree/main/hd-vila}{HD-VILA} \cite{Hdvila-CVPR-2022} & HD-VILA-100M \cite{Hdvila-CVPR-2022} & 103M & VTC & MLM & ME & Bert \cite{Bert-NAACL-2019} & CVPR'22 \\

\href{https://github.com/microsoft/XPretrain}{LF-VILA} \cite{Sun-NeurIPS-2022} & LF-VILA-8M \cite{Sun-NeurIPS-2022} & 8M & VTC, VTM, MTC & MLM  & ME & Transformer \cite{Vaswani-NeurIPS-2017} & NeurIPS'22 \\

\href{https://github.com/zinengtang/TVLT}{TVLT} \cite{Tang-NeurIPS-2022} & HowTo100M \cite{Howto100m-ICCV-2019}, YTTemporal180M \cite{Yttemporal-NeurIPS-2021} & 316M & VAM & MSM  & JE & MAE \cite{MAE-CVPR-2022} & NeurIPS'22 \\

\href{https://github.com/airsplay/vimpac}{Vimpac} \cite{Vimpac-arxiv-2021} &  HowTo100M \cite{Howto100m-ICCV-2019} & 136M & VCC & MTP & ED & BERT \cite{Bert-NAACL-2019}, SimCLR \cite{Simclr-ICLR-2020} & arxiv'22 \\

\href{https://github.com/mayuelala/SimVTP}{SimVTP} \cite{Ma-arxiv-2022} & WebVid-2M \cite{WebVid2M-CVPR-2021} & 2M & VTC, VTM & MSM  & ED & BERT \cite{Bert-NAACL-2019}, VideoMAE \cite{Tong-NeurIPS-2022} & arxiv'22 \\

\href{https://github.com/tsujuifu/pytorch_violet}{Violet} \cite{Violet-arxiv-2022} & YT-Temporal \cite{Yttemporal-NeurIPS-2021}, WebVid-2.5M \cite{WebVid2M-CVPR-2021}, CC-3M \cite{CC3M-ACL-2018} & 186M & VTM & MLM, MVM & ME & Video-Swin \cite{VideoSwin-CVPR-2022}, LE \cite{Zhang-arxiv-2016}, VQ-VAE \cite{Vqvae-NeurIPS-2017} & arxiv'22 \\

\href{https://github.com/showlab/all-in-one}{All-in-One} \cite{Wang-CVPR-2023} & HowTo100M \cite{Howto100m-ICCV-2019}, CC3M \cite{CC3M-ACL-2018}, WebVid-2.5M \cite{WebVid2M-CVPR-2021}  & 110M & VTM & MLM & JE & ViT \cite{Vaswani-NeurIPS-2017} & CVPR'23 \\

Hitea \cite{Hitea-CVPR-2023}  & WebVid-2M \cite{WebVid2M-CVPR-2021}, CC3M \cite{CC3M-ACL-2018} & 5M & VTC, VTM, MTRE, CME & MLM, PrefixLM & ME & MVit-Base \cite{Mvit-CVPR-2022}, BERT-Base \cite{Bert-NAACL-2019} & CVPR'23 \\

\href{https://github.com/LeeYN-43/Clover}{Clover} \cite{Clover-CVPR-2023} & WebVid-2M \cite{WebVid2M-CVPR-2021}, CC3M \cite{CC3M-ACL-2018} & 5M & TMA & MLM, MVM & ME &  Video-Swin \cite{VideoSwin-CVPR-2022}, BERT \cite{Bert-NAACL-2019} & CVPR'23 \\

\href{https://github.com/klauscc/VindLU}{VindLu} \cite{Vindlu-CVPR-2023} & WebVid-10M \cite{WebVid2M-CVPR-2021}, CC3M \cite{CC3M-ACL-2018}, CC12M \cite{Conceptual12m-CVPR-2021} & 25M & VTC, VTM & MLM, MVM & DE &  ViT \cite{Vit-arxiv-2020}, BERT \cite{Bert-NAACL-2019} & CVPR'23 \\

\href{https://github.com/tsujuifu/pytorch_empirical-mvm}{VioletV2} \cite{Violetv2-CVPR-2023a} & WebVid-2M \cite{WebVid2M-CVPR-2021}, CC3M \cite{CC3M-ACL-2018} & 5M & VTM & MLM, MVM & ME & Video-Swin \cite{VideoSwin-CVPR-2022}, LE \cite{Zhang-arxiv-2016}, VQ-VAE \cite{Vqvae-NeurIPS-2017}, DPT-L \cite{Dpt-ICCV-2021}, RAFT-L \cite{Raft-ECCV-2020}, SWIN-B \cite{Swin-ICCV-2021}, DALL-E \cite{Dalle-ICML-2021}, CLIP-Vit-B \cite{Clip-ICML-2021} & CVPR'23 \\
 
MuLTI \cite{Xu-arxiv-2023} &  WebVid-2M \cite{WebVid2M-CVPR-2021}, CC-3M \cite{CC3M-ACL-2018} & 5M &  VTM, VTC, MCM & MLM & ME & ViT \cite{Vit-arxiv-2020}, BERT \cite{Bert-NAACL-2019} & arxiv'23 \\

\bottomrule
\end{tabular}
}%
\label{tab:video_based}
\end{table*}

\section{Video-based Models}
\label{sec:video_based_vifm}

Video-based models, trained on datasets containing videos, seek to generalize across various video understanding tasks. We classify these models into three primary categories based on their pretraining objectives: generative, discriminative, and hybrid. These categories are detailed in Table \ref{tab:video_based}.

\subsection{Generative Pretraining Objective}

In this category we consider the foundation models with a generative objective, often mask-reconstruction. The generative objective encompasses MVM, MFM, MIM, and GVM. In mask modeling, applied masking schemes have evolved over time, transitioning from discrete token masking to random token masking, with some exploring even intelligent masking strategies. We classify approaches based their masking-scheme in this section. 

\noindent
\textbf{Discrete Token Masking.}
Earlier approaches \cite{Wang-CVPR-2022, Song-arxiv-2022, Vimpac-arxiv-2021} in masked video modeling are based on the prediction of discrete tokens, where each discrete token corresponds to a visual cube from a video. These discrete tokens are generated using dVAE in VQGAN \cite{VQGAN-CVPR-2021}. \textit{Bevt} \cite{Wang-CVPR-2022} jointly reconstructs discrete visual tokens within the image and video domains, facilitating the separation of spatial and temporal modeling. \textit{MAM$^2$} \cite{Song-arxiv-2022} propose an encoder-regressor-decoder network followed by two separate decoders to disentangle spatiotemporal modeling. The spatial and temporal decoders in this case reconstruct discrete mask tokens, and RGB difference respectively. 

\noindent
\textbf{Random Masking.}
Due to the limitations imposed by the size of the visual codebook, these methods have been replaced by simpler approaches that directly reconstruct masked visual patches. \textit{ST-MAE} \cite{Feichtenhofer-NeurIPS-2022} extends the concept of MAE \cite{MAE-CVPR-2022} (for the image domain) to videos, where they propose reconstruction by randomly masking 90\% of space-time patches as a challenging pretext task for videos. Different from that, \textit{VideoMAE} \cite{Tong-NeurIPS-2022} considers time as a third independent dimension and proposes masking cubes instead of space-time patches. ST-MAE \cite{Feichtenhofer-NeurIPS-2022} also observe that randomly masking 90\% of video cubes results in effective representation learning.  Building upon the VideoMAE \cite{Tong-NeurIPS-2022} framework, \textit{VideoMAEv2} \cite{VideoMAEv2-CVPR-2023a} introduces a dual masking strategy that effectively removes cubes from both the encoder and decoder networks, significantly enhancing the model's performance. Additionally, VideoMAEv2 \cite{VideoMAEv2-CVPR-2023a} expands its capabilities by incorporating data from multiple sources, further increasing its scale and pre-training efficiency. These approaches incorporate random masking, which might not always result in an optimal representation that can generalize across multiple tasks. 

\noindent
\textbf{Intelligent Masking Schemes.}
Some approaches \cite{Mats-arxiv-2023, Mgmae-ICCV-2023} propose intelligent masking schemes resulting in an effective representation and reducing the computational complexity of the model.  \textit{MATS} \cite{Mats-arxiv-2023} introduces motion-aware token selection using a pair of adjacent frames. Additionally, this approach introduces motion-aware adaptive frame sampling to further reduce computational complexity. \textit{MGMAE} \cite{Huang-CVPR-2023} introduce motion information while masking using optical flow and thus propose to generate temporally consistent masking/visible volume. Approaches like MVD \cite{Mvd-CVPR-2023} propose improving the representation by predicting the feature maps instead of raw pixel values. \textit{MVD} \etal~\cite{Mvd-CVPR-2023} propose a dual decoder architecture for efficient spatiotemporal modelling, where one decoder predicts the features of a pre-trained image backbone, and the second decoder predicts the features of a pretrained video backbone. 

\noindent
\textbf{Large Multi-modal Modals (LMMs).} 
Several recent works explore video understanding and generation using large language models (LLMs). \textit{ChatVideo} \cite{Chatvideo-arxiv-2023} and \textit{MM-VID} \cite{Mmvid-arxiv-2023} convert videos into text for improved comprehension. \textit{VideoChatGPT} \cite{Video_Chatgpt-arxiv-2023} and \textit{VideoChat} \cite{VideoChat-arxiv-2023} enhance video-based conversations by integrating visual encoders with LLMs and instruction tuning. \textit{Valley} \cite{Valley-arxiv-2023} creates video assistants using curated instruction datasets and a projection module. \textit{PaLM2-VAdapter} \cite{Palm2_VAadapter-arxiv-2024} progressively aligns vision and language features using a vision-language adapter module. \textit{VideoDirectorGPT} \cite{VideoDirectorGPT-arxiv-2023} demonstrates LLMs' potential in video generation tasks with a unique framework. It employs LLMs to plan video content and guide scene-specific video generation, showcasing the versatility of LLMs in both video understanding and creation.

So far, we discuss unimodal approaches and LMMs in this category. \textit{LAVENDER} \cite{Li-ICCV-2023} and \textit{AudVis MAE} \cite{Georgescu-ICCV-2023} are two multimodal approaches based on generative pretraining objective. \textit{LAVENDER} \cite{Li-ICCV-2023} employs text as additional modality and MLM as training objective, whereas \textit{AudVis MAE} \cite{Georgescu-ICCV-2023} proposed unified encoding of audio-visual modalities. 



\subsection{Discriminative Pretraining Objective}

Multi-modal contrastive large-scale pretraining has emerged as a dominant trend, surpassing mask reconstruction approaches due to its superior ability to generalize models across different domains. This is because multi-modality incorporates information from multiple sources, such as text and vision, leading to richer and more versatile representations. Research \cite{Clip-ICML-2021, Videoclip-EMNLP-2021} shows that text is the most frequent modality used in conjunction with the visual domain for multi-modal contrastive pretraining. These models are often trained using VTA (discriminative) and VTC (discriminative) as a common pretraining objective. We discuss the research using different variants of discriminative objectives in this section. 

\noindent
\textbf{Simple Approaches.} Approaches like \textit{ViClip} \cite{InternVid-arxiv-2023} and \textit{VideoClip} \cite{Videoclip-EMNLP-2021}, aim to create a video counterpart to CLIP \cite{Clip-ICML-2021} (Image-Text Contrastive). These methods rely on video-text contrastive learning as their primary objective. Notably, ViClip \cite{InternVid-arxiv-2023} further validates the impact of large and diverse training datasets on the quality of learned representations. However, video data poses a greater challenge compared to simpler image-text pretraining due to the additional temporal dimension inherent in video. 

\noindent
\textbf{Introducing Temporal Consistency and Action Understanding.}
While simple video-text pretraining struggles with capturing the flow of time in videos, ViFMs like \textit{TACT} \cite{Bagad-ICLRW-2023}, \textit{PAXION} \cite{Paxion-NeurIPS-2023}, \textit{HierVL} \cite{Ashutosh-CVPR-2023} and \textit{VFC} \cite{Momeni-ICCV-2023} offer promising solutions. \textit{TACT} \cite{Bagad-ICLRW-2023} and \textit{PAXION} \cite{Paxion-NeurIPS-2023} focus on improving temporal understanding, with TACT enforcing correct event order and PAXION leveraging a knowledge patcher and a specific objective. \textit{HierVL \cite{Ashutosh-CVPR-2023}}, on the other hand, aims for comprehensive understanding by analyzing videos at different scales and summarizing both short clips and entire videos. Finally, Verb-focused Contrastive (VFC) \cite{Momeni-ICCV-2023} excels at capturing fine-grained action details through challenging contrastive examples and precise verb alignment.

\noindent
\textbf{Long-form Video Understanding.}
Long-form video understanding presents challenges due to the memory requirements and model capacities, with only a few attempts extending existing models for this purpose. \textit{LaViLa} \cite{Zhao-CVPR-2023} investigates how pre-trained Large Language Models (LLMs) can be utilized. This method turns LLMs into "Narrators" by giving them visual inputs, enabling them to automatically create detailed descriptions of long videos. These descriptions are then used to train a video-language model. In a similar direction, \textit{MovieChat} \cite{Song-arxiv-2023} combines ViFMs with LLMs using a Q-former and a projection layer. MovieChat tackles the challenge of processing lengthy videos by introducing a memory management mechanism that reduces complexity and cost while enhancing comprehension.

\subsection{Hybrid Pretraining Objective}

Hybrid pre-training combines generative tasks (like mask reconstruction) with discriminative tasks (like contrastive loss). By integrating both generative and discriminative objectives, these hybrid approaches aim to enhance the learned representations. We'll delve into an overview of such methodologies in this section.

\noindent
\textbf{Simple Approaches.} \textit{VIMPAC} \cite{Vimpac-arxiv-2021} is a basic uni-modal approach. It combines a generative task (mask reconstruction) with a contrastive objective (VTC). During contrastive learning, clips from the same video are considered positive pairs, while clips from different videos are considered negative. Conversely, \textit{VideoBERT} \cite{Sun-ICCV-2019} represents another straightforward approach, leveraging the robust BERT \cite{Bert-NAACL-2019} architecture to accommodate the temporal characteristics of video data. 

\noindent
\textbf{Advanced Approaches.} \textit{UniVL} \cite{Luo-arxiv-2020}, and \textit{Clover},  \cite{Clover-CVPR-2023} takes a more advanced approach. Instead of naively combining objectives, \textit{UniVL} employs stage-by-stage pre-training for both discriminative and generative tasks. Additionally, UniVL also innovates in masking by applying a 15\% probability of masking the entire text during video generation. These advancements allow UniVL to excel in both video understanding and generation tasks. On the other hand, \textit{Clover} employs an additional pre-training objective called Tri-Modal Alignment (TMA) to improve cross-modal understanding. It further leverages "pair-wise ranking loss" to ensure fine-grained discriminative ability. 

Recent advancements extend ViFMs beyond basic video understanding tasks. For example, \textit{MMVG} \cite{Fu-CVPR-2023b} tackles video storytelling by generating stories from textual prompts. It achieves this by treating videos as sequences of tokens, training the model on text-video relationships through an additional Text-to-Video Completion (TVC) objective. \textit{HD-VILA} \cite{Hdvila-CVPR-2022} focuses on versatility by leveraging a diversely sourced dataset (HD-VILA-100M) for pre-training, enhancing performance across different tasks. Finally, \textit{TVLT} \cite{Tang-NeurIPS-2022} explores understanding multimedia content by focusing solely on raw video and audio, eliminating the need for language.

\noindent
\textbf{Improving Action Understanding and Temporal Reasoning.}
Building on the limitations of basic video-text contrastive pre-training, recent advancements with the hybrid pretraining objective also strive to improve temporal understanding in ViFMs. Approaches like \textit{ALPRO} \cite{Alpro-CVPR-2022} combine contrastive loss with specialized techniques (e.g., Prompting Entity Modeling (PEM)) for finer-grained video analysis. \textit{Hitea} \cite{Hitea-CVPR-2023} delves deeper, capturing details of individual moments and their connection to text descriptions through methods like Cross-Modal Moment Exploration (CME). \textit{LF-VILA}  \cite{Sun-NeurIPS-2022} tackles long-range dependencies and temporal relationships across modalities with its Multimodal Temporal Contrastive (MTC) and Hierarchical Temporal Window Attention (HTWA) mechanisms. These efforts showcase the ongoing push to strengthen ViFMs' ability to grasp the flow of time within videos and extract valuable action knowledge.

\vspace{0.1cm}
\noindent
\textbf{Efficient-Effective Approaches.}
Video Foundation Models (ViFMs) face a trade-off between efficiency and performance. ViFMs like \textit{VIOLET} \cite{Violet-arxiv-2022} and \textit{VIOLETv2} \cite{Violetv2-CVPR-2023a} prioritize complex end-to-end transformer models (e.g., VideoSWIN \cite{VideoSwin-CVPR-2022}) for handling spatiotemporal dynamics, leading to higher computational cost. While \textit{VIOLET} employs discrete token modeling as its pre-training objective, \textit{VIOLETv2} expands upon this by incorporating eight additional pre-text tasks, increasing computational overhead.

To address this challenge, approaches like \textit{All-in-One} \cite{Wang-CVPR-2023}, \textit{SimVTP} \cite{Ma-arxiv-2022} and \textit{MuLTI} \cite{Xu-arxiv-2023} address this trade-off with different strategies. \textit{All-in-One} proposes a single, streamlined model that can simultaneously process raw video pixels and text tokens, eliminating separate encoders for better efficiency. It further introduces a "token rolling" operation for effective temporal encoding. \textit{SimVTP} focuses on simplicity by utilizing masked autoencoders with high masking ratios (90\% video, 75\% text) within a single encoder network. This forces the model to develop robust video-text representations for reconstruction. Finally, \textit{MuLTI} introduces a "MultiWay-Sampler" to condense textual features, enabling efficient computation. It also introduces a "Multiple Choice Modeling" pre-training task, leading to enhanced performance. These advancements exemplify the ongoing effort to create more efficient and effective VL-FMs for video-language tasks.

\vspace{0.1cm}
\noindent
\textbf{Optimizing and Evaluating ViFMs.}
In contrast to methods focused on specific aspects of pre-training, \textit{VindLU} \cite{Vindlu-CVPR-2023} offers a comprehensive roadmap for effective VL-FM pre-training.  This work delves into architecture design, fusion techniques, pre-training objectives, data selection, training protocols, and scaling strategies, providing a valuable guide for researchers developing future VL-FMs.
Furthermore, \textit{MELTR} \cite{Ko-CVPR-2023} presents a methodology for fine-tuning VL-FMs to enhance their generalizability across diverse downstream tasks. This approach addresses the challenge of adapting pre-trained models to new applications.
\textit{VideoGLUE} \cite{VideoGLUE-arXiv-2023} establishes a standardized evaluation protocol for Video Foundation Models (ViFMs). This benchmark dataset and suite of tasks enables researchers to compare the performance of different ViFMs.
Finally, \textit{Video-Bench} \cite{Video_Bench-arxiv-2023} provides a comprehensive benchmark and toolkit, which aims at evaluating the true potential of Video-LMMs towards achieving human-like comprehension and decision-making.

\begin{table*}
\centering
\caption{Table presents the Universal Foundational Models (UFM) combining multiple modalities other than video and text. These methods are divided into three categories based on the pretraining objective as: generative, discriminative, and hybrid (combining both generative and discriminative). The hyperlink in the approach points to the corresponding implementation.}
\footnotesize
\resizebox{1\textwidth}{!}{%
\begin{tabular}{|l|x{3.6cm}x{1.0cm}|x{1.6cm}x{2cm}|lx{2.0cm}|r|}
\toprule
 \multirow{2}{*}{Method} & \multicolumn{2}{|c}{\underline{Pretraining data}} & \multicolumn{2}{|c}{\underline{Pretraining Objectives}} & \multicolumn{2}{|c|}{\underline{Architecture}} & \multirow{2}{*}{Venue} \\ 
 & Dataset(s) & Size & Contrastive & Generative & Type & Base &  \\ 
  \midrule
\href{https://github.com/facebookresearch/fairseq/tree/main/examples/MMPT}{VLM} \cite{VLM-IJCNLP-2021} & HowTo100M \cite{Howto100m-ICCV-2019} & 1.1M & - & MFM, MLM, MMM & JE & BERT \cite{Bert-NAACL-2019} & ACL'21 \\

\href{https://github.com/facebookresearch/SlowFast.git}{MaskFeat} \cite{MaskFeat-CVPR-2022} & IN-21K \cite{ImageNet-CVPR-2009}, K400 \cite{Kinetics400-arxiv-2017} & 400M & - & MFP & ED & MViT \cite{Mvit-CVPR-2022} & CVPR'22 \\

OmniVL \cite{Omivl-NeurIPS-2022} & IN-1K \cite{ImageNet-CVPR-2009}, Something-Something V2 \cite{SomethingSomethingv2-ICCV-2017},  & 1.4M  & - & MDM & JE & BERT \cite{Bert-NAACL-2019}, TimeSformer \cite{Timesformer-ICML-2021} & NeurIPS'22 \\

\href{https://github.com/facebookresearch/omnivore/tree/main/omnimae}{OmniMAE} \cite{Omnimae-CVPR-2023} & IN-1K \cite{ImageNet-CVPR-2009}, Something-Something V2 \cite{SomethingSomethingv2-ICCV-2017},  & 1.4M  & - & MDM & JE & ViT \cite{Vit-arxiv-2020} & CVPR'23 \\

\href{https://github.com/jy0205/LaVIT}{Video-LaVIT} \cite{VideoLavit-arxiv-2024} & WebVid-10M \cite{WebVid2M-CVPR-2021}, CC3M \cite{CC3M-ACL-2018}, CC12M \cite{Conceptual12m-CVPR-2021}, SBU-Captions \cite{Sbcaptions-NeurIPS-2011}, BLIP-Capfilt \cite{Blip-ICML-2022}, RedPajama \cite{Redpajama-2023}, Instructional data \cite{Video_Chatgpt-arxiv-2023, Liu-arxiv-2023} & 103M & - & N(I/M/T)TG & JE & EVA-CLIP-ViTg \cite{Evaclip-arxiv-2023} & arxiv'24 \\

VideoPoet \cite{Videopoet-arxiv-2023} & Web Scraping & 1.25B & - & T2V, T2I, FP, Central inpainting and outpainting, AVCont & MLE & LLM \cite{Tay-ICLR-2022} & arxiv'24 \\

\midrule


\href{https://github.com/google-research/google-research/tree/master/vatt}{VATT} \cite{Akbari-NeurIPS-2021} & HowTo100M \cite{Howto100m-ICCV-2019}, AudioSet \cite{AudioSet-ICASSP-2017} & 27M & VAC, VTC & - & MLE & Tranfomers \cite{Vaswani-NeurIPS-2017} & NeurIPS'21 \\

VITO \cite{Parthasarathy-NeurIPS-2023} &  K400 \cite{K400-CVPR-2017} & 400M & AGC & - & ED & ViT \cite{Vit-arxiv-2020}, VQ-VAE \cite{Vqvae-NeurIPS-2017} & arxiv'22 \\

\href{https://github.com/FoundationVision/GLEE}{GLEE} \cite{Glee-arxiv-2023} & Object365 \cite{Objects365-ICCV-2019}, OpenImages \cite{Openimages-IJCV-2020}, COCO \cite{Coco-ECCV-2014}, LVIS \cite{Lvis-CVPR-2019}, BDD \cite{Bdd100k-CVPR-2020}, YTVIS19 \cite{Ytvis19-ICCV-2019}, YTVIS21 \cite{Ytvis21-2021}, OVIS \cite{Ovis-IJCV-2022}, RefCOCO \cite{RefCoco-ECCV-2016}, RefCOCO+ \cite{RefCoco-ECCV-2016}, RefCOCOg \cite{RefCocoG-ECCV-2016}, VisualGenome \cite{Krishna-IJCV-2017}, RVOS \cite{Urvos-ECCV-2020}, SA1B \cite{SegmentAnything-ICCV-2023}, GRIT \cite{Grit-arxiv-2023} & 5M & DIS, GRS, MOT, VIS, VPS & - & MLE & ResNet-50 \cite{Resnet-TPAMI-2016}, Swin-L \cite{Swin-ICCV-2021}, EVA-02-L \cite{Eva02-arxiv-2023} & arxiv'23 \\

\href{https://github.com/PKU-YuanGroup/LanguageBind}{LanguageBind} \cite{Zhu-arxiv-2023} & VIDAL-10M & 10M & MMC & - & MLE & Open-CLIP \cite{OpenCLIP-2021} & arxiv'23 \\

\href{https://github.com/zengyan-97/X2-VLM}{X$^2$VLM} \cite{Zeng-arxiv-2023} & COCO \cite{Coco-ECCV-2014}, Visual-Gnome \cite{Krishna-IJCV-2017}, SBU Captions \cite{Sbcaptions-NeurIPS-2011}, CC \cite{CC3M-ACL-2018}, Object365 \cite{Objects365-ICCV-2019}, OpenImages \cite{Openimages-IJCV-2020}, WebVid-2.5M \cite{WebVid2M-CVPR-2021}, HowTo100M \cite{Howto100m-ICCV-2019}, Yt-Temporal \cite{Yttemporal-NeurIPS-2021} & 28M & MGA, MGL & - & ME & Transformer \cite{Vaswani-NeurIPS-2017} & arxiv'23 \\

\midrule

\href{https://github.com/rowanz/merlot}{MERLOT} \cite{Merlot-NeurIPS-2021} & YT-Temporal-180M \cite{Yttemporal-NeurIPS-2021} & 6M & FTM, TR & MLM & DE & ViT \cite{Vaswani-NeurIPS-2017}, RoBERTa \cite{Roberta-arxiv-2019} & NeurIPS'21 \\

\href{https://github.com/OpenGVLab/InternVideo}{InternVideo} \cite{Wang-arxiv-2022} & Kinetics-400 \cite{Kinetics400-arxiv-2017}, WebVid-2M \cite{WebVid2M-CVPR-2021}, WebVid-10M \cite{WebVid2M-CVPR-2021}, HowTo100M \cite{Howto100m-ICCV-2019}, AVA \cite{Ava-CVPR-2018}, Something-Something V2 \cite{SomethingSomethingv2-ICCV-2017}, Kinetics-710 \cite{Kinetics700-arxiv-2019} & 12M & VTC & MVM & DE & ViT \cite{Vaswani-NeurIPS-2017}, UniformerV2 \cite{Uniformerv2-arxiv-2022} & arxiv'22 \\

ViC-MAE \cite{Hernandez-arxiv-2023} & MiT \cite{Moments-TPAMI-2019}, K400 \cite{K400-CVPR-2017} & 1.2M & ITC & MIM & DE & ViT \cite{Vit-arxiv-2020} & arxiv'22 \\

\href{https://github.com/zinengtang/Perceiver_VL}{Perciever-VL} \cite{Tang-WACV-2023} & CC3M \cite{CC3M-ACL-2018}, WebVid-2.5M \cite{WebVid2M-CVPR-2021} & 5M & VTM & MLM & JE & ViT \cite{Vit-arxiv-2020}, BERT \cite{Bert-NAACL-2019} & WACV'23 \\

Smaug \cite{Smaug-CVPR-2023}  & COCO \cite{Coco-ECCV-2014}, Visual-Gnome \cite{Krishna-IJCV-2017}, SBU Captions \cite{Sbcaptions-NeurIPS-2011}, CC3M \cite{CC3M-ACL-2018}, CC12M \cite{Conceptual12m-CVPR-2021}, WebVid-2M \cite{WebVid2M-CVPR-2021} & 17M & VTM, VTC & MVM, MLM & ME & ViT \cite{Vit-arxiv-2020} & CVPR'23 \\

\href{https://github.com/yuangongnd/cav-mae}{CAV-MAE} \cite{Gong-arxiv-2023} & AudioSet2M \cite{AudioSet-ICASSP-2017} & 2M & VAC & MSM & ME & ViT \cite{Vaswani-NeurIPS-2017}  & arxiv'23 \\

VAST \cite{Vast-arxiv-2023} & VAST-27M \cite{Vast-arxiv-2023} & 27M & OM-VCC, OM-VCM & OM-VCG &
 ME & BERT \cite{Bert-NAACL-2019}, BEAT \cite{BEAT-ICML-2023}, EVAClip-ViT-G \cite{Evaclip-arxiv-2023} & NeurIPS'23 \\
 
mPLUG-2 \cite{Mplug2-arxiv-2023} & MS-COCO \cite{Coco-ECCV-2014}, Visual Genome \cite{Krishna-IJCV-2017}, CC12M \cite{CC3M-ACL-2018}, SBU Captions \cite{Ordonez-NeurIPS-2011}, WebVid-2M \cite{WebVid2M-CVPR-2021}, WikiCorpus \cite{Bert-NAACL-2019}, Crawled data & 30M & VTC, VTM & MLM & ME & BERT \cite{Bert-NAACL-2019}, Transformer \cite{Vaswani-NeurIPS-2017} & arxiv'23 \\

\href{https://github.com/TXH-mercury/VALOR}{VALOR} \cite{Valor-arxiv-2023} & VALOR-1M \cite{CC3M-ACL-2018}, WebVid-2.5M \cite{WebVid2M-CVPR-2021}, CC14M \cite{CC3M-ACL-2018}, HD\_VILA\_100M \cite{Hdvila-CVPR-2022} & 119M & MGA & MGC & MLE & BERT \cite{Bert-NAACL-2019}, CLIP \cite{Clip-ICML-2021}, VideoSwin \cite{VideoSwin-CVPR-2022}, AST \cite{Ast-arxiv-2021} & arxiv'23 \\
 
InternVideo2 \cite{InternVideo2-arxiv-2024} & K-Mash \cite{InternVideo2-arxiv-2024}, MVid \cite{InternVideo2-arxiv-2024}, WebVid \cite{WebVid2M-CVPR-2021}, InternVid \cite{Internvideo-arxiv-2022} & 6B &  Con, VTC & MLM & MLE & ViT \cite{Vit-arxiv-2020}, BERT \cite{Bert-NAACL-2019}, BEATs \cite{BEAT-ICML-2023} & arxiv'24 \\

VideoPrism \cite{Videoprism-arxiv-2024} & Anonymous Corpus \cite{Videoprism-arxiv-2024} & 1.3B &  VTC & MVM, distillation & JE & ViT \cite{Vit-arxiv-2020}, ViViT \cite{AArnab-ICCV-2021} & arxiv'24 \\
\bottomrule
\end{tabular}
}%
\label{tab:unified_image_video}
\end{table*}
\section{Universal Foundational Models (UFMs)}
\label{sec:universal_fms}

Multi-modal foundation models \cite{Mplug2-arxiv-2023, Smaug-CVPR-2023, Internvideo-arxiv-2022} are aiming towards generalization by integrating additional modalities like audio and sensor data, apart from visual modalities. Table \ref{tab:unified_image_video} contains a list of such approaches. We classify these approaches based on their pretraining objective (generative, discriminative, hybrid) in this section.

\subsection{Generative Pretraining Objective}

Similar to video-based ViFMs, UFMs primarily utilize mask reconstruction as a key pretraining objective for generative pretraining. A significant trend in this domain is moving towards unifying architectures and datasets. 
\textit{VLM} \cite{VLM-IJCNLP-2021} paves the way with a single, streamlined encoder that handles both video and text input in a task-agnostic manner. This simplified architecture, fueled by innovative masking techniques like Masked Modality Model (MMM), fosters robust cross-modal understanding without sacrificing individual modality capabilities.
\textit{OmniVL} \cite{Omivl-NeurIPS-2022} takes unification a step further by proposing a single architecture for both image-language and video-language tasks. This is achieved by first unifying the pre-training datasets and then employing a single encoder for the visual (image and video) domain. Novel contrastive and masked language modeling objectives further support this approach. Consequently, OmniVL excels in both visual-only tasks like image classification and cross-modal tasks like video question answering.
\textit{OmniMAE} \cite{Omnimae-CVPR-2023} utilizes masked autoencoding to train a single encoder-decoder network for both images and videos. This approach could be easily generalized to other visual modalities such as thermal images and 3D point clouds. 
Noteworthy is OmniVORE \cite{Omnivore-CVPR-2022}, which combines various visual modalities in a similar manner but for supervised classification tasks.
A different approach, \textit{MaskFeat} \cite{MaskFeat-CVPR-2022} focuses on redefining the pretext task as masked feature prediction (MFP) to design a unified architecture for both image and video understanding, where features in this case are HOG (Histogram of Oriented Gradients). 

\vspace{0.1cm}
\noindent
\textbf{LLM-based Approaches.}
Recent advancements have integrated multiple modalities, including image, video, text, and audio, into Large Language Models (LLMs), leading to the development of general-purpose large multimodal models (LMMs). Some approaches focus on combining different visual modalities such as image and video. \textit{VideoLaVIT} \cite{VideoLavit-arxiv-2024} breaks down videos into keyframes and temporal motions, enabling unified pre-training across diverse modalities, including images and videos. Moving towards image-video understanding, \textit{VideoLLaVA} \cite{Video_Llava-arxiv-2023} aligns visual representations from images and videos before projecting them onto the LLM, refining them through instruction tuning. Additionally, \textit{Chat-UniVi} \cite{Chat_Univi-arxiv-2023} proposes a unified approach for image and video understanding using dynamic visual tokens and a multi-scale architecture to efficiently represent and perceive semantics and details simultaneously.

On the other hand, some approaches focus on integrating the audio modality along with the visual modalities into LLMs. \textit{Video-LLaMA} \cite{Video_Llama-arxiv-2023} pioneers the integration of visual and auditory information through separate Q-formers and pre-trained encoders. \textit{FAVOR} \cite{Favor-arxiv-2023} tailors a framework for audio-visual LLMs, incorporating a "Causal Q-Former" that considers causal relationships between video frames. Meanwhile, 

The trend is now slightly moving towards combining all modalities, including image, video, audio, and text, using a single foundational model. \textit{Macaw-LLM} \cite{Macaw_llm-arxiv-2023} directly integrates visual, audio, and textual features, facilitating a unified understanding of videos. Conversely, \textit{VideoPoet} \cite{Videopoet-arxiv-2023} utilizes a decoder-only transformer architecture similar to LLMs for generating high-quality videos with matching audio based on textual input, particularly excelling in "zero-shot" scenarios. Advancements continue towards audio-visual grounding, with PG-Video-LLaVa \cite{Pg_Video_Laava-arxiv-2023} enhancing LLMs for video comprehension and object grounding by introducing pixel grounding capabilities through object tracking and audio transcription.

\noindent
\textbf{Large-scale Models for Generative Tasks.}
With the advent of large-scale pretraining, two main lines of video generation approaches have emerged: autoregressive transformers \cite{Cogvideo-arxiv-2022, Nuwa-ECCV-2022, Videopoet-arxiv-2023, Videocomposer-NeurIPS-2023, VideoDirectorGPT-arxiv-2023, SORA-2024, Mora-arxiv-2024} and diffusion models \cite{MakeAVideo-arxiv-2022, Pyoco-ICCV-2023, SVD-arxiv-2023, Stablevideo-ICCV-2023, Imagen-arxiv-2023, Control-arxiv-2023, Lumiere-arxiv-2024}. Autoregressive transformers \cite{Katharopoulos-ICML-2020} generate sequences (like text or video) one element at a time, considering previously generated elements to predict the next. Some autoregressive approaches, like SORA \cite{SORA-2024}, VideoPoet \cite{Videopoet-arxiv-2023}, and VideoDirectorGPT \cite{VideoDirectorGPT-arxiv-2023}, demonstrate the ability to generalize across multiple video tasks (e.g., VideoQA, Video Generation) by incorporating autoregressive language models into their architectures. Diffusion models \cite{Denoising-NeurIPS-2020, Sohl_Dickstein-ICML-2015}, on the other hand, gradually add noise to a video sample and then learn to reverse the process to synthesize videos from this noise. Large-scale diffusion models achieve impressive results on specific video generation tasks such as video editing \cite{Dreamix-arxiv-2023, Stablevideo-ICCV-2023, Fatezero-ICCV-2023}, video synthesis \cite{Blattmann-CVPR-2023, Chen-arxiv-2023a}, and text-to-video generation \cite{MakeAVideo-arxiv-2022, SVD-arxiv-2023, Pyoco-ICCV-2023}. 

\subsection{Discriminative Pretraining Objective}

Universal models aim to achieve comprehensive understanding by processing different modalities (e.g., image, video, audio, text) together. \textit{VITO} specifically targets the fusion of image and video modalities through attention-guided contrastive learning (AGC) and harnesses a large-scale video dataset known as \textit{VideoNet}, akin to ImageNet but tailored for videos. Similarly, \textit{VATT} \cite{Akbari-NeurIPS-2021} adopts a strategy of projecting and aligning different modalities—audio, video, and text—using a cross-modal encoder to facilitate multi-modal comprehension. In contrast, \textit{$X^2VLM$} \cite{Zeng-arxiv-2023} proposes a modular architecture that offers the flexibility to integrate additional modalities seamlessly without necessitating the retraining of the entire framework. Meanwhile, \textit{LanguageBind} \cite{Zhu-arxiv-2023} addresses the expansion of modalities by leveraging language as a central anchor. It employs a pre-trained video-language model, preserving its language encoder while training new encoders for supplementary modalities like audio or depth through contrastive learning with a \textit{multi-modal contrastive (MMC)} objective. This process aligns all modalities within a shared feature space, enhancing the model's overall understanding. Notably, for object-centric video tasks, existing ViFMs may not be suitable. Recently, the introduction of GLEE \cite{Glee-arxiv-2023}, an object-centric foundation model, extends the scope of research in ViFMs for video tasks by incorporating visual prompts alongside vision-text input.

\subsection{Hybrid Pretraining Objective}

While generative pre-training objectives like masked modeling enhance the spatio-temporal understanding of videos, multi-modal contrastive learning improves the semantic understanding. Hybrid approaches aim to achieve the best of both worlds by combining these techniques during pre-training.  \textit{MERLOT} \cite{Merlot-NeurIPS-2021} exemplifies this by employing Frame-Transcript Matching (FTM) and Temporal Reordering (TR) to align video frames with their captions, alongside Masked Language Modeling (MLM) for deeper language grasp. \textit{InternVideo} \cite{Wang-arxiv-2022} takes a similar approach, leveraging Masked Video Modeling (MVM) to capture video actions and Multimodal Contrastive Learning (VLC) to create a shared semantic space for video and language. It further strengthens this representation with supervised action classification and cross-modal attention. Finally, \textit{ViC-MAE} \cite{Hernandez-arxiv-2023} utilizes Masked Autoencoders (MAEs) to capture local features in video patches for fine-grained understanding. It then employs contrastive learning and pooling across video frames to extract global features representing the entire video.

\noindent
\textbf{Efficient Approaches.}
Efficiency in pre-training video foundation models (ViFMs) is critical due to their computational expensiveness, especially with multiple or hybrid objectives. To tackle this challenge, researcher incorporates specialized attention mechanisms and strategies. For instance, \textit{Perceiver-VL} \cite{Tang-WACV-2023} utilizes iterative latent attention, a technique that bypasses the computational bottleneck of standard self-attention in transformers, leading to significant efficiency gains. Additionally, \textit{Smaug} \cite{Smaug-CVPR-2023} leverages masked autoencoders (MAEs) for efficient pre-training, masking both visual and textual inputs to reduce costs and improve cross-modal alignment. It further employs a space-time token sparsification module to strategically select informative regions and frames, minimizing computational demands.

\noindent
\textbf{Effective Approaches.}
Recent advancements have pushed the boundaries of ViFMs by incorporating multiple modalities beyond just video. \textit{VAST} \cite{Vast-arxiv-2023} pioneers an "omni-modal symphony" by incorporating vision, audio, subtitles, and text, leveraging the VAST-27M dataset for comprehensive multimodal training. Similarly, \textit{CAV-MAE} \cite{Gong-arxiv-2023} extends the MAE paradigm to video, introducing audio reconstruction during pretraining and integrating both masked modeling and contrastive objectives for enhanced comprehension. Building on prior work, \textit{InterVideo2} \cite{InternVideo2-arxiv-2024} introduces the audio modality and a progressive training approach to generalize across multiple video and audio interaction tasks. Pushing the boundaries of multimodal learning, \textit{VALOR} \cite{Valor-arxiv-2023} proposes vision-audio-language omni-perception models with discriminative and generative pretraining tasks, facilitating cross-modal connections and empowering the model for diverse tasks like retrieval and captioning. Meanwhile, \textit{VideoPrism} \cite{Videoprism-arxiv-2024} adopts a two-stage training process, refining spatio-temporal representations with video-text data and employing techniques like global-local distillation, yielding versatile representations for varied video understanding tasks. 

Modular architectures, exemplified by \textit{$X^2VLM$} \cite{Zeng-arxiv-2023} and \textit{mPLUG-2} \cite{Mplug2-arxiv-2023}, further enhance ViFM flexibility, with separate encoders for each modality and shared attention and contrastive learning modules, enabling tailored models for specific tasks and improved transferability across domains. This modular approach fosters collaboration while addressing the evolving needs of multimodal understanding in ViFMs.

\section{Results and Discussion}
\label{sec:discussion}

We compare state-of-the-art performance on six video understanding tasks: action recognition, zero-shot and open-vocabulary action recognition, text-to-video retrieval, video question answering, video captioning, and text-to-video generation (found in Tables \ref{tab:action_recognition}, \ref{tab:action_recognition_multimodal}, \ref{tab:vqa_captioning}, \ref{tab:retrieval}, and \ref{tab:generative}). Each table categorizes models as either generalist --- capable of handling multiple tasks (further subdivided into image-based, video-only, and unified image-video types, as discussed in previous sections), or specialist --- excelling at a single task. 

\begin{table}
    \centering
    \caption{Comparing the finetuned performance of state-of-the-art (SOTA) generalist and specialist models for \textbf{video action recognition} on K400 \cite{Kinetics400-arxiv-2017}, HMDB51 \cite{Jhmdb-ICCV-2011}, UCF101 \cite{Ucf101-arxiv-2012}, and SSv2 \cite{SomethingSomethingv2-ICCV-2017} datasets. The highlighted entries indicate the best performing methods in both fine-tuned and zero-shot settings.}
    \centering
    \setlength\tabcolsep{4.0pt}
    \small
    \begin{tabular}{|c|l|c|cc|cc|cc|cc|}
    \toprule
     & \multirow{2}{*}{Method} & Arch. & \multicolumn{2}{c}{K400 \cite{Kinetics400-arxiv-2017}} & \multicolumn{2}{|c|}{HMDB51 \cite{Jhmdb-ICCV-2011}} & \multicolumn{2}{c|}{UCF101 \cite{Ucf101-arxiv-2012}} & \multicolumn{2}{c|}{SSv2 \cite{SomethingSomethingv2-ICCV-2017}} \\ 
     \cline{4-11} 
     & & Type & Top-1 & Top-5 & Top-1 & Top-5 & Top-1 & Top-5 & Top-1 & Top-5 \\
    \midrule
    \multicolumn{11}{|c|}{Fine-Tune} \\
    \hline
    \multirow{16}{*}{\rotatebox{90}{Generalist}} 
    & VideoCoca \cite{Yan-arxiv-2023b} & Enc-Dec & 72.0 & 90.5 & 58.7 & 84.5 & \textbf{86.6} & \textbf{98.4} & - & - \\
    & EVL \cite{Evl-ECCV-2022} & Dual-Enc & 82.9 & - & - & - & - & - & 61.7 & - \\
    & DualPath \cite{Park-CVPR-2023} & Dual-Enc & 85.4 & 97.1  & - & - & - & - & 70.3 & 92.9 \\
    & UMT-B \cite{Li-ICCV-2023} & Dual-Enc & 87.4 & 97.5  & - & - & - & - & 70.8 & 92.6  \\
    & UMT-L \cite{Li-ICCV-2023} & Dual-Enc & 90.6 & \textbf{98.7} & - & - & - & - & 74.7 & 94.7 \\
    \cline{3-11} 
    & All-in-One \cite{Wang-CVPR-2023} & Joint-Enc & 53.2 & 83.5 & 55.2 & \textbf{89.1} & 84.1 & 95.7 & - & - \\ 
    & BEVT \cite{Wang-CVPR-2022} & Enc-Dec & 80.6 & - & - & - & - & - & 70.6 & - \\
    & MAM$^2$ \cite{Song-arxiv-2022} & Enc-R-Dec & 85.3 & 96.7 & - & - & - & - & 71.3 & 93.1 \\
    & MG-MAE \cite{Mgmae-ICCV-2023} & Enc-Dec & 81.8 & 95.0  & - & - & - & - & 72.3 & 93.5  \\
    & ST-MAE \cite{Feichtenhofer-NeurIPS-2022} & Enc-Dec & 81.3 & 94.9 & - & - & - & - & 72.1 & 93.9 \\
    & VATT \cite{Akbari-NeurIPS-2021} & Joint-Enc & 79.9 & 94.6 & - & - & - & - & - & - \\ 
    & VideoMAE-B \cite{Tong-NeurIPS-2022} & Enc-Dec & 81.5 & 95.1 & - & - & - & - & 70.8 & 92.4 \\
    & VideoMAE-H \cite{Tong-NeurIPS-2022} & Enc-Dec & 86.1 & 97.3 & - & - & - & - & 75.4 & 95.2 \\
    & VideoMAEv2-H \cite{VideoMAEv2-CVPR-2023a} & Enc-Dec & 88.6 & 97.9 & - & - & - & - & 76.8 & 95.8 \\
    & VideoMAEv2-g \cite{VideoMAEv2-CVPR-2023a} & Enc-Dec & 90.0 & 98.4 & - & - & - & - & 77.0 & \textbf{95.9} \\
    & VIMPAC \cite{Vimpac-arxiv-2021} & Enc-Only & 77.4 & - & - & - & - & - & 68.1 & - \\
    \cline{3-11}
    & InternVideo \cite{Wang-arxiv-2022} & Dual-Enc & \textbf{91.1} & - & \textbf{89.3} & - & - & - & \textbf{77.2} & - \\
    & mPLUG-2 \cite{Mplug2-arxiv-2023} & Mix-Enc & 87.1 & 97.7 & - & - & - & - & - & -  \\
    & MaskFeat-S \cite{MaskFeat-CVPR-2022} & Enc-Dec & 82.2 & 95.1 & - & - & - & - & - & - \\
    & MaskFeat-L \cite{MaskFeat-CVPR-2022} & Enc-Dec & 86.4 & 97.1 & - & - & - & - & 74.4 & 94.6 \\
    & OmniMAE \cite{Omnimae-CVPR-2023} & Joint-Enc & 84.0 & - & - & - & - & - & 69.5 & - \\
    & OmniVL \cite{Omivl-NeurIPS-2022} & Joint-Enc & 79.1 & 94.5 & - & - & - & - & - & - \\
    & VideoPrism-B \cite{Videoprism-arxiv-2024} & Joint-Enc & 84.2 & - & - & - & - & 63.6 & - & - \\
    & VideoPrism-g \cite{Videoprism-arxiv-2024} & Joint-Enc & 87.2 & - & - & - & - & - & 68.5 & - \\
    \hline
    \multirow{8}{*}{\rotatebox{90}{Specialist}} & AIM \cite{Aim-ICLR-2022} & Dual-Enc & 84.7 & 96.7  & - & - & - & - & 69.1 & 92.2 \\
    & ActionCLIP \cite{Actionclip-arxiv-2021} & Dual-Enc & 83.8 & 97.1 & - & - & - & - & - & - \\
    & BIKE \cite{Wu-CVPR-2022} & Dual-Enc & 88.6 & 98.3  & - & - & - & - & - & - \\
    & Frozen \cite{WebVid2M-CVPR-2021} & Dual-Enc & 50.5 & 80.7 & 54.3 & 88.0 & 81.3 & 94.3 & - & -\\
    & MIL-NCE \cite{Miech-CVPR-2020} & Dual-Enc & - & - & 53.1 & 87.2 & 82.7 & - & - & - \\ 
    & M$^2$-CLIP \cite{M2clip-arxiv-2024} & Dual-Enc & 84.1 & 96.8 & - & - & - & - & 69.1 & 91.8 \\
    & ST-Adapter \cite{St_adapter-NeurIPS-2022} & Dual-Enc & 82.7 & 96.2 & - & - & - & - & 69.5 & 92.6 \\
    & Vita-CLIP \cite{Vitaclip-CVPR-2023} & Dual-Enc & 82.9 & 96.3 & - & - & - & - & 48.7 & - \\
    & X-CLIP \cite{Ni-ECCV-2022a} & Dual-Enc & 87.7 & 97.4 & - & - & - & - & - & - \\
    \bottomrule
    \end{tabular}
    \label{tab:action_recognition}
\end{table}

\begin{table}
\centering
\caption{Comparing the zero-shot and base-to-novel generalization performance of SOTA ViFMs for video action recognition on K400 \cite{Kinetics400-arxiv-2017}, HMDB51 \cite{Jhmdb-ICCV-2011}, UCF101 \cite{Ucf101-arxiv-2012}, and SSv2 \cite{SomethingSomethingv2-ICCV-2017} datasets. The highlighted entries indicate the best performing methods in both zero-shot and base-to-novel generalization settings.}
\setlength\tabcolsep{7.0pt}
\small
\begin{tabular}{|l|c|c|c|c|c|c|c|}
\toprule
\multirow{2}{*}{Method} & Arch. & HMDB51 & UCF101 & K400 & HMDB51 & UCF101 & SSv2 \\ 
& Type & Top-1 & Top-1 & HM & HM & HM & HM  \\
\midrule
& & \multicolumn{2}{|c|}{Zero-Shot} & \multicolumn{4}{|c|}{Base-To-Novel}\\
\hline
ActionCLIP \cite{Actionclip-arxiv-2021} & Dual-Encoder & 40.8 & 58.3 & 52.6 & 48.5 & 70.7 & 11.5  \\
AIM \cite{Aim-ICLR-2022} & Dual-Encoder & - & - & 68.0 & 57.1 & 82.6 & 8.2 \\
BIKE \cite{Wu-CVPR-2022} & Dual-Encoder & 52.8 & 80.8 & - & - & - & -  \\
EVA-CLIP \cite{Evaclip-arxiv-2023} & Dual-Encoder  & - & 76.8 & - & - & - & - \\
EZ-CLIP \cite{Ezclip-arxiv-2023} & Dual-Encoder & 55.2 & 82.6 & 66.3 & \textbf{66.3} & 85.4 & \textbf{14.8} \\
FitCLIP \cite{Fitclip-arxiv-2022} & Dual-Encoder  & - & 73.3 & - & - & - & - \\
Froster \cite{Froster-ICLR-2024} & Dual-Encoder & - & - & \textbf{70.4} & 65.1 & \textbf{87.0} & 14.6 \\
IMP \cite{Akbari-NeurIPS-2023}  & Multi-Encoder & \textbf{59.1} & \textbf{91.5} & - & - & - & - \\
LSS \cite{Ranasinghe-NeurIPS-2023} & Dual-Encoder & 51.4 & 74.2 & - & - & - & - \\
M$^2$-CLIP \cite{M2clip-arxiv-2024} & Dual-Encoder & 47.1 & 78.7 & - & - & - & - \\
MAXI \cite{Maxi-ICCV-2023} & Dual-Encoder & 52.3 & 78.2 & - & - & - & - \\
MOV \cite{Mov-arxiv-2022} & Dual-Encoder & 57.8 & 80.9 & - & - & - & - \\
PromptCLIP \cite{Ju-ECCV-2022}  & Dual-Encoder & - & 66.6 & - & - & - & -  \\
St-Adapter \cite{St_adapter-NeurIPS-2022} & Dual-Encoder & - & - & 67.3 & 55.9 & 80.9 & 8.8 \\
ViFi-CLIP \cite{VifiClip-CVPR-2023} & Dual-Encoder & 51.3 & 76.8 & 67.9 & 61.9 & 78.3 & 13.9 \\
Vita-CLIP \cite{Vitaclip-CVPR-2023} & Dual-Encoder  & 48.6 & 75.0 & - & - & - & - \\
X-CLIP \cite{Ni-ECCV-2022a} & Dual-Encoder & 44.6 & 72.0 & 64.0 & 55.0 & 71.2 & 7.4 \\
\bottomrule
\end{tabular}
\label{tab:action_recognition_multimodal}
\end{table}

\begin{table}
\centering
\caption{Comparing the fine-tuned and zero-shot performance of state-of-the-art (SOTA) generalist and specialist models for \textbf{text-to-video retrieval} tasks on MSR-VTT \cite{Msrvtt-CVPR-2016}, DiDeMo \cite{Didemo-ICCV-2017}, and LSMDC \cite{Lsmdc-CVPR-2015} dataset. The highlighted indicate the best performing methods in both fine-tuned and zero-shot settings.}
\vspace{-0.3cm}
\setlength\tabcolsep{6.0pt}
\footnotesize
\begin{tabular}{|c|l|c|ccc| ccc| ccc|}
\toprule
 & \multirow{2}{*}{Method} & Arch. & \multicolumn{3}{c}{MSR-VTT \cite{Msrvtt-CVPR-2016}} & \multicolumn{3}{c}{DiDeMo \cite{Didemo-ICCV-2017}} & \multicolumn{3}{c|}{LSMDC \cite{Lsmdc-CVPR-2015}} \\ 
 \cline{4-12}
 & & Type & R@1 & R@5  & R@10  &  R@1 & R@5  & R@10 & R@1 & R@5  & R@10 \\
\midrule
\multicolumn{12}{|c|}{Fine-Tune} \\
\hline
\multirow{7}{*}{\rotatebox{90}{Generalist}} & PromptCLIP \cite{Ju-ECCV-2022} & Dual-Enc & 36.5 & 64.6 & - & 36.1 & 64.8 & - & 13.4 & 29.5 & - \\
& RTQ \cite{Wang-ACMMM-2023} & Dual-Enc & 53.4 & 76.1 & 84.4 & 57.6 & 84.1 & 89.8 & - & - & - \\
& Singularity \cite{Lei-arxiv-2022} & Dual-Enc & 36.8 & 65.9 & 75.5 & 47.4 & 75.2 & 84.0 & - & - & - \\
& VideoCoca \cite{Yan-arxiv-2023b} & Enc-Dec & 34.3 & 57.8 & 67.0 & - & - & - & - & - & - \\
& UMT-B \cite{Li-ICCV-2023} & Dual-Enc & 51.0 & 76.5 & 84.2 & 61.6 & 86.8 & 91.5 & 32.7 & 54.7 & 63.4 \\
& UMT-L \cite{Li-ICCV-2023} & Dual-Enc & 58.8 & 81.0 & 87.1 & 70.4 & 90.1 & 93.5 & 43.0 & 65.5 & 73.0 \\
\cline{3-12}
& All-in-One \cite{Wang-CVPR-2023} & Joint-Enc & 37.9 & 68.1 & 77.1 & 32.7 & 61.4 & 73.5 & - &- & - \\
& ALPRO \cite{Alpro-CVPR-2022} & Mix-Enc & 33.9 & 60.7 & 73.2 & 35.9 & 67.5 & 78.8 & - & - &  \\
& Clover \cite{Clover-CVPR-2023} & Mix-Enc & 40.5 & 69.8 & 79.4 & 50.1 & 76.7 & 85.6 & 24.8 & 44.0 & 54.5 \\
& HD-VILA \cite{Hdvila-CVPR-2022} & Mix-Enc & 35.6 & 65.3 & 78.0 & 28.8 & 57.4 & 69.1 & 17.4 & 34.1 & 44.1 \\
& Hitea \cite{Hitea-CVPR-2023} & Mix-Enc & 44.4 & 69.3 & 78.9 & 51.8 & 79.1 & 85.3 & 27.1 & 46.2 & 54.5 \\
& LAVANDER \cite{Li-CVPR-2023} & Joint-Enc & 37.8 & 63.8 & 75.0 & 47.4 & 74.7 & 82.4 & 22.2 & 43.8 & 53.5 \\
& SimVTP \cite{Ma-arxiv-2022} & Enc-Dec & 53.6 & 81.9 & \textbf{90.7} & - & - & - & - & - & - \\
& Smaug \cite{Smaug-CVPR-2023} & Mix-Enc & 44.0 & 70.4 & 78.8 & 55.6 & 80.8 & 88.4 & - &- & - \\
& VideoCLIP \cite{Videoclip-EMNLP-2021} & Dual-Enc & 30.9 & 55.4 & 66.8 & - & - & -  & - & - & -  \\
& VindLU-L \cite{Vindlu-CVPR-2023} & Dual-Enc & 48.8 & 72.4 & 82.2 & 59.8 & 86.6 & \textbf{91.5} & - & - & - \\
& VIOLET \cite{Violet-arxiv-2022} & Mix-Enc & 34.5 & 63.0 & 73.4 & 32.6 & 62.8 & 74.7 & 16.1 & 36.6 & 41.2 \\
\cline{3-12}
& InternVideo \cite{Wang-arxiv-2022} & Dual-Enc & 55.2 & 79.6 & 87.5 & 57.9 & 82.4 & 88.9 & 34.0 & 53.7 & 62.9 \\
& mPLUG-2 \cite{Mplug2-arxiv-2023} & Mix-Enc & 53.1 & 77.6 & 84.7 & 56.4 & 79.1 & 85.2 & \textbf{34.4} & \textbf{55.2} & \textbf{65.1} \\
& OmniVL \cite{Omivl-NeurIPS-2022} & Dual-Enc & 47.8 & 74.2 & 83.8 & 52.4 & 79.5 & 85.4 & - & - & - \\
& X$^2$VLM \cite{Zeng-arxiv-2023} & Mix-Enc & 49.6 & 76.7 & 84.2 & - & - & - & - & - & - \\
& VALOR \cite{Valor-arxiv-2023} & Multi-Enc & 54.4 & 79.8 & 87.6 & 57.6 & 83.3 & 88.8 & 31.8 & 52.8 & 62.4 \\
& VAST \cite{Vast-arxiv-2023} & Mixed-Enc & \textbf{63.9} & \textbf{84.3} & 89.6 & \textbf{72.0} & \textbf{89.0} & 91.4 & - & - & - \\
\hline
\multirow{7}{*}{\rotatebox{90}{Specialist}} & CAMoE \cite{Cheng-arxiv-2021} & Dual-Enc & 47.3 & 74.2 & 84.5 & 43.8 & 71.4 & 79.9 & 25.9 & 46.1 & 53.7 \\
& Clip4Clip \cite{Luo-Neurocomputing-2022} & Dual-Enc & 42.1 & 71.9 & 81.4 & 43.4 & 70.2 & 80.6 & 21.6 & 41.8 & 49.8 \\
& CrossTVR \cite{Dai-arxiv-2023} & Mix-Enc & 54.0 & 77.5 & 85.3 & 55.0 & 77.6 & - & 27.7 & 48.5 & - \\
& Frozen \cite{WebVid2M-CVPR-2021} & Dual-Enc & 32.5 & 61.5 & 71.2 & 31.0 & 59.8 & 72.4 & 15.0 & 30.8 & 39.8 \\
& MCQ \cite{Ge-CVPR-2022} & Dual-Enc & 37.6 & 64.8 & 75.1 & 37.0 & 62.2 & 73.9 & 17.9 & 35.4 & 44.5 \\
& MILES \cite{Miles-ECCV-2022} & Dual-Enc & 37.7 & 63.6 & 73.8 & 36.6 & 63.9 & 74.0 & 17.8 & 35.6 & 44.1 \\
& ProST \cite{Li-ICCV-2023b} & Dual-Enc & 46.9 & 73.3 & 82.9 & 47.5 & 75.5 & 84.4 & 26.3 & 46.1 & 55.2 \\
& OA-Trans \cite{Wang-CVPR-2022a} & Dual-Enc & 35.8 & 63.4 & 76.5 & 34.8 & 64.4 & 75.1 & 18.2 & 34.3 & 43.7 \\
& QB-Norm \cite{Bogolin-CVPR-2022} & - & 47.2 & 73.0 & 83.0 & 43.3 & 71.4 & 80.8 & 22.4 & 40.1 & 49.5 \\
& TMVM \cite{Lin-NeurIPS-2022} & Dual-Enc & 36.2 & 64.2 & 75.7 & 36.5 & 64.9 & 75.4 & 17.8 & 37.1 & 45.9 \\
& X-CLIP \cite{Ni-ECCV-2022a} & Dual-Enc & 46.1 & 73.0 & 83.1 & 45.2 & 74.0 & - & 23.3 & 43.0 & - \\
\hline
\multicolumn{12}{|c|}{Zero-Shot} \\
\hline
\multirow{10}{*}{\rotatebox{90}{Mix}} & MCQ \cite{Ge-CVPR-2022} & Dual-Enc & 26.0 & 46.4 & 56.4 & 25.6 & 50.6 & 61.1 & 12.2 & 25.9 & 32.2 \\
& MILES \cite{Miles-ECCV-2022} & Dual-Enc & 26.1 & 47.2 & 56.9 & 27.2 & 50.3 & 63.6 & 11.1 & 24.7 & 30.6 \\
& OA-Trans \cite{Wang-CVPR-2022a} & Dual-Enc & 23.4 & 47.5 & 55.6 & 23.5 & 50.4 & 59.8 & - & - & - \\
& UMT-B & Dual-Enc & 35.2 & 57.8 & 66.0 & 41.2 & 65.4 & 74.9 & 19.1 & 33.4 & 42.2 \\
& UMT-L & Dual-Enc & 40.7 & 63.4 & 71.8 & 48.6 & 72.9 & 79.0 & \textbf{24.9} & \textbf{41.7} & \textbf{51.8} \\ 
& FitCLIP \cite{Fitclip-arxiv-2022} & Dual-Enc & 33.8 & 59.8 & 69.4 & 28.5 & 53.7 & 64.0 & - & - & - \\
& Frozen \cite{WebVid2M-CVPR-2021} & Dual-Enc & 18.7 & 39.6 & 51.6 & 21.1 & 46.0 & 56.2 & 9.3 & 22.0 & 30.1 \\
& ALPRO \cite{Alpro-CVPR-2022} & Mix-Enc & 24.1 & 44.7 & 55.4 & 23.8 & 47.3 & 57.9 & - & - & - \\
& Clover \cite{Clover-CVPR-2023} & Mix-Enc & 26.4 & 49.5 & 60.0 & 29.5 & 55.2 & 66.3 & 14.7 & 29.2 & 38.2 \\
& VideoCLIP \cite{Videoclip-EMNLP-2021} & Dual-Enc & 10.4 & 22.2 & 30.0 & 16.6 & 46.9 & -  & - & - & -  \\
& UniVL \cite{Luo-arxiv-2020} & Mix-Enc & 21.2 & 49.6 & 63.1 & - & - & - & - & - & -  \\
& VIOLET \cite{Violet-arxiv-2022} & Mix-Enc & 25.9 & 49.5 & 59.7 & 23.5 & 49.8 & 59.8 & - & - & - \\
& InternVideo \cite{Wang-arxiv-2022} & Mix-Enc & 40.0 & 65.3 & \textbf{74.1} & 31.5 & 57.6 & 68.2 & 17.6	& 32.4 & 40.2 \\
& OmniVL \cite{Omivl-NeurIPS-2022} & Dual-Enc & 34.6 & 58.4 & 66.6 & 33.3 & 58.7 & 68.5 & - & - & - \\
& VAST \cite{Vast-arxiv-2023} & Mixed-Enc & \textbf{49.3} & \textbf{68.3} & 73.9 & \textbf{55.5} & \textbf{74.3} & \textbf{79.6}  & - & - & - \\
\bottomrule
\end{tabular}
\label{tab:retrieval}
\end{table}

\subsection{Video Content Understanding}

This section evaluates how well state-of-the-art ViFMs handle tasks involving video analysis and understanding. We compare their performance on video action recognition, multi-modal action recognition (including zero-shot and open vocabulary settings), spatio-temporal action localization, and text-to-video retrieval tasks.

\subsubsection{Video Action Recognition}
While state-of-the-art (SOTA) performance in video action recognition varies across datasets as shown in Table \ref{tab:action_recognition}, highlighting the strengths of different model types, universal foundational models (UFM) like InternVideo \cite{Wang-arxiv-2022} consistently achieve top-1 accuracy across K400 \cite{Kinetics400-arxiv-2017}, HMDB51 \cite{Jhmdb-ICCV-2011}, and SSv2 \cite{SomethingSomethingv2-ICCV-2017} datasets. On K400 \cite{Kinetics400-arxiv-2017}, it surpasses UMT-L \cite{Li-ICCV-2023} (image-based ViFM) in top-1 accuracy, while UMT-L \cite{Li-ICCV-2023} excels in top-5. Similarly, InternVideo \cite{Wang-arxiv-2022} outperforms All-in-one \cite{Wang-CVPR-2023} (video-only) in top-1 accuracy on HMDB51 \cite{Jhmdb-ICCV-2011}, while All-in-one \cite{Wang-CVPR-2023} takes the top-5 spot. Notably, VideoCoCa \cite{Yan-arxiv-2023b}(image-based ViFM) dominates both top-1 and top-5 accuracy on UCF101 \cite{Ucf101-arxiv-2012}, showcasing the diverse capabilities of various foundation models. Finally, on SSv2 \cite{SomethingSomethingv2-ICCV-2017}, InternVideo \cite{Wang-arxiv-2022} retains its top-1 dominance, while VideoMAEv2-g \cite{VideoMAEv2-CVPR-2023a} (video-only) emerges as the top-5 performer. This diverse landscape underscores the valuable contributions of different foundation model types (unified, image-based, and video-only) within video action recognition, with each demonstrating strengths depending on the specific task and evaluation metric.



\subsubsection{Zero-shot and Open-vocabulary Action Recognition.}
Table \ref{tab:action_recognition_multimodal} compares Top-1 accuracy of ViFMs on zero-shot and harmonic-mean (HM) accuracy on base-to-novel (open-vocabulary) action recognition tasks. Most approaches report zero-short results smaller size datasets like HMDB51 \cite{Jhmdb-ICCV-2011} and UCF101 \cite{Ucf101-arxiv-2012}. The results show that IMP \cite{Akbari-NeurIPS-2023} (a UFM) integrates multiple modalities (text, image, video, and audio) via a unified encoder performs best on both datasets. Harmonic mean in case of base-novel generalization is computed by taking ahrmonic mean of accuracy of based classes and novel classes. The results shows that froster outperforms on K400 and UCF101 datasets whereas EZ-Clip (image-based) outperforms on JHMDB and SSv2 datasets. 

Both zero-shot and base-to-novel action recognition requires video-text understanding. The low performance in most cases (except UCF101 dataset) indicate that multi-modal action reconition has a huge scope for improvement in performance. Especially SSv2 dataset,where the actions are more complicated (e.g. opening/closing doors). 

\noindent
\textbf{Discussion.}
Overall, while ViFMs achieve very good fine-tuned performance on large datasets like Kinetics400 and Something-Something V2 (SSv2), limitations arise when dealing with smaller datasets like HMDB51 and UCF101. The significant size disparity makes fine-tuning large-scale models on these datasets challenging, resulting in a scarcity of reported results. Additionally, the zero-shot and base-to-novel performance on across most benchmarks leaves significant room for improvement. Techniques for constructing better representational spaces capable of capturing the crucial spatio-temporal context and vision-language semantics within videos are necessary for further progress.

Furthermore, only a few foundation models, like LaViLa \cite{Sun-NeurIPS-2022} and Avion \cite{Avion-arxiv-2023}, are currently designed for action recognition on datasets like EpicKitchen \cite{EpicKitchen-ECCV-2018} and Ego-4D \cite{Ego4D-CVPR-2022}. These datasets pose a unique challenge due to their complexity. They involve longer videos and are captured from an egocentric perspective (first-person view), which significantly differs from the typical third-person perspective videos used during pre-training. Even highly generic foundation models like VideoPrism \cite{Videoprism-arxiv-2024} struggle to adapt to this view translation, highlighting the need for models specifically designed to handle such complexities.

\subsubsection{Retrieval.}
Table \ref{tab:retrieval} compares state-of-the-art (SOTA) models for text-to-video retrieval, revealing a key insight: generalist models consistently outperform specialist models across diverse datasets (MSR-VTT \cite{Msrvtt-CVPR-2016}, DiDeMo \cite{Didemo-ICCV-2017}, and LSMDC \cite{Lsmdc-CVPR-2015}) and settings (fine-tuned and zero-shot). 

Delving deeper, unified models that incorporate not just video and text, but potentially audio modalities as well, seem to be achieving superior performance across various datasets. For instance, VAST excels on MSRVTT and DiDeMo benchmarks, while mPlug-2 (focusing on unified image-video) dominates on LSMDC under fine-tuned settings. Interestingly, when looking at zero-shot results, VAST remains the leader on MSRVTT and DiDeMo, whereas unmasked teacher models, built by inflating image models, perform best on LSMDC. This suggests that datasets like LSMDC, containing movie descriptions, benefit from the strong spatio-temporal understanding provided by unmasked teacher models employing the MDM objective.

Overall, the findings highlight the general promise of unified image-video approaches in video understanding tasks, as evidenced by their consistent performance across various benchmarks. InternVideo's \cite{Wang-arxiv-2022} exceptional performance further emphasizes the potential of well-designed generalist models for tackling diverse video understanding challenges.

\noindent
\textbf{Discussion.}
While ViFMs have achieved significant progress in recognition tasks, text-to-video retrieval remains a challenging area.  This task inherently requires effective multi-modal architectures that can capture the interaction between video content and textual descriptions. While standard multi-modal setups can enhance semantic understanding within the video domain, aligning video content with textual descriptions presents a significant hurdle. The challenge lies in bridging the gap between complex structures present in videos (including spatio-temporal aspects) and simple structures of language.

As previously discussed, masked video modeling techniques like those employed in VideoMAE \cite{Tong-NeurIPS-2022}, ST-MAE \cite{Feichtenhofer-NeurIPS-2022} can significantly improve a model's spatio-temporal understanding of videos. InternVideo takes this a step further by combining masked video modeling with a multi-modal contrastive objective using cross-model attention. This combined approach allows InternVideo to outperform most other approaches on text-to-video retrieval tasks by a substantial margin. However, despite this success, there's still significant room for improvement in overall retrieval performance. We infer that ViFMs should not only focus on advanced techniques for modeling causal behavior and temporal reasoning within videos, but also prioritize effective alignment between such complex video components and their corresponding textual descriptions. This will lead to improved video-language interaction and ultimately enhance retrieval performance.

\subsubsection{Spatio-temporal Video Understanding}

Table \ref{tab:tal_stal} compares the performance of various foundation models on two action localization tasks: Temporal Action Localization (TAL) and Spatio-temporal Action Localization (STAL). The TAL evaluation utilizes the ActivityNet \cite{Activitynet-CVPR-2015} and THUMOS14 \cite{Thumos-CVIU-2017} datasets, while the STAL evaluation employs the AVA \cite{Ava-CVPR-2018} and AVA-Kinetics \cite{Ava_Kinetic-arxiv-2005} datasets. As evident from the table, InternVideo \cite{Internvideo-arxiv-2022} emerges as the clear leader, achieving the best performance across all tasks.

This observation highlights the scarcity of ViFMs specifically designed for action localization tasks. Notably, all existing approaches in Table \ref{tab:tal_stal} leverage masked data modelling (MDM) (Generative Objective) during pre-training.   Specifically, MaskFeat \cite{MaskFeat-CVPR-2022}, ST-MAE \cite{Feichtenhofer-NeurIPS-2022}, and VideoMAE \cite{Tong-NeurIPS-2022} are unimodal approaches built only upon mask modelling. Conversely, UMT \cite{Li-ICCV-2023}, InternVideo \cite{Internvideo-arxiv-2022}, and VideoPrism \cite{Videoprism-arxiv-2024} combine both mask modelling (generative objective) and video-text contrastive learning (discriminative objective) for improved performance.

\begin{table}
\centering
\caption{Comparing the fine-tuned performance of state-of-the-art (SOTA) Vi-FMs on ActivityNet \cite{Activitynet-CVPR-2015} for Temporal Action Localization (TAL) and AVA \cite{Ava-CVPR-2018} and AVA-Kinetics \cite{Ava_Kinetic-arxiv-2005} for Spatio-temporal action localization (STAL). The highlighted entries indicate the best performing methods.}
\setlength\tabcolsep{6.0pt}
\small
\begin{tabular}{|l|cc|cc|}
\toprule
 \multirow{2}{*}{Method} & \multicolumn{2}{c|}{TAL} & \multicolumn{2}{c|}{STAL} \\
 \cline{2-5}
 & ActivityNet & THUMOS14 & AVA & AVA-Kinetics \\
\midrule
MaskFeat-L \cite{MaskFeat-CVPR-2022} & - & - & 37.8 & - \\
ST-MAE-L \cite{Feichtenhofer-NeurIPS-2022} & - & - & 37.3 & - \\
VideoMAE-L \cite{Tong-NeurIPS-2022} & - & - & 39.3 & - \\
VideoMAEv2 \cite{VideoMAEv2-CVPR-2023a} & - & 69.6 &  \textbf{42.6} &  \textbf{43.9} \\
UMT-B \cite{Li-ICCV-2023} & - & - & 33.5 & - \\
UMT-L \cite{Li-ICCV-2023} & - & - & 39.8 & - \\
InternVideo \cite{InternVid-arxiv-2023} &  \textbf{39.0} &  \textbf{71.6} & 41.0 & 42.5 \\
VideoPrism-B \cite{Videoprism-arxiv-2024} & 36.6 & - & 30.6 & 31.8 \\
VideoPrism-g \cite{Videoprism-arxiv-2024} & 37.8 & - & 36.2 & 37.3 \\
\bottomrule
\end{tabular}
\label{tab:tal_stal}
\end{table}

\noindent
\textbf{Discussion.}
The dominance of MDM in pre-training the ViFMs presented in Table \ref{tab:tal_stal} highlights its effectiveness in modeling spatio-temporal interactions within videos. This capability is crucial for action localization tasks, allowing video-only models to achieve good performance. However, video-only models often struggle with capturing the semantic aspects of actions.Here, additional modalities like text can provide valuable semantic cues. While video-text interaction improves performance, it may still lack specific action knowledge as noted by Wang \etal~\cite{Paxion-NeurIPS-2023}.  Therefore, effectively combining generative and discriminative objectives, such as mask modelling and video-text contrastive learning, is crucial for achieving a comprehensive understanding of motion, temporality, and semantics within a video.


While the presented ViFMs demonstrate strong performance, it's important to acknowledge the existence of additional action localization datasets like UCF24 \cite{Ucf101-arxiv-2012} and JHMDB \cite{Jhmdb-ICCV-2011}.  Our observations suggest that these existing ViFMs are not being evaluated on these benchmark datasets. We hypothesize that the smaller scale of these datasets may limit the effectiveness of fine-tuning large-scale models.  Datasets like UCF-MAMA\cite{modi2022video}, VIRAT\cite{Tinyvirat-ICPR-2021} are high resolution, contain multiple actors and actions happening at the \textit{same} time : benchmarking existing models on those should give a reasonable estimate of their ability to understand the arrow of time. Future research could explore strategies to adapt these models for optimal performance on diverse action localization tasks and datasets.

This subsection highlights that while our models achieve good performance on various action recognition benchmarks, action localization remains a significant challenge. Action localization involves recognizing action class as well as predicting a box around an actor. Recognition step presents a significant challenge since action-prediction is a higher level concept, and not simple perceptual recognition. Therefore, qunatitative results on action-detection are observed to be lower in practice. Although unimodal models like VideoMAE v2 utilize masked data modeling (MDM) to capture spatio-temporal information, incorporating vision-language understanding can further improve performance. This is because multimodal approaches equip ViFMs with additional semantic understanding, leading to more robust action localization. Furthermore, such multimodal foundation models demonstrate the potential for extending action recognition towards open-vocabulary tasks, as evidenced by recent works \cite{Xue-arxiv-2023a, Chatterjee-NeurIPS-2024}.


\subsection{Descriptive Understanding Tasks} 
This section focuses on evaluating how well state-of-the-art ViFMs perform on tasks that rely on understanding video content through text descriptions. We compare their performance on Video Question Answering (VideoQA) and Video Captioning tasks.

\begin{table*}[t]
\centering
\caption{Comparing the fine-tuned performance of state-of-the-art (SOTA) generalist and specialist models for \textbf{VideoQA} (accuracy) on MSRVTT \cite{Msrvtt_vqa-ECCV-2018}, LSMDC \cite{Lsmdc_vqa-arxiv-2016}, and MSVD \cite{Msvd_vqa-ACMMM-2017} datasets, and \textbf{Zero-shot Video Captioning} (CIDEr) tasks on MSRVTT \cite{Msrvtt-CVPR-2016}, MSVD \cite{Msvd-ACLHLT-2011}, and YooCook2 \cite{Youcook2-AAAI-2018} datasets. The highlighted entries indicate the best performing methods.}
\setlength\tabcolsep{4.0pt}
\footnotesize
\begin{tabular}{|c|l|c|cc|cc|c|ccc|}
\toprule
 & \multirow{3}{*}{Method} & \multirow{2}{*}{Arch.} & \multicolumn{5}{c|}{VQA} & \multicolumn{3}{c|}{Captioning} \\
 \cline{4-11}
  & & & \multicolumn{2}{c}{\underline{MSRVTT}} & \multicolumn{2}{c}{\underline{LSMDC}} & \underline{MSVD} & MSRVTT & MSVD & YouCook2 \\ 
 & & Type & MC & QA & MC & FiB & QA & & & \\
\midrule
\multicolumn{11}{|c|}{Fine-Tuned} \\
\hline
\multirow{12}{*}{\rotatebox{90}{Generalist}} & ClipBERT \cite{Clipbert-CVPR-2021} & Dual-Enc & 88.2 & 37.4 & - & - & - & - & - & - \\ 
& RTQ \cite{Wang-ACMMM-2023}  & Dual-Enc & - & 42.1 & - & - & - & 69.3 & 123.4 & - \\
& Singularity \cite{Lei-arxiv-2022} & Dual-Enc & 92.1 & 43.5 & - & - & - & - & - & - \\
& UMT-B \cite{Li-ICCV-2023} & Dual-Enc & 96.3 & 44.9 & - & - & 49.5 & - & - & - \\
& UMT-L \cite{Li-ICCV-2023} & Dual-Enc & 97.3 & 47.1 & - & - & 55.2 & - & - & - \\
& VideoCoca \cite{Yan-arxiv-2023b} & Enc-Dec & - & 46.3 & - & - & 56.9 & 73.2 & - & 128.0 \\
\cline{3-11}
& All-in-One \cite{Wang-CVPR-2023}  & Joint-Enc & 92.3 & 46.8 & 84.4 & - & 48.3 & - & - & - \\
& ALPRO \cite{Alpro-CVPR-2022} & Mix-Enc & - & 42.1 & - & - & 46.3 & - & - & - \\
& Clover \cite{Clover-CVPR-2023} & Mix-Enc & 95.2 & 44.1 & 83.7 & 54.1 & 52.4 & - & - & - \\
& Hitea \cite{Hitea-CVPR-2023} & Mix-Enc & 97.2 & 45.4 & 85.8 & 54.6 & 55.6 & 62.5 & 145.1 & - \\
& LAVANDER \cite{Li-CVPR-2023} & Joint-Enc & \textbf{97.4} & 45.0 & \textbf{87.0} & \textbf{57.1} & 56.6 & 60.1 & 150.7 & - \\
& SimVTP \cite{Ma-arxiv-2022} & Enc-Dec & 93.6 & 44.7 & 83.7 & - & 48.9 & - & - & - \\ 
& UniVL \cite{Luo-arxiv-2020} & Mix-Enc & - & - & - & - & - & - & - & 127.0 \\
& VindLU-L \cite{Vindlu-CVPR-2023} & Dual-Enc & 95.5 & 44.6 & - & - & - & - & - & - \\ 
& VIOLET \cite{Violet-arxiv-2022} & Dual-Enc & 91.9 & 43.9 & 82.8 & 53.7 & 47.9 & - & - & - \\
\cline{3-11}
& mPLUG-2 \cite{Mplug2-arxiv-2023} & Mix-Enc & - & 48.0 & - & - & 58.1 & \textbf{80.3} & 165.8 & - \\
& MaMMUT \cite{Mammut-MLR-2023} & Dual-Enc & - & 49.5 & - & - & 60.2 & 73.6 & \textbf{195.6} & - \\
& MERLOT \cite{Merlot-NeurIPS-2021} & Dual-Enc & 90.9 & 43.9 & 82.8 & 53.7 & 47.9 & - & - & - \\
& OmniVL \cite{Omivl-NeurIPS-2022} & Dual-Enc & - & 44.1 & - & - & 51.0 & - & - & 116.0 \\
& Smaug \cite{Smaug-CVPR-2023} & Mix-Enc & 92.9 & 44.5 & - & - & - & - & - & - \\
& X$^2$VLM \cite{Zeng-arxiv-2023} & Mix-Enc & - & 45.5 & - & - & 54.6 & - & - & - \\
& VALOR \cite{Valor-arxiv-2023} & Multi-Enc & - & 49.2 & - & - & 60.0 & 74.0 & 178.5 & - \\
& VAST \cite{Vast-arxiv-2023} & Mixed-Enc & - & 50.1 & - & - & 60.2 & 78.0 & - & \textbf{198.8} \\
\cline{3-11}
& VideoChat \cite{VideoChat-arxiv-2023} & LMM & - & 45.0 & - & - & 56.3 & - & - & - \\
& Video-LLaMA \cite{Video_Llama-arxiv-2023} & LMM & - & 29.6 & - & - & 51.6 & - & - & - \\
& Video-LLaVA \cite{Video_Llava-arxiv-2023} & LMM & - & \textbf{59.2} & - & - & \textbf{70.7} & - & - & - \\
& Video-ChatGPT \cite{Video_Chatgpt-arxiv-2023} & LMM & - & 49.3 & - & - & 64.9 & - & - & - \\
\hline
\multirow{4}{*}{\rotatebox{90}{Special}} & Just-Ask \cite{Yang-ICCV-2021}  & Dual-Enc & - & 41.5 & - & - & 46.3 & - & - & - \\
& CLIP4Caption \cite{tang2021clip4caption}  & Dual-Enc & - & - & - & - & - & 57.7 & - & -\\
& SwinBERT \cite{Swinbert-CVPR-2022}  & Dual-Enc & - & - & - & - & - & 53.8 & 120.6 & - \\
& MV-GPT \cite{Seo-CVPR-2022}  & Dual-Enc & - & - & - & - & - & 60.0 & - & - \\
& Text-KG \cite{Gu-CVPR-2023} & Mix-Enc & - & - & - & - & - & 60.8 & - & 133 \\
\hline
\multicolumn{11}{|c|}{Zero-Shot} \\
\hline
\multirow{7}{*}{\rotatebox{90}{Generalist}} & VideoCoca \cite{Yan-arxiv-2023b} & Enc-Dec & - & - & - & - & - & 27.1 & - & 34.3 \\
& Hitea \cite{Hitea-CVPR-2023} & Mix-Enc & - & 21.7 & - & - & 37.4 & - & - & - \\
& mPlug-2 \cite{Mplug2-arxiv-2023} & Mix-Enc & - & 43.8 & - & - & 55.3 & - & - & - \\
& Distill-VLM \cite{Zhao-arxiv-2024} & LMM & - & 24.4 & - & - & - & 48.2 & - & - \\
& PaML2-VAdapter \cite{Palm2_VAadapter-arxiv-2024} & LMM & - & 19.6 & - & - & 40.5 & 47.7 & - & - \\
& VideoPrism-B \cite{Videoprism-arxiv-2024} (w/PaLM-1B) & Dual-Enc & - & 28.5 & - & - & 39.5 & 40.3 & - & 52.3 \\
& VideoPrism-B \cite{Videoprism-arxiv-2024} (w/PaLM-8B) & Dual-Enc & - & 32.0 & - & - & 47.1 & 38.5 & - & 63.6 \\
\bottomrule
\end{tabular}
\label{tab:vqa_captioning}
\end{table*}  

\subsubsection{Video Question Answering (VideoQA)}
Table \ref{tab:vqa_captioning} (Left column) compares the performance of different models on VideoQA (MSR-VTT \cite{Msrvtt_vqa-ECCV-2018}, LSMDC \cite{Lsmdc_vqa-arxiv-2016}, and MSVD \cite{Msvd_vqa-ACMMM-2017} datasets) task. For VideoQA, different models are better depending on the type of question. LAVENDER \cite{Li-CVPR-2023} (a video-based ViFM) does better on multiple-choice (MC) and Fill-in-Blank (FiB) questions on both MSR-VTT \cite{Msrvtt_vqa-ECCV-2018} and LSMDC \cite{Lsmdc_vqa-arxiv-2016} datasets. On the other hand, Video-LLaVa \cite{Video_Llava-arxiv-2023} (a Large Multi-modal Model) is better at answering open-ended questions on both MSRVTT \cite{Msrvtt_vqa-ECCV-2018} and MSVD \cite{Msvd_vqa-ACMMM-2017} benchmarks. 

We can infer from the table that video-based models like LAVENDER \cite{Li-CVPR-2023} excel at multiple-choice (MC) and fill-in-the-blank (FiB) questions. This is likely because these tasks require strong cross-modal understanding, which LAVENDER achieves by encoding both vision and text information within a single joint encoder. This allows the model to learn relationships between the modalities from the very beginning. Additionally, its Masked Language Modeling (MLM) pre-training objective closely resembles the FiB setting, further boosting performance. However, open-ended questions demand not only video-text interaction but also rich natural language understanding. Fortunately, modern LLMs demonstrate impressive language semantic knowledge. Combining these LLMs with video models leads to state-of-the-art performance on open-ended VideoQA tasks.

\noindent
\textbf{Discussion.}
Beyond question types (MC vs. open-ended), another crucial dimension in VideoQA is reasoning capability. This can be broken down into two categories: Factoid reasoning \cite{Castro-ACL-2022, Castro-ICCL-2022, Msvd_vqa-ACMMM-2017, Msrvtt-CVPR-2016} and Inference reasoning \cite{Xiao-CVPR-2021, Li-CVPR-2022, Star-NeurIPS-2021, Clevrer-ICLR-2019}. Existing ViFMs primarily focus on Factoid questions \cite{Merlot-NeurIPS-2021, Clover-CVPR-2023, Li-CVPR-2023}, which involve retrieving factual information directly from the video. However, a gap remains in tackling Inference-based questions that require a deeper understanding of the video content, including both dense spatio-temporal relationships (how objects and actions unfold over time) and causal relationships (understanding cause-and-effect). 

Overall, Multi-modal video tasks like retrieval and VQA remain far from achieving their full potential. A deeper analysis reveals a more critical issue: the limited scale of video-text datasets compared to image-text datasets. This data scarcity hinders effective learning for video foundation models.  Further compounding the problem is the data quality of many video-text datasets, often generated by scraping online videos and text, which introduces significant noise. The presence of noisy data further complicates the already challenging task of multi-modal learning for video models.

\subsubsection{Video Captioning.}
Table \ref{tab:vqa_captioning} (right column) compares video captioning performance on various datasets (MSRVTT \cite{Msrvtt-CVPR-2016}, MSVD \cite{Msvd-ACLHLT-2011}, and YouCook2 \cite{Youcook2-AAAI-2018} datasets). The results show specialization: mPLUG-2 \cite{Mplug2-arxiv-2023} (image-video-text) performs well on MSRVTT \cite{Msrvtt-CVPR-2016}, MaMMUT \cite{Mammut-MLR-2023} (image-video-text) excels on MSVD \cite{Msvd-ACLHLT-2011}, and VAST \cite{Vast-arxiv-2023} (video-audio-text) dominates YouCook2 \cite{Youcook2-AAAI-2018} (cooking videos). This suggests the importance of incorporating relevant modalities for specific video content. Notably, VAST's strong performance on YouCook2 highlights the informative nature of audio in video captioning. Overall, the superiority of \textit{unified models} across datasets underscores the potential of exploring additional modalities for captioning tasks.

\noindent
\textbf{Discussion.}
While Table \ref{tab:vqa_captioning} focuses on video captioning \cite{Vinyals-TPAMI-2016, Gu-arxiv-2017, Msrvtt-CVPR-2016}, dense video captioning \cite{Chang-NN-2022, Deng-CVPR-2021}, which generates multiple captions describing events throughout a video, remains an unsolved challenge. This task demands a deeper understanding of spatio-temporal relationships, an area where ViFMs are still under development.

\subsection{Video Content Generation and Manipulation}

We discuss Text-To-Video(T2V) generation tasks, generates video given its textual description. For T2V generation GANs, auto-regressive, and diffusion models are the dominant choice. 

\begin{table}[bht]
\centering
\caption{Comparing the fine-tuned performance of state-of-the-art (SOTA) generalist and specialist models for zero-shot Text-To-Video Generation on MSR-VTT \cite{Msrvtt-CVPR-2016} and UCF-101 \cite{Ucf101-arxiv-2012} dataset. The highlighted entries indicate the best performing methods.}
\setlength\tabcolsep{6.0pt}
\small
\begin{tabular}{|c|l|cc|cc|}
\toprule
 & \multirow{3}{*}{Method} & \multicolumn{2}{c|}{MSR-VTT}  & \multicolumn{2}{c|}{UCF-101} \\
 \cline{3-6}
 & & ClipSIM ($\uparrow$) & FVD ($\downarrow$) & FVD ($\downarrow$) & IS ($\uparrow$) \\
\midrule
\multirow{2}{*}{\rotatebox{90}{Gen}} & InternVid \cite{InternVid-arxiv-2023} & 0.2951 & - & 616.51 & 21.04 \\
& Make-A-Video \cite{MakeAVideo-arxiv-2022} & \textbf{0.3049} & - & 367.23 & 33.00 \\
& PYoCo \cite{Pyoco-ICCV-2023} & - & - & 355.19 & 47.76 \\
& SVD \cite{SVD-arxiv-2023} & - & - & \textbf{242.02} & - \\
& VideoPoet \cite{Videopoet-arxiv-2023} & \textbf{0.3049} & \textbf{213.00} & 355.00 & \textbf{38.44} \\
& VideoLaVIT \cite{VideoLavit-arxiv-2024} & 0.3010 & 169.51 & 274.96 & 37.96 \\
& VideoLDM \cite{VideoLDM-CVPR-2023} & 0.2929 & - & 550.61 & 33.45 \\
& VideoComposer \cite{Videocomposer-NeurIPS-2023} & 0.2932 & 580.00 & - & - \\
\hline
\multirow{4}{*}{\rotatebox{90}{Special}} & CogVideo \cite{Cogvideo-arxiv-2022} & 0.2631 & 1294.00 & 702 & 25.27 \\
& MagicVideo \cite{Magicvideo-arxiv-2022} & - & 998.00 & 655 & - \\
& VideoFactory \cite{Videofactory-arxiv-2023} & 0.3005 & - & 410 & - \\
\bottomrule
\end{tabular}
\label{tab:generative}
\end{table}   

\subsubsection{Text-To-Video Generation}
Table \ref{tab:generative} showcases the performance of different models on the Text-to-Video generation task, tested on MSR-VTT \cite{Msrvtt-CVPR-2016} and UCF101 \cite{Ucf101-arxiv-2012} datasets.

Looking at the MSR-VTT \cite{Msrvtt-CVPR-2016} results, VideoPoet \cite{Videopoet-arxiv-2023}, a model utilizing a unified image-video framework, stands out by outperforming all others across both CLIPSim \cite{Godiva-arxiv-2021} and FVD \cite{FVD-arxiv-2018} metrics. However, on UCF101, different strengths emerge. InternVid, another ViFM, takes the lead when using the IS \cite{Saito-IJCV-2020} metric. Interestingly, for this dataset, SVD \cite{SVD-arxiv-2023}, a diffusion-based foundation model, surpasses its competitors in terms of FVD \cite{FVD-arxiv-2018}. This indicates SVD's \cite{SVD-arxiv-2023} ability to generate videos with exceptionally high visual fidelity within the context of UCF101 \cite{Ucf101-arxiv-2012}.

\noindent
\textbf{Discussion.}
Despite significant progress in video generation over the past few years, real-world integration remains challenging. High computational complexity and time-consuming processes are major hurdles. Generating a single minute of video can take hours. However, generative models are continuously improving their ability to maintain temporal consistency. This paves the way for integration with existing ViFMs to enhance the representation space.

\begin{figure*}[th]
    \centering  
    \begin{subfigure}
        \centering
	    \includegraphics[width=1\textwidth]{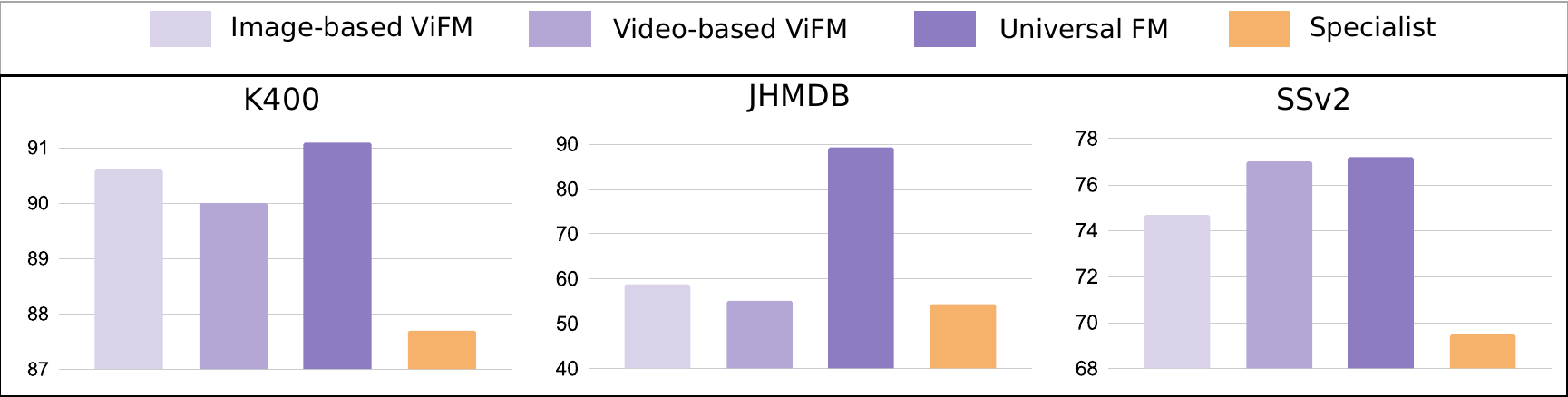}
        \caption{Fine-tuning results of Image-based, Video-based, and Universal foundational models against Specialist models for \textbf{video action recognition} task on k400 \cite{Kinetics400-arxiv-2017} (left), JHMDB \cite{Jhmdb-ICCV-2011} (center) and SSv2 \cite{SomethingSomethingv2-ICCV-2017} (right) datasets.}
        \label{fig:action_recognition_graphs}
        \vspace{0.2cm}
    \end{subfigure}
    \begin{subfigure}
        \centering
	\includegraphics[width=1\textwidth]{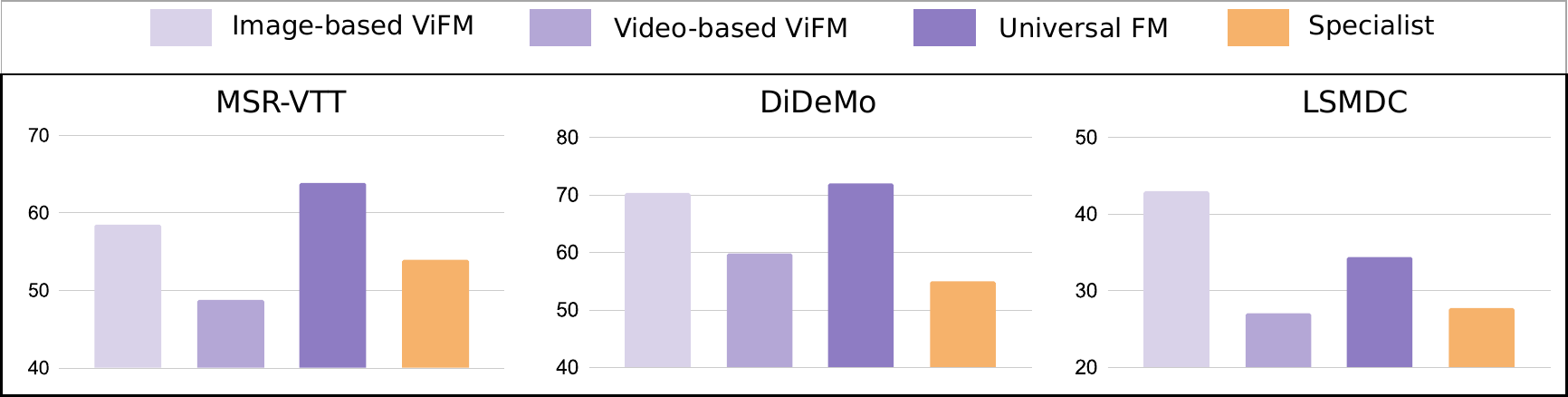}
        \caption{Fine-tuning results of Image-based, Video-based, and Universal foundational models against Specialist models for \textbf{video retrieval} task on MSRVTT \cite{Msrvtt-CVPR-2016} (left), DIDEMO \cite{Didemo-ICCV-2017} (center) and LSMDC \cite{Lsmdc-CVPR-2015} (right) datasets.}
        \label{fig:video_retrieval_results}
        \vspace{0.2cm}
    \end{subfigure}
    \begin{subfigure}
        \centering
	\includegraphics[width=1\textwidth]{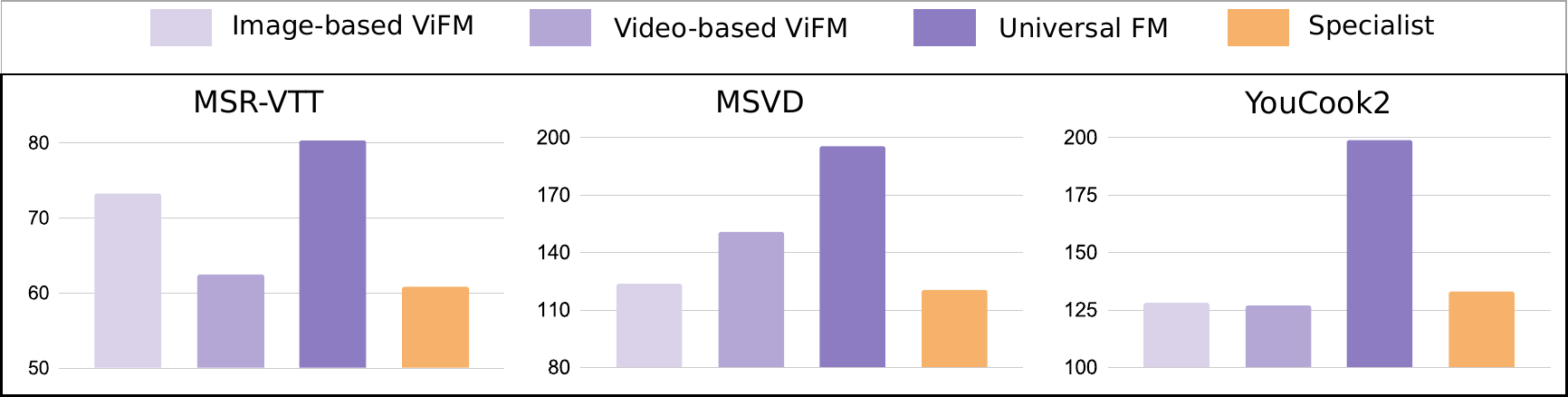}
        \caption{Fine-tuning results of Image-based, Video-based, and Universal foundational models against Specialist models for \textbf{video captioning} task on MSRVTT \cite{Msrvtt-CVPR-2016} (left), MSVD \cite{Msvd-ACLHLT-2011} (center) and YouCook2 \cite{Youcook2-AAAI-2018} (right) datasets.}
        \label{fig:video_captioing_results}
    \end{subfigure}
    \label{fig:analysis_results}
\end{figure*}

\subsection{Analysis of Results} 
Figure \ref{fig:action_recognition_graphs}, \ref{fig:video_retrieval_results}, and \ref{fig:video_captioing_results} presents a general comparison of image-based foundational models, video-based models, unified-models and task-specific specialist models. Note that task-specific models are finetuned for a particular task being analyzed and have been shown by a separate color (red). Across all the tasks a \textit{general} trend is being observed: i.e. Image based foundational models perform better than video based models. We attribute this to the fact that there is a large amount of pretraining data available for image-based models whereas video-based models significantly lag behind this aspect. A notable exception is video models performing better on SSV2 in Action Recognition: this might be because SSV2 contains challenging action classes whose definition changes with the order of frames and requires understanding the arrow of time. Video Modelling helps establish much needed temporal correspondance in this case. 

Across all the tasks, unified models perform much better than both image/video based models: this shows that joint representational learning of image/video modalities helps achieve significant performance gains. Finally, we note that all unified models outperform task-specific models. This highlights a desirable generalization behaviour: we are able to train foundational models on massive datasets, outperform task-specific specialists, as well as \textit{adapt} a given model for several downstream tasks with minimal compute.

\section{Challenges and Future Directions}
\label{sec:future}

Multi-modal video foundation models encounter challenges in accessing extensive training data, limiting their development compared to image-text models. Existing video-text datasets are often small and noisy, hindering robust representation learning. Future research must prioritize creating high-quality, large-scale video-text datasets and exploring data cleaning, augmentation, and alternative pretraining methods. Additionally, existing video foundation models struggle with temporal consistency, object-centric tasks, and adaptability to diverse contexts. However, recent advancements, such as the generative auto-regressive model SORA \cite{SORA-2024}, offer insights into addressing temporal consistency issues. ViFMs can draw inspiration from such models to improve their performance. Integrating diverse self-supervised learning techniques \cite{Clmae-WACV-2024} and adapter modules provides avenues for domain adaptation and various video understanding tasks. The subsequent subsections present possible future research directions in the domain of ViFMs.

\subsection{Addressing Ethical Considerations.}
As Video Foundation Models (ViFMs) find increasing application in real-world scenarios, akin to ChatGPT \cite{ChatGPT-2023} and Amazon SageMaker \cite{AmazonSagemaker-2021}, addressing ethical concerns becomes crucial. Future research should focus on mitigating biases through debiasing datasets and fairness metrics, promoting transparency and explainability to build trust, and establishing responsible use guidelines throughout the ViFM development lifecycle. By actively addressing these ethical considerations, we can ensure ViFMs are deployed responsibly, maximizing their positive impact on real-world applications.

\subsection{Long-Form Video Understanding.} 
Achieving long-form video understanding \cite{Egoschema-NeurIPS-2024} with ViFMs presents a significant challenge due to the extensive memory requirements for processing extended sequences. Recent efforts have recognized this hurdle and begun to explore solutions, such as memory consolidation mechanisms \cite{Song-arxiv-2023, Balavzevic-arxiv-2024} and memory-efficient attention \cite{RingAttention-arxiv-2024}. However, to truly unlock the potential of ViFMs in this domain, integrating causal reasoning could be pivotal. By incorporating causal reasoning \cite{Clevrer-ICLR-2019} into ViFMs, we can enhance their ability to comprehend extended video content by enabling them to answer fundamental questions like "why," "what next," and "what if." This deeper understanding facilitated by causal reasoning could revolutionize long-form video understanding, allowing ViFMs to recognize event order, direction of causality, and detailed relationships between actors, actions, and objects. Moreover, integrating causal reasoning may enhance the robustness \cite{Schiappa-CVPR-2023} of ViFMs and improve their ability to handle occlusions \cite{Modi-NeurIPS-2023} and other challenges commonly encountered in real-world video data. Therefore, while addressing the memory constraints is crucial, integrating causal reasoning into ViFMs offers a promising avenue for achieving comprehensive long-form video understanding.

\subsection{Viewpoint Invariance.}
ViFMs excel in traditional video settings (i.e., third-person viewpoint), but limitations arise in understanding different viewpoints like egocentric \cite{Ego4D-CVPR-2022} or birds-eye view \cite{Peng-WACV-2023}. Future research can delve into viewpoint-invariant representations for dynamic scenes. Inspired by the human ability to mentally rotate objects, neural representations like NeRF \cite{Mildenhall-ECCV-2020} can be explored to encode a continuous 3D representation within ViFMs. Additionally, methods that project an agent's limited view to a common reference frame \cite{Philion-ECCV-2020} and establish correspondences across views hold promise for learning robust representations despite dynamic exploration. By pursuing these directions, ViFMs can be equipped to handle different viewpoints and varying camera paths, ultimately leading to significant advancements in action localization and a deeper understanding of dynamic video content.

\subsection{Domain Adaptation}
Domain adaptation refers to the ability of models to perform well in new environments (lighting, locations, etc.) or domains that differ from those they were trained on. Some studies \cite{Schiappa-ACM-2023, Kumar-arxiv-2023} provide directions to make traditional video models robust against such changes by suggesting augmentations \cite{Zara-CVIU-2024} and specific tuning \cite{Kareer-arxiv-2024}. However, research in domain adaptation for foundation models is still lacking. To seamlessly integrate these powerful models into real-world applications, future work should explore methods to make ViFMs robust against domain shifts.

\subsection{Improving Efficiency}
Despite promising results, ViFM's resource demands pose a challenge for edge deployment. These models often have hundreds of millions to billions of parameters, leading to longer training and inference times. This consequently limits their deployment on edge devices for real-time inference. To address this challenge, a key future direction in the integration of video foundation models for edge devices involves developing efficient deployment strategies to overcome resource constraints and enable seamless inferencing at the edge, citing \cite{Vila-arxiv-2023} as an example. This entails exploring novel approaches to optimize model architecture and reduce computational overhead, as well as investigating innovative techniques for model compression and quantization to facilitate deployment on resource-constrained edge devices without compromising performance. By addressing these challenges, researchers can pave the way for widespread adoption of video foundation models in edge computing environments, unlocking their potential to power a diverse range of high-impact applications across industries.

\section{Conclusion}
This survey offers a comprehensive and, to the best of our knowledge, the first-of-its-kind in-depth exploration of Video Foundation Models (ViFMs). We commenced by establishing a foundation with discussions on video understanding tasks, relevant architectures, pretraining datasets, and approaches for ViFM pretraining. We categorize core methodologies for ViFM creation into three primary techniques: Inflating Image Models, Video-based models (focusing on video or video-text pretraining), and Unified Image-Video ViFMs (applicable to both image and video tasks). By comparing the performance of various ViFMs on video tasks and offering insights based on methodologies and results, we aim to equip the research community with a comprehensive overview of existing ViFMs, while also highlighting critical areas for future exploration. This, we believe, will foster further advancements in video modeling and unlock the full potential of ViFMs.
\section{Acknowledgements}
This project is funded by \textit{Innovation Funds}, Denmark.


\bibliographystyle{ACM-Reference-Format}
\bibliography{references}

\end{document}